\documentclass[10pt,twocolumn,letterpaper]{article}

\usepackage{cvpr}              

\usepackage{graphicx}
\usepackage{amsmath}
\usepackage{amssymb}
\usepackage{booktabs}
\usepackage{cases}

\usepackage{amsmath,amsfonts,bm}

\def\figref#1{figure~\ref{#1}}

\def\secref#1{section~\ref{#1}}

\def\eqref#1{equation~\ref{#1}}

\def\1{\bm{1}}

\def\vf{{\bm{f}}}
\def\vg{{\bm{g}}}
\def\vh{{\bm{h}}}

\def\vp{{\bm{p}}}
\def\vq{{\bm{q}}}
\def\vr{{\bm{r}}}

\def\vu{{\bm{u}}}
\def\vv{{\bm{v}}}
\def\vw{{\bm{w}}}
\def\vx{{\bm{x}}}

\def\vz{{\bm{z}}}

\def\mQ{{\bm{Q}}}

\def\mT{{\bm{T}}}

\DeclareMathAlphabet{\mathsfit}{\encodingdefault}{\sfdefault}{m}{sl}
\SetMathAlphabet{\mathsfit}{bold}{\encodingdefault}{\sfdefault}{bx}{n}

\newcommand{\softmax}{\mathrm{softmax}}

\DeclareMathOperator*{\argmin}{arg\,min}

\usepackage[utf8]{inputenc} 
\usepackage[T1]{fontenc}    
\usepackage[pagebackref=true,breaklinks=true,letterpaper=true,colorlinks,bookmarks=false]{hyperref}      
\usepackage{url}            
\usepackage{booktabs}       
\usepackage{amsfonts}       
\usepackage{nicefrac}       
\usepackage{microtype}
\usepackage{hhline}
\usepackage{xcolor}
\usepackage{color}
\usepackage{times}
\usepackage{epsfig}
\usepackage{graphicx}
\usepackage{amsmath}
\usepackage{amssymb}
\usepackage{amsthm}
\usepackage{array}
\usepackage{multirow}
\usepackage{booktabs}
\usepackage{float}

\renewcommand{\figref}[1]{Fig.~\ref{fig:#1}}

\newcommand{\tabref}[1]{Tab.~\ref{tab:#1}}
\newcommand{\equref}[1]{Eq.~(\ref{equ:#1})}
\renewcommand{\secref}[1]{Sec.~\ref{sec:#1}}

\newcommand{\myinput}[1]{}
\renewcommand{\paragraph}[1]{\noindent{\bf{#1}}}

\usepackage[accsupp]{axessibility}  

\usepackage{multirow}
\usepackage{subcaption}
\usepackage[font={small}]{caption}
\usepackage{booktabs, dcolumn}
\usepackage{pifont} 
\usepackage[capitalize]{cleveref}
\crefname{section}{Sec.}{Secs.}
\Crefname{section}{Section}{Sections}
\Crefname{table}{Table}{Tables}
\crefname{table}{Tab.}{Tabs.}

\def\cvprPaperID{6427} 
\def\confName{CVPR}
\def\confYear{2022}

\begin{document}

\title{RCP: Recurrent Closest Point for Scene Flow Estimation on 3D Point Clouds}

\author{ 
Xiaodong Gu$^{1}$\hspace{0.6cm}
Chengzhou Tang$^{2}$\hspace{0.6cm}
Weihao Yuan$^{1}$\hspace{0.6cm}
Zuozhuo Dai$^{1}$\hspace{0.6cm}
Siyu Zhu$^{1}$\hspace{0.6cm}
Ping Tan$^{12}$\hspace{0.6cm}\\ 
${}^{1}$Alibaba Group\hspace{1.5cm}${}^{2}$Simon Fraser University 
}

\maketitle

\begin{abstract}
3D motion estimation including scene flow and point cloud registration has drawn increasing interest.
Inspired by 2D flow estimation, recent methods employ deep neural networks to construct the cost volume for estimating accurate 3D flow.
However, these methods are limited by the fact that it is difficult to define a search window on point clouds because of the irregular data structure.
In this paper, we avoid this irregularity by a simple yet effective method.
We decompose the problem into two interlaced stages, where the 3D flows are optimized point-wisely at the first stage and then globally regularized in a recurrent network at the second stage. 
Therefore, the recurrent network only receives the regular point-wise information as the input.
In the experiments, we evaluate the proposed method on both the 3D scene flow estimation and the point cloud registration task. 
For 3D scene flow estimation, we make comparisons on the widely used FlyingThings3D~\cite{ft3d} and KITTI~\cite{kitti2015} datasets. 
For point cloud registration, we follow previous works and evaluate the data pairs with large pose and partially overlapping from ModelNet40~\cite{modelnet}. 
The results show that our method outperforms the previous method and achieves a new state-of-the-art performance on both 3D scene flow estimation and point cloud registration, which demonstrates the superiority of the proposed zero-order method on irregular point cloud data.
Our source code is available at \url{https://github.com/gxd1994/RCP}.

\end{abstract}
\begin{figure}[t]
  \centering

  \begin{subfigure}{0.48\columnwidth}
      \centering
      \includegraphics[width=1\columnwidth, trim={0cm 0cm 0cm 1cm}, clip]{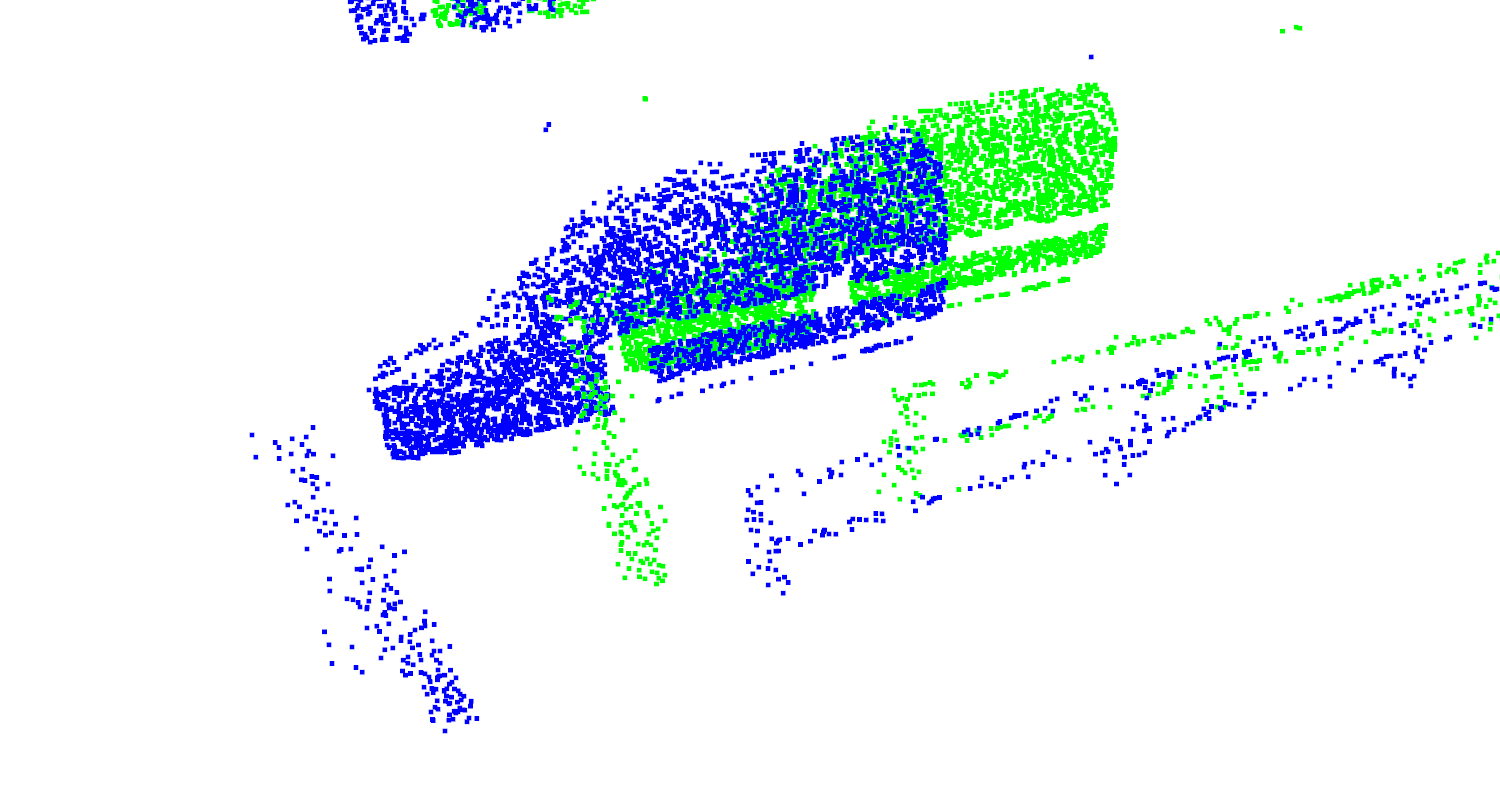}
      \caption*{iter=0}
  \end{subfigure}
  \begin{subfigure}{0.48\columnwidth}
      \centering
      \includegraphics[width=1\columnwidth, trim={0cm 0cm 0cm 1cm}, clip]{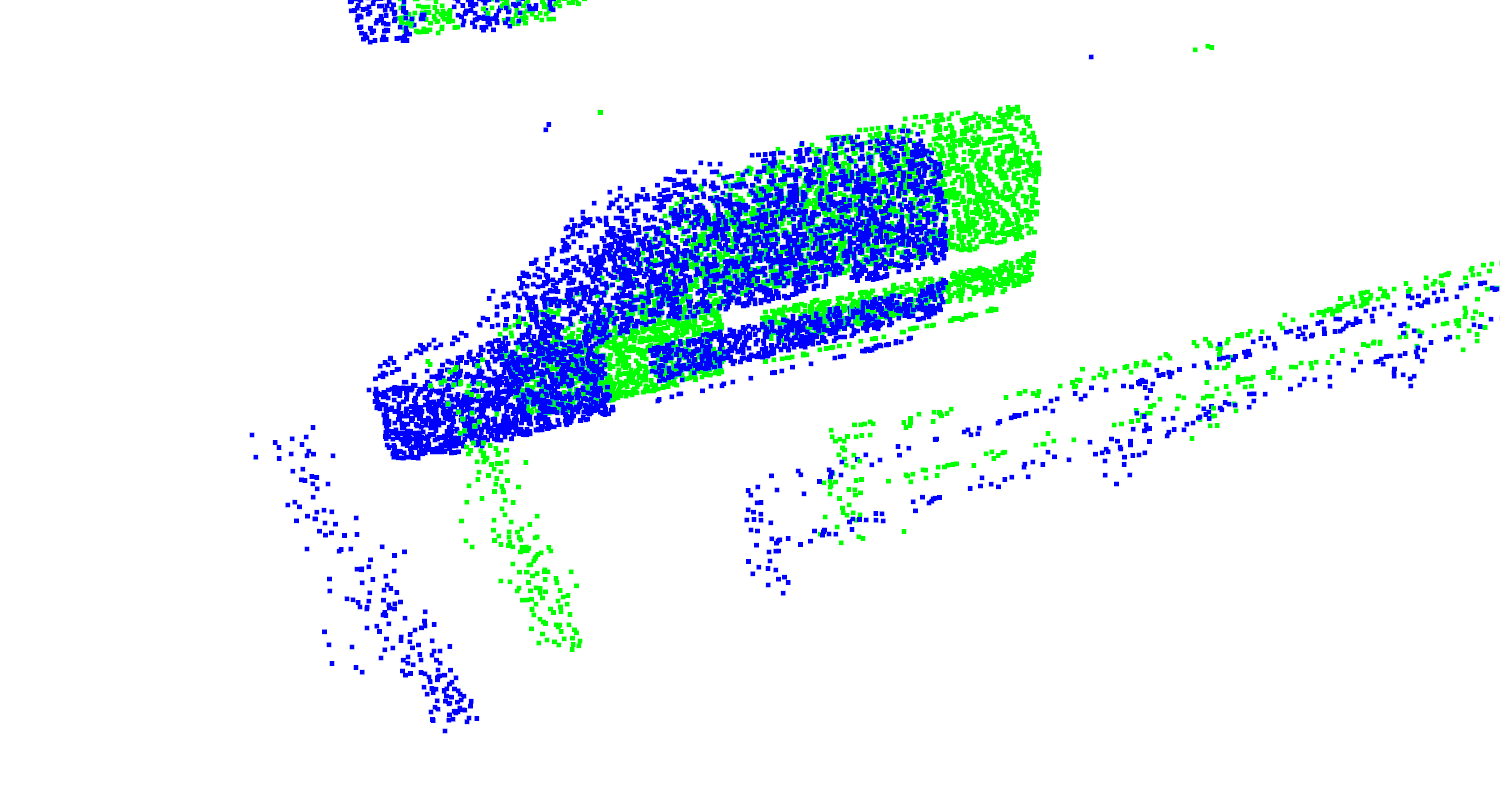}
      \caption*{iter=3}
  \end{subfigure}
  \begin{subfigure}{0.48\columnwidth}
      \centering
      \includegraphics[width=1\columnwidth, trim={0cm 0cm 0cm 1cm}, clip]{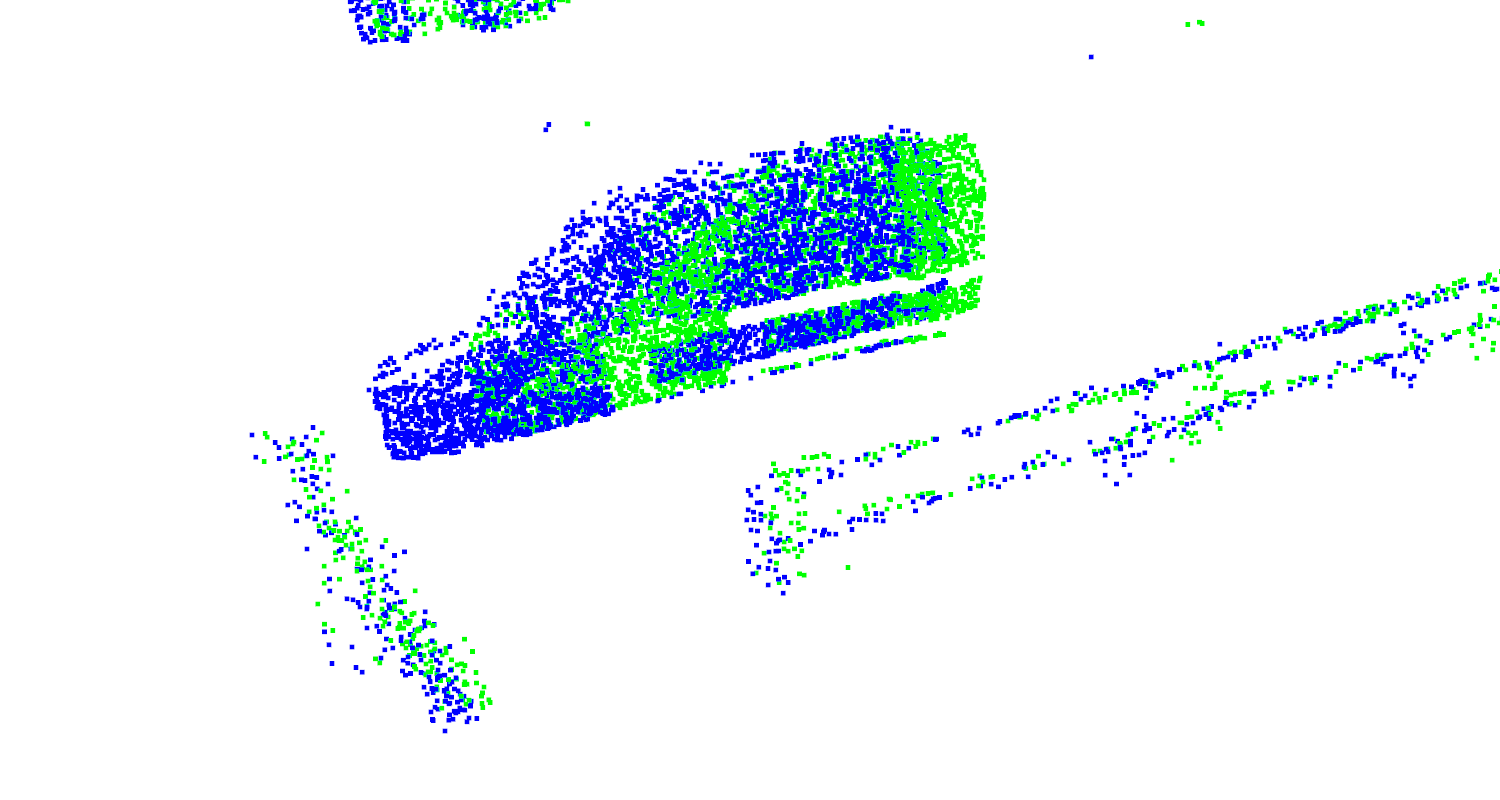}
      \caption*{iter=7}
  \end{subfigure}
  \begin{subfigure}{0.48\columnwidth}
      \centering
      \includegraphics[width=1\columnwidth, trim={0cm 0cm 0cm 1cm}, clip]{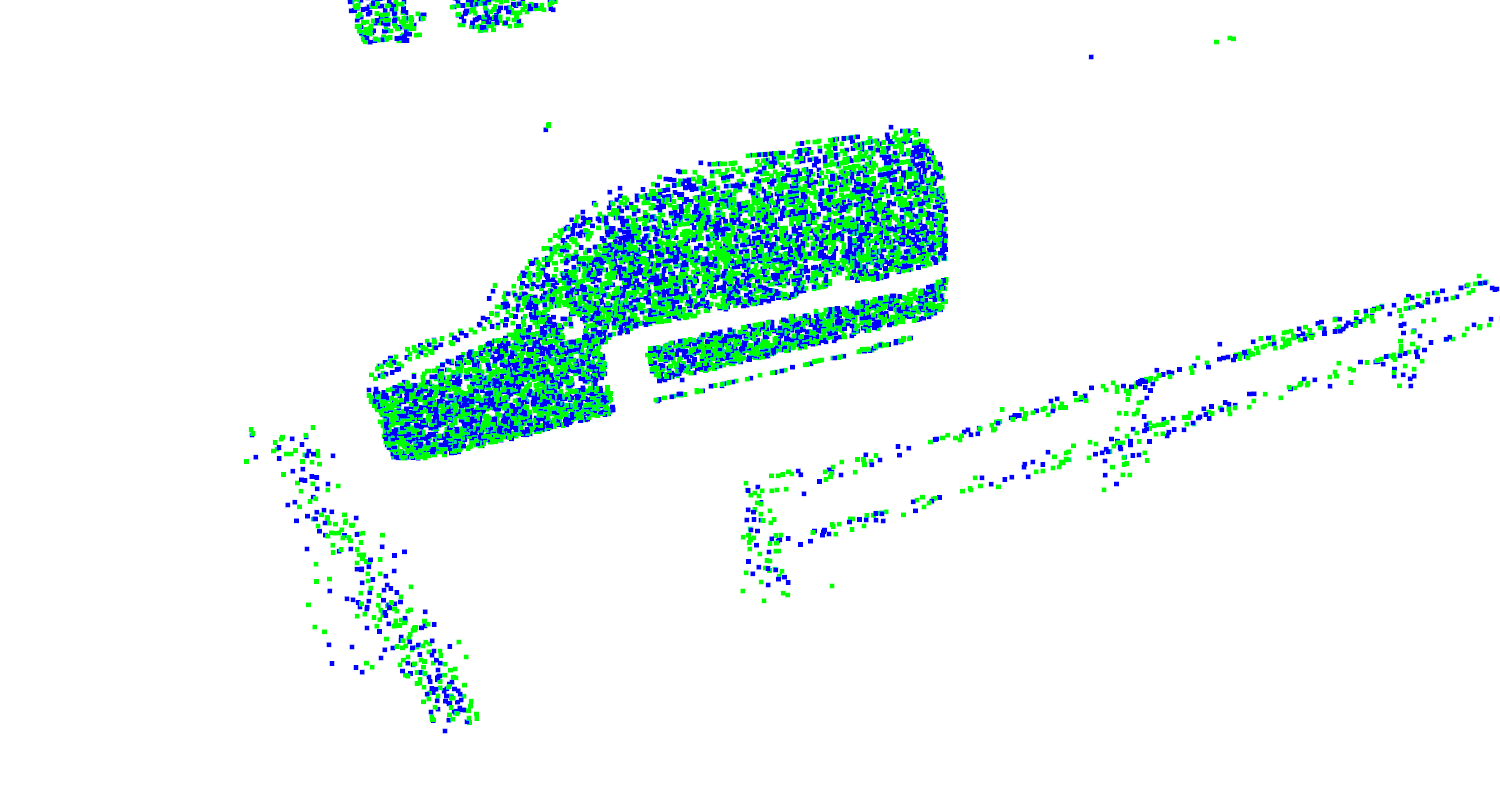}
      \caption*{iter=14}
  \end{subfigure}

  \caption{Visualization of the results on the KITTI scene flow dataset. With the increasing number of alternate optimizations, the source point cloud (in green) is gradually aligned with the target point cloud (in blue).}
  \label{fig:optimization}
  \vspace{-4mm}

\end{figure}

\section{Introduction}

Motion estimation is a fundamental building block for numerous applications such as robotics, augmented reality and autonomous driving. The low-level motion cues can serve other higher-level tasks such as object detection and action recognition. Given a pair or a sequence of images, we can estimate 2D flow fields from optical flow estimation by either classic variational methods or modern deep learning methods~\cite{flownet2, pwcnet, raft}.

Different from the scene flow methods that extend 2D optical flow to stereoscopic or RGB-D image sequences~\cite{Variational, RGBD-sf}, increasing attention has been paid to the direct 3D flow estimation on point clouds recently, which has several advantages over image based methods for a variety of applications. For example, one of the most prominent benefits is that it avoids the image sensor readings and the additional computation of depth from images for autonomous driving, which enables low-latency 3D flow estimation for a high-speed driving vehicle. While for augmented reality, especially for AR glasses, the 3D flow estimation on point clouds enables the computation distribution on the cloud server because it saves much more transmission bandwidth than images. It also protects the privacy of the surrounding people by not using images.

Therefore, some learning based methods~\cite{wang2018deep, flownet3d, HPLFlowNet, pointflowrigid, flowstep3d} utilizes the recent advances made for high-level tasks and customize the scene flow estimation specifically for point clouds. These methods predict the 3D flow vectors from cost-volumes, where similarity costs between 3D points from two point cloud sets are measured. However, different from the cost-volumes in 2D optical flow that search a fixed regular neighborhood around a pixel in consecutive images~\cite{flownet2, pwcnet, raft}, it is impossible to define such a search window on point clouds because of the irregular data structure. Therefore, previous works like ~\cite{flownet3d, pointPWC, flowstep3d,HPLFlowNet}, they designed some complicated layers to measure the point-to-patch cost or patch-to-patch cost. 

In this paper, we avoid this irregularity by a simple and effective method. We decompose the problem into two interlaced stages, where the 3D flows are optimized point-wisely at the first stage and then globally regularized in a recurrent network at the second stage. Therefore, the recurrent network only receives the regular point-wise information as the input.
Besides the scene flow estimation, our method also enables another important motion estimation task that registers two point clouds with the different 6-DOF pose. Since we only measure the point-to-point costs, we avoid the discretization of the 6DOF solution space, which is difficult because the rotation vector and the translation vector are two different variables that have different scales and ranges.

To evaluate the proposed method, we conduct experiments on both the 3D scene flow estimation and the point cloud registration. For 3D scene flow estimation, we made comparisons on the widely used FlyingThings3D~\cite{ft3d} and KITTI~\cite{kitti2015} benchmark. For point cloud registration, we follow the previous works and generate data pairs with large pose and partially overlapping from ModelNet40~\cite{modelnet}. We have achieved state-of-the-art results on both 3D scene flow estimation and point cloud registration, which demonstrate the superiority of the proposed zero-order method on irregular point cloud data.

\section{Related work}
\setlength{\parskip}{0.5em}

\paragraph{Point Cloud Networks. }
Prior to the emergency of neural networks, various hand-crafted 3D feature descriptors have been proposed based on the heuristic knowledge~\cite{featdesc-review}.
These descriptors usually accumulate measurements to histograms based on the spatial coordinates~\cite{3DSC, johnson1999using, USC} or the geometry attributes~\cite{LSP, SHOT}.
Other works such as PFH~\cite{PFH} and FPFH~\cite{FPFH} have been proposed the descriptors which are rotation-invariant.
Traditional features, however, are designed by hand and cannot be too complicated.
More recently, more methods have begun to employ the deep neural networks to learn the features.
3DMatch~\cite{zeng20163dmatch} utilizes a contrastive loss to train a descriptor with 3D convolutional neural networks, but the voxelization leads to a loss of feature quality.
To solve this, PPFNet~\cite{ppfnet} uses the PointNet~\cite{pointnet} to directly learn the point cloud features. Also, the salient keypoints are detected to describe the point clouds~\cite{li2019usip, 3dfeatnet}.
In this work, we use PointNet++~\cite{pointnet++} to extract the point cloud features considering its strong feature representation ability.

\paragraph{3D Flow Estimation on Point Clouds. }
The task of scene flow estimation is first introduced in~\cite{vedula1999three} and then developed from images~\cite{Variational, RGBD-sf} to point clouds~\cite{dewan2016rigid, ushani2017learning, ushani2018feature}.
\cite{dewan2016rigid} formulates the scene flow estimation problem as an energy minimization problem.~\cite{ushani2017learning} constructs occupancy grids and filters the background before the energy minimization and then refine the results with filtering after. An encoding network is introduced in a following up work to learn features from the occupancy grid~\cite{ushani2018feature}.

Recently, more methods use deep neural networks for 3D flow estimation~\cite{wang2018deep, flownet3d, HPLFlowNet, pointflowrigid, flowstep3d, pointPWC}.
A network that learns hierarchical features of point clouds and flow embeddings representing point motions is designed in FlowNet3D~\cite{flownet3d}. 
A bilateral convolutional layer which projects the point cloud to the permutohedral lattice is proposed in HPLFlowNet~\cite{HPLFlowNet}. 
PointPWC-Net~\cite{pointPWC} proposes a learnable cost volume layer along with upsample and warping layers to build a coarse-to-fine deep network to efficiently handle point cloud.
FLOT~\cite{FLOT} uses optimal transport tools to estimate the scene flow without using multiscale analysis.
Recent works~\cite{raft, unrol_img_de, unrol_graph_de, gu2021dro} get promising results by using the model unrolling method,   FlowStep3D~\cite{flowstep3d} adopts GRU\cite{GRU} to iteratively refine the scene flow and show excellent results.
However, all the above methods need to define the cost volume, which is difficult to design considering the irregularity of the point clouds.
In contrast, we avoid this irregularity by optimizing the cost point-wisely and sending the regular point-wise information to a recurrent network.

\begin{figure*}[t]
    \centering

    \includegraphics[width=1.9\columnwidth, trim={0cm 0cm 0cm 0cm}, clip]{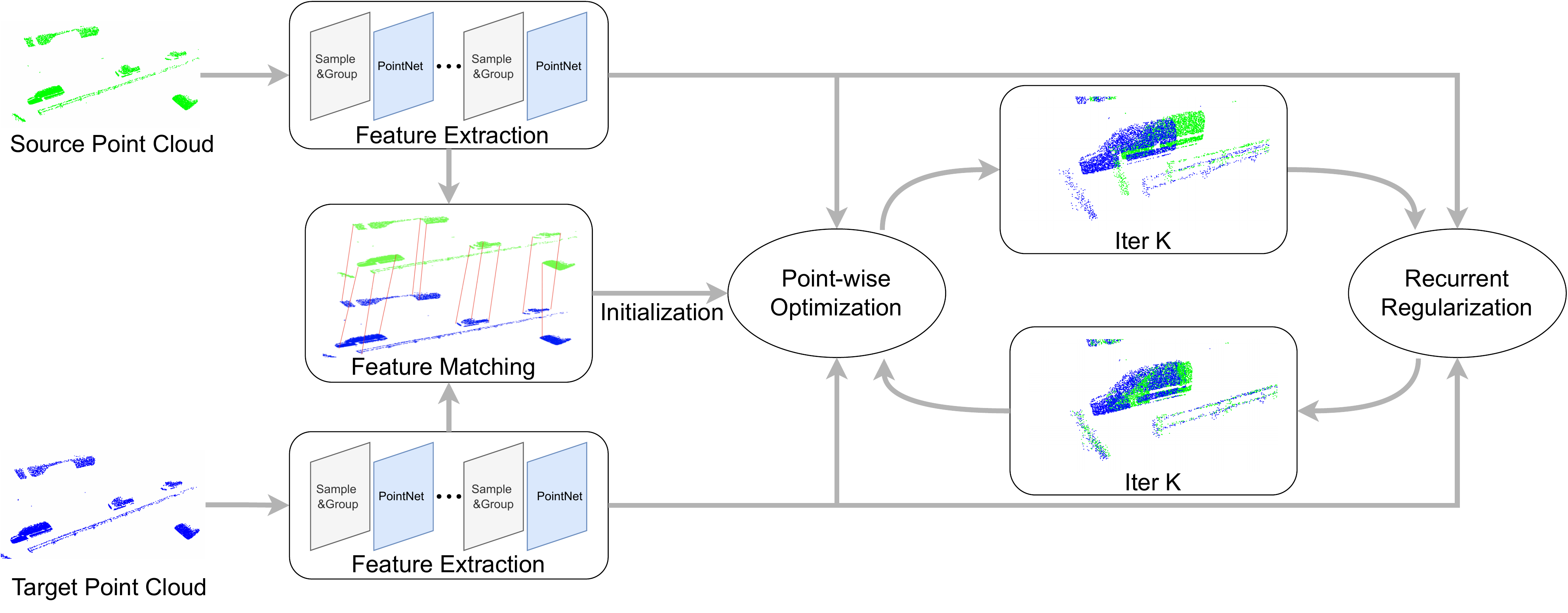}
  
    \caption{Overview of our framework. We first extract the point cloud features by a shared PointNet++~\cite{pointnet++} backbone, followed by the sinkhorn algorithm~\cite{sinkhorn} to get the initialized results from feature matching. Then, we decompose the problem into two sub-problems and utilize the proposed point-wise optimization and recurrent regularization modules to estimate either the 3D scene flow or the transformation.}
    \label{fig:framework}
    \vspace{-4mm}

\end{figure*}

\paragraph{Point Cloud Registration. }
The task of point cloud registration is to find the spatial transformation between two point clouds. Traditional methods are mainly based on the iterative closest point (ICP)\cite{besl1992method,segal2009generalized} and its variants\cite{sinkhorn, kernelcorr, goicp, godin-weights, rusinkiewicz2019symmetric, pomerleau2015review, rusinkiewicz2001efficient}. Recent works learn deep neural networks for point cloud registration and can be divided into two streams. One stream\cite{fgr,chui-featuregmm,rpm,bsc,igsp,pointnetlk,huang2021predator} studies how to leverage better point cloud feature presentation in off-the-shelf optimization. PointNetLK\cite{pointnetlk} defines a feature-metric distance between point clouds, which is measured by the features from PointNet\cite{pointnet}, and unrolls the classical Lucas \& Kanade (LK) algorithm into a recurrent network to minimize this distance.

Another stream\cite{pcrnet,deepcp,deepicp, prnet,fu2021robust, huang2021predator} studies the optimization process itself, which usually makes the optimization algorithm differentiable and incorporate it into a deep learning pipeline. To improve the robustness against noise, PCRNet\cite{pcrnet} proposes a framework which estimates the pose with a deep MLP network instead of the traditional LK algorithm. Deep Closest Point\cite{deepcp} proposes an end-to-end pipeline with a point cloud embedding network, an attention-based matching module, and a differentiable singular value decomposition (SVD) layer to estimate the spatial transformation.
RPM~\cite{rpm} and RPMNet~\cite{rpmnet} use Sinkhorn normalization~\cite{sinkhorn} to handle outliers and partial visibility.
To further decrease the sensibility to outlier points, RGM\cite{fu2021robust} develops a deep graph matching network which considers both the local geometry of each point and its structure and topology in a larger range so that more correspondences can be found.
Inspired by RAFT~\cite{raft}, we use a recurrent network directly to estimate the residual of the transformation to find more high-quality correspondences.

\section{Method}
\subsection{Overview}

Given two point clouds $\mathcal{P}=\{\vp_{1},\vp_{2},\cdots\vp_M\}$ and $\mathcal{Q}=\{\vq_{1},\vq_{2},\cdots\vq_N\}$, where $\mathcal{P}$ and $\mathcal{Q}$ do not necessarily have the same number of points or have any exact correspondence between their points, our goal is to estimate the transformation~$\mathcal{X}$ between them, which can be either a set of point-wise 3D flow vectors $\{\vx_1, \vx_2,\cdots\vx_M\}$ for a dynamic scene or a 6-DOF transformation $\{\mQ,\mT\}$ for a rigid scene, where $\mQ$ is the rotation quaternion and $\mT$ is the 3D translation vector. As shown in~\figref{framework}, we first extract the feature of  each 3D point in $\mathcal{P}$ and $\mathcal{Q}$ by a shared PointNet++~\cite{pointnet++} in~\secref{feature}, and then estimate the initial transformation $\mathcal{X}_{0}$ via the sinkhorn~\cite{FLOT,sarlin2020superglue,cuturi2013sinkhorn,sinkhorn} based feature matching. Then the initial solution $\mathcal{X}^{0}$ is further updated by minimizing the following objective function:
\begin{equation}
E(\mathcal{X})=D_{\mathcal{P},\mathcal{Q}}(\mathcal{X})+R(\mathcal{X}),
\label{equ:raw}
\end{equation}
where the data term $D_{\mathcal{P},\mathcal{Q}}$ measures the feature distance between $\mathcal{P}$ and $\mathcal{Q}$ under a transformation $\mathcal{X}$, and the regularization term $R(\mathcal{X})$ measures the object-aware smoothness of $\mathcal{X}$ itself,~\ie, neighboring points on the same object should have similar flow vectors.

However, optimizing~\equref{raw} directly is difficult, and one of the major reasons is that a transformed point $\vp_{m}+\vx_{m}$ does not exactly correspond to any point $\vq_n$ in $\mathcal{Q}$, so the feature distance between $\vp_{m}$ and $\vp_{m}+\vx_{m}$ is invalid, and the data term $D_{\mathcal{P},\mathcal{Q}}$ is non-differentiable respect to the transformation $\mathcal{X}$.
Inspired by the quadratic relaxation in previous works~\cite{steinbrucker2009large}, we introduce an auxiliary transformation $\mathcal{Z}$ and convert~\equref{raw} into:
\begin{equation}
E(\mathcal{X},\mathcal{Z})=D_{\mathcal{P},\mathcal{Q}}(\mathcal{Z})+R(\mathcal{X})+\|\mathcal{Z}-\mathcal{X}\|^{2}_{2}.
\label{equ:joint}
\end{equation}
The auxiliary variable $\mathcal{Z}$ decouples the data term and the regularization term, so they can be optimized separately. 
Meanwhile, the additional term $\|\mathcal{Z}-\mathcal{X}\|^{2}_{2}$ enforces $\mathcal{Z}$ and $\mathcal{X}$ close to each other through the optimization process. The $\mathcal{Z}$ and the $\mathcal{X}$ are updated interlacedly as:
\begin{numcases}{}
\mathcal{Z}^{k}=\argmin_{\mathcal{Z}} D_{\mathcal{P},\mathcal{Q}}(\mathcal{Z})+\|\mathcal{Z}-\mathcal{X}^{k-1}\|^{2}_{2},\label{equ:data_term}\\
\mathcal{X}^{k}=\argmin_{\mathcal{X}}\|\mathcal{Z}^{k}-\mathcal{X}\|^{2}_{2}+R(\mathcal{X})\label{equ:reg_term}.
\end{numcases}
where $k$ represents the $k$-th iteration. At each iteration, we first solve $\mathcal{Z}^k$ via the point-wise optimization in~\secref{point_wise} and then $\mathcal{X}^k$ via the recurrent regularization implicitly in~\secref{rec_reg}.



\subsection{Feature Extraction}
\label{sec:feature}
First, we extract features that are used through the sinkhorn feature matching~\cite{sinkhorn}, point-wise optimization of~\equref{data_term} as well as the recurrent regularization of~\equref{reg_term}. We use PointNet++~\cite{pointnet++} to extract features for each 3D point, and denote $\vf_{\mathcal{P}}(\cdot)$ and $\vf_{\mathcal{Q}}(\cdot)$ as the feature extraction operation on $\mathcal{P}$ and $\mathcal{Q}$ respectively. PointNet++ ~\cite{pointnet++} designs the $set\_conv$ layer which includes the sample layer, the group layer and the PointNet~\cite{pointnet} layer. The sample layers subsample the point clouds into $1/4$ of the original number by using iterative farthest point sampling. The group layer groups the $k$-nearest neighbor ($k=32$) points around each point and then uses max-pooling for feature aggregation. 

\begin{figure}[t]
  \centering

  \begin{subfigure}{0.48\columnwidth}
      \centering
      \includegraphics[width=1\columnwidth, trim={0cm 0cm 0cm 0cm}, clip]{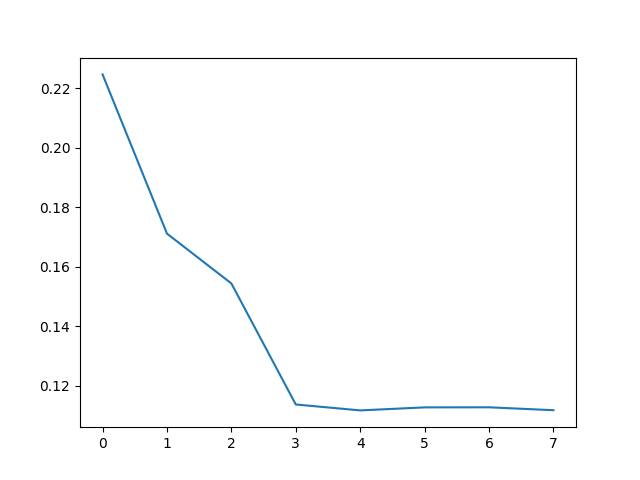}
      \caption*{Cost}
  \end{subfigure}
  \begin{subfigure}{0.48\columnwidth}
      \centering
      \includegraphics[width=1\columnwidth, trim={0cm 0cm 0cm 0cm}, clip]{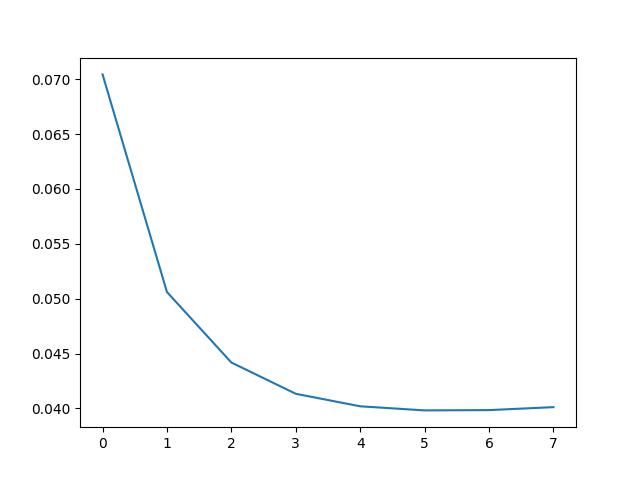}
      \caption*{EPE3D}
  \end{subfigure}

  \caption{Curves of the cost and EPE3D error with respect to the iterations on the Flyingthigs3D dataset. As the alternate optimization progresses, the cost drops step by step, while the EPE3d error is decreased gradually.}
  \label{fig:optimization}
  \vspace{-1mm}

\end{figure}

\subsection{Point-wise Optimization}
\label{sec:point_wise}
Given the solution $\mathcal{X}^{k-1}=\{\vx_{{1}}^{k-1}, \vx_{{2}}^{k-1},\cdots\vx_{M}^{k-1}\}$ from the previous iteration, we first minimize~\equref{data_term} point-wisely to update $\mathcal{Z}^{k}$. The auxiliary transformation $\mathcal{Z}^{k}=\{\vz_{{1}}^{k-1}, \vz_{{2}}^{k-1},\cdots\vz_{M}^{k-1}\}$ contains the auxiliary flow vectors for each point in $\mathcal{P}$. An auxiliary flow vector $\vz_{m}^{k}$ minimizes both the feature distance between $\vf_{\mathcal{P}}(\vp_{m})$ and $\vf_{\mathcal{Q}}(\vp_{m}+\vz^{k}_{m})$ as well as the euclidean distance between $\vz_{m}^{k}$ and $\vx_{m}^{k-1}$:
\begin{equation}
\vz_{m}^{k}=\argmin_{\vz} \|\vf_{\mathcal{P}}(\vp_m)-\vf_{\mathcal{Q}}(\vp_m+\vz)\|+\|\vz-\vx_{m}^{k-1}\|.
\label{equ:point_wise1} 
\end{equation}
However, $\vp_{m}+\vz$ usually does not exactly correspond to any point $\vq_n$ in $\mathcal{Q}$, so we use $\vu_{n}=\vq_n-\vp_m$ as a candidate for $\vz_{m}^{k}$ and search for the best $\vq_n$ within a local neighborhood $\Omega$ around $\vp_{m}+\vx_{m}^{k-1}$. This winer-take-all selection is further softened as a bilateral interpolation:
\begin{equation}
\footnotesize
\vz_{m}^{k}=\frac{1}{W}\sum_{\vq_{n}\in\Omega}\exp(-\frac{\|\vf_{\mathcal{P}}(\vp_m)-\vf_{\mathcal{Q}}(\vq_{n})\|}{\sigma_{\vf}}-\frac{\|\vu_{n}-\vx_{m}^{k-1}\|}{\sigma_{\vu}})\vu_{n},
\label{equ:point_wise2} 
\end{equation}
where 
$W$ is the normalization term that sum all the bilateral weights.
Empirically, the reversed distance $-(\|\vf_{\mathcal{P}}(\vp_m)-\vf_{\mathcal{Q}}(\vq_{n})\|)/\sigma_{\vf})$ and $-(\|\vu_{n}-\vx_{m}^{k-1}\|)/\sigma_{\vu}$ can be further replaced with cosine similarity for better performance, which converts~\equref{point_wise2} into:
\begin{equation}
\vz_{m}^{k}=\sum_{\vq_{n}\in\Omega}\softmax(\vg^{\top}_{m}\vg_{n})\vu_{n},
\label{equ:point_wise3}
\end{equation}
where $\vg_{m}$ concatenates $\vf_{\mathcal{P}}(\vp_{m})$ and the positional encoding of $\vp_{m}+\vx_{m}^{k-1}$, while $\vg_{n}$ concatenates $\vf_{\mathcal{Q}}(\vq_{n})$ and the positional encoding of $\vq_{n}$.~\equref{point_wise3} is similar to the attention
~\cite{attention} mechanism, but uses the same feature operator for the key and the queries.

For point cloud registration, two additional steps are required.
First, before~\equref{point_wise3}, the point-wise flow vector $\vx_{m}^{k-1}$ is calculated by the difference between $\vp_{m}$ and its corresponding projection transformed by $\mathcal{X}^{k-1}=\{\mQ^{k-1},\mT^{k-1}\}$. Second, after $\vz_{m}^{k}$ is estimated, the auxiliary 6-DOF transformation $\mathcal{Z}^{k}$ is computed via the PnP algorithm~\cite{pnp}~from the auxiliary 3D flow vectors $\vz_{1\cdots M}^{k}$.

\begin{figure}[t]
    \centering
    
    \includegraphics[width=0.98\columnwidth, trim={0cm 0cm 0cm 0cm}, clip]{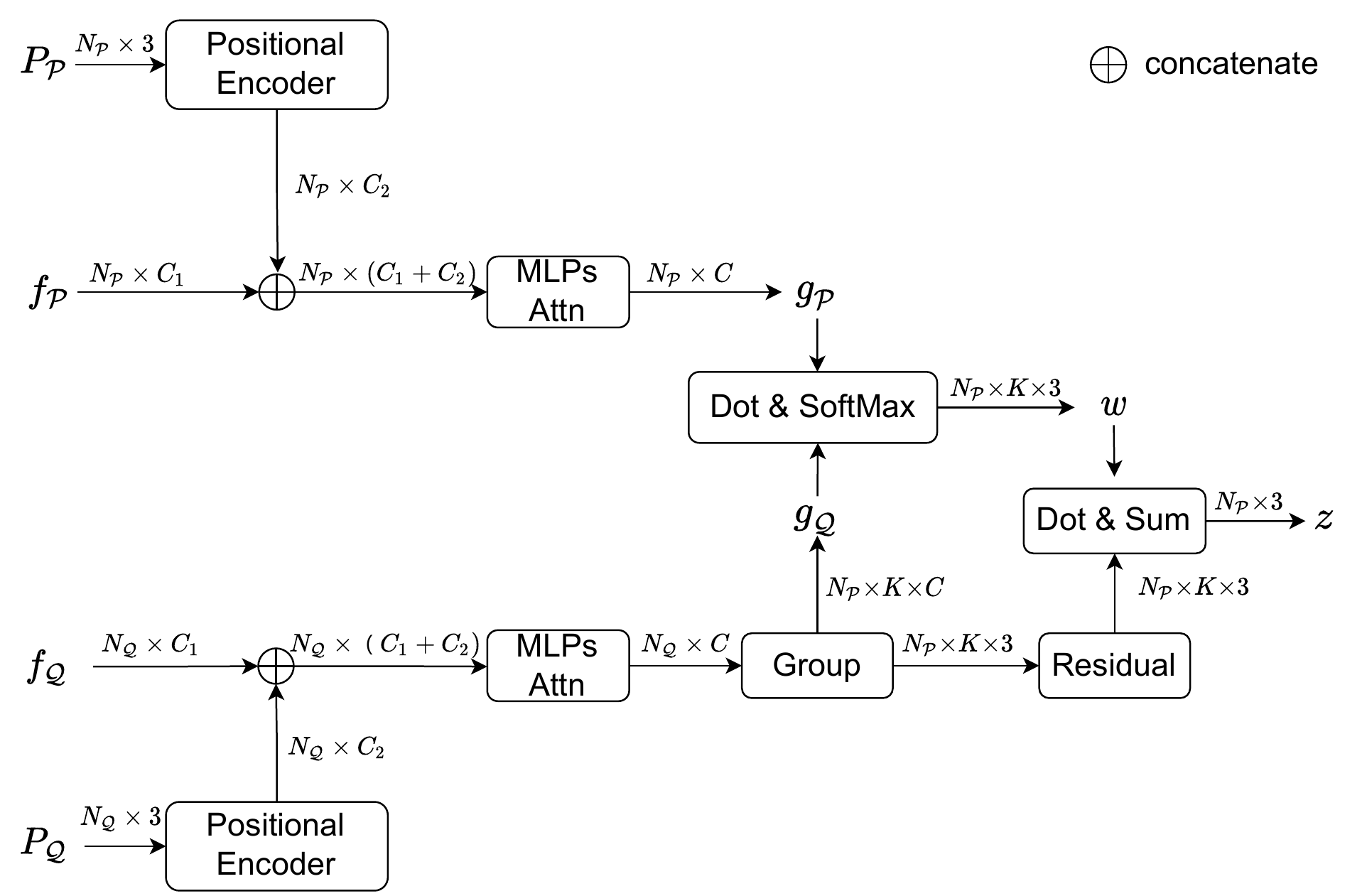}
    
    \caption{Details of point-wise optimization module. The positional encoder and the MLPs-attention blocks are both composed of two fully-connected layers. The parameters of these two blocks for point cloud $\mathcal{P}$ and $\mathcal{Q}$ are shared. The residual block is to obtain the position residual of currently estimated points and their neighborhood points.}
    \label{fig:pointwise}
     \vspace{-5mm}
\end{figure}

\subsection{Recurrent Regularization}
\label{sec:rec_reg}
Given the auxiliary transformation $\mathcal{Z}^{k}$ from the last section, we further implicitly regularize $\mathcal{Z}^{k}$ into $\mathcal{X}^{k}$ by a recurrent network. The recurrent network receives previous iteration's hidden state $\vh_{k-1}$ and the current iteration information $\vv_{k}=[\vf_{\mathcal{P}}(\vp_m)-\vf_{\mathcal{Q}}(\vp_m+\vx_{m}^{k-1}),\vz_{m}^{k}]$ as the inputs, where $\vf_{\mathcal{Q}}(\vp_m+\vx_{m}^{k-1})$ is interpolated similarly to~\equref{point_wise3}. Then an updated hidden state $\vh_k$ is produced from a recurrent unit such as GRU~\cite{GRU} as follows:
\vspace{-2mm}
\begin{equation}
\begin{aligned}
\vw_k &= \sigma(set\_conv_\vw([\vh_{k-1}, \vv_{k}])),\\
\vr_k &= \sigma(set\_conv_\vr([\vh_{k-1}, \vv_{k}])),\\
\tilde{\vh}_k &= \tanh(set\_conv_\vh([\vr_k\odot \vh_{k-1}, \vv_{k}])),\\
\vh_k &= (1-\vw_k)\odot \vh_{k-1} + \vw_k\odot \tilde{\vh}_k,\\
\label{equ:recurrent}
\end{aligned}
\vspace{-5mm}
\end{equation}
where $\odot$ is the element-wise product, $[\cdot,\cdot]$ is a concatenation and $\sigma(\cdot)$ is the sigmoid activation function. To initialize the first iteration's hidden state $\vh_0$, we set $\mathcal{Z}^{1}=\mathcal{X}_{0}$ and pass $\vv_{1}$ through two $set\_conv$ layers.

After the new hidden state $\vh_k$ has been produced, we use a transformation predictor consisting of two $set\_conv$ layers to estimate the residual $\Delta\mathcal{Z}^k$ between $\mathcal{Z}^k$ and the regularized transformation $\mathcal{X}^k$, and update the transformation as $\mathcal{X}^k = \mathcal{Z}^k+\Delta\mathcal{Z}^k$. For point cloud registration, we additionally max-pool the hidden state $\vh_k$ overall points, and predict the residual 6-DOF transformation $\Delta\mathcal{Z}^k$ from the max-pooled feature vector, and update the transformation in $SE(3)$ manifold as $\mathcal{X}^k = \Delta\mathcal{Z}^k\otimes\mathcal{Z}^k$.

\section{Training Loss}
Similar to previous methods~\cite{pointPWC, flowstep3d}, our networks can be trained in either a supervised way or a self-supervised way.
\subsection{Supervised Training}
In the training of the scene flow network, given the ground truth 3D flow vectors$\{\vx^*_{\vp_{i}}\}$ at each point $\vp_{i}$, the $L_{1}$ loss is adopted as:
\begin{equation}
    \mathcal{L}_{flow} = \frac{1}M\sum_i 
            \| \hat{\vx}_{\vp_{i}} - \vx^*_{\vp_{i}}\|,
\end{equation}
where $\{ \hat{\vx}_{\vp_{i}} \}$ is the predicted flow from the recurrent regularization in the last iteration.

In the training of the point cloud registration network, we first transform the source point cloud $\mathcal{P}$ with the ground-truth transformation $\{\mQ^*,\mT^*\}$ and the predicted transformation $\hat{\mQ}, \hat{\mT}$ respectively, and then compute the difference between the transformed point clouds as:
\begin{equation}
    \mathcal{L}_{register} = \frac{1}M \sum_i  
            \|(\hat{\mQ} \cdot \vp_i + \hat{\mT}) - 
              (\mQ^* \cdot \vp_i + \mT^*) \|.
\end{equation}

\subsection{Self-supervised Training}
The ground truth 3D flow is expensive, so the network needs to utilize the geometric priors as supervision when there is no access to the ground truth. Following previous work~\cite{pointPWC}, our self-supervised loss for scene flow is composed of three terms: Chamfer distance, Smoothness regularization, and Laplacian regularization~\cite{pixel2mesh, laplacian_mesh}.

\paragraph{Chamfer Term }
The Chamfer distance enforces the source point cloud $\mathcal{P}$ to move close to the target point cloud $\mathcal{Q}$ with the mutual closest points as: 
\begin{equation}
\begin{aligned}
    \mathcal{L}_{C}
    =&\sum_{\vp_i\in \mathcal{P}'}\min_{\vq_j\in \mathcal{Q}}\|\vp_i-\vq_j\|_2^2 +\sum_{\vq_j\in \mathcal{Q}}\min_{\vp_i\in \mathcal{P}'}\|\vq_j-\vp_i\|_2^2,
\end{aligned}
\end{equation}
where $\mathcal{P}'$ is transformed from $\mathcal{P}$ by the predicted 3D flow vectors.

\paragraph{Smoothness Term }
The smoothness constraint encourages the adjacent points to have similar 3D flow predictions as: 
\begin{equation}
    \mathcal{L}_{S}=\sum_i \frac{1}{|L(\vp_i)|}
            \sum_{\vp_j \in L(\vp_i)} 
            || \hat{\vx}_{\vp_j} -  \hat{\vx}_{\vp_i}||_2^2,
\end{equation}
where $L(\vp_i)$ is the local neighbor region of $\vp_i$, and $|L(\vp_i)|$ it the number of points in this local region.

\paragraph{Laplacian Term }
The Laplacian coordinate vector approximates the local shape and curvature around a point $\vp_{i}$ as: 
\begin{equation}
    \delta_i = \sum_i \frac{1}{|L(\vp_i)|}
            \sum_{\vp_j \in L(\vp_i)} 
            (\vp_j -  \vp_i).
\end{equation}
Ideally, the transformed point cloud $\mathcal{P}'$ should have the same Laplacian coordinate with the target point cloud $\mathcal{Q}$. So we adopt a regularization term based on the Laplacian coordinate similar to~\cite{pointPWC}, as:
\begin{equation}
    \mathcal{L}_{R} = \sum_{\vp_i' \in \mathcal{P}'} 
    \|\delta(\vp_i') - \delta (\vq_{inter})\|_2^2 ,
\end{equation}
where $\delta (\vq_{inter})$ is the interpolated Laplacian coordinate from point cloud $Q$ at the same position as $\vp_i'$.

In summary, the self-supervised loss for the scene flow is defined as the weighted sum of these three terms, as:
\begin{equation}
    \mathcal{L}_{selflow} = \alpha_1 \mathcal{L}_C + \alpha_2 \mathcal{L}_S + \alpha_3 \mathcal{L}_R.
\end{equation}

\setlength{\parskip}{0.5em}

\begin{table*}[t]
\centering
\begin{tabular}{c | c | c | c c c c}
\toprule
Datasets & Method & Sup. & EPE3D$\downarrow$ & Acc3DS$\uparrow$ & AccDR$\uparrow$ & Outliers3D$\downarrow$ \\
\midrule
\multirow{10}{*}{FlyingThings3D}

& ICP\cite{icp} & Self & $0.4062$  & $0.1614$ & $0.3038$ & $0.8796$ \\
& Ego-motion\cite{Ego} & Self & $0.1696$  & $0.2532$ & $0.5501$ & $0.8046$ \\
& PointPWC-Net\cite{pointPWC} & Self & $0.1246$  & $0.3068$ & $0.6552$ & $0.7032$ \\
& FlowStep3D\cite{flowstep3d} & Self & $0.0852$  & $0.5363$ & $0.8262$ & $0.4198$ \\
& Ours & Self & $\mathbf{0.0765}$ & $\mathbf{0.5858}$ & $\mathbf{0.8602}$  & $\mathbf{0.4142}$ \\
\cmidrule(l{0em}r{0em}){2-7}

& FlowNet3D\cite{flownet3d} & Full & $0.1136$ & $0.4125$ & $0.7706$ & $0.6016$ \\
& HPLFlowNet\cite{HPLFlowNet} & Full & $0.0804$  & $0.6144$ & $0.8555$ & $0.4287$ \\ 
& PointPWC-Net\cite{pointPWC} & Full & $0.0588$  & $0.7379$ & $0.9276$ & $0.3424$ \\
& FLOT\cite{FLOT} & Full & $0.0520$  & $0.7320$ & $0.9270$ & $0.3570$ \\
& FlowStep3D~\cite{flowstep3d} & Full & $0.0455$  & $0.8162$ & $0.9614$ & $0.2165$ \\
& Ours & Full & $\mathbf{0.0403}$  & $\mathbf{0.8567}$ & $\mathbf{0.9635}$ & $\mathbf{0.1976}$ \\
\midrule

\multirow{10}{*}{KITTI}
& ICP\cite{icp} & Self & $0.5181$  & $0.0669$ & $0.1667$ & $0.8712$ \\
& Ego-motion\cite{Ego} & Self & $0.4154$  & $0.2209$ & $0.3721$ & $0.8096$ \\
& PointPWC-Net\cite{pointPWC} & Self & $0.2549$  & $0.2379$ & $0.4957$ & $0.6863$ \\
& FlowStep3D\cite{flowstep3d} & Self & $0.1021$  & $0.7080$ & $0.8394$ & $0.2456$ \\
& Ours & Self & $\mathbf{0.0763}$  & $\mathbf{0.7856}$ & $\mathbf{0.8921}$ & $\mathbf{0.1849}$ \\

\cmidrule(l{0em}r{0em}){2-7}
& FlowNet3D\cite{flownet3d} & Full & $0.1767$ & $0.3738$ & $0.6677$ & $0.5271$ \\
& HPLFlowNet\cite{HPLFlowNet} & Full & $0.1169$  & $0.4783$ & $0.7776$ & $0.4103$ \\
& PointPWC-Net\cite{pointPWC} & Full & $0.0694$  & $0.7281$ & $0.8884$ & $0.2648$ \\
& FLOT\cite{FLOT} & Full & $0.0560$  & $0.7550$ & $0.9080$ & $0.2420$ \\
& FlowStep3D~\cite{flowstep3d} & Full & $0.0546$  & $0.8051$ & $0.9254$ & $0.1492$ \\
& Ours & Full & $\mathbf{0.0481}$ & $\mathbf{0.8491}$ & $\mathbf{0.9448}$ & $\mathbf{0.1228}$ \\

\bottomrule
\end{tabular}
\caption{Evaluation results on the FlyingThings3D~\cite{ft3d} and KITTI~\cite{kitti2015} datasets. ``Self" denotes training under the self-supervised setting while ``Full" denotes under the supervised setting. All methods are only trained on FlyingThings3D~\cite{ft3d}. In default setting, we set the iteration number to 7 and 14 for FlyingThings3D~\cite{ft3d} and KITTI~\cite{kitti2015}, respectively.}
\label{tab:sceneflow}
\vspace{-5mm}
\end{table*}

\section{Experiments}

\subsection{3D Flow Estimation}

\paragraph{Dataset}
To make a fair comparison, we follow previous works~\cite{flownet3d, pointPWC, FLOT, flowstep3d} to evaluate all participating methods on FlyingThings3D~\cite{ft3d} and KITTI~\cite{kitti2015} for 3D flow estimation. The FlyingThings3D is a synthetic dataset that is rendered from scenes with multiple randomly sampled moving objects from the ShapeNet\cite{chang2015shapenet} dataset. It contains around 32k stereo images with ground truth disparity and optical flow maps. To use it for 3D flow estimation, we post-process it as in~\cite{pointPWC} into 19640 training pairs and 3824 testing pairs of point clouds, and each point cloud contains 8192 3D points on average. Similarly, the KITTI scene flow dataset~\cite{kitti2015} is also originally designed to evaluate the image based methods. We follow~\cite{pointPWC} and post-process it into 142 testing pairs~\emph{only} for evaluation.

\paragraph{Training Details}
We train and evaluate our method with both the self-supervised and the supervised setting as described in Sec.4. To speed up the training, we adopt a two-stage training strategy: first, we set the batch size to 8 on 4 GTX 2080Ti GPUs, and alternate between the~\equref{point_wise3} and the~\equref{recurrent} for 3 times. 
The learning rate adopts the step decay strategy, where the initial learning rate is set to 1e-3 and then halved every 25 epochs, and 90 epochs were trained in total; second, we fine-tune the trained model from the first stage, but iterate more times and train for fewer epochs.
We reduce the batch size to 4 to enable 7 iterations of~\equref{point_wise3} and~\equref{recurrent}.
The initial learning rate is also reduced to 1.25e-4 and then decayed every 5 epochs, and the model is trained for 10 epochs.

\paragraph{Quantitative Comparisons}
We follow previous works~\cite{flownet3d, pointPWC, FLOT, flowstep3d} to use the following metrics for evaluation:
\begin{enumerate}
    \item[$\bullet$] EPE3D (m): $\sum||F_{pred} - F_{gt}||_2/N$ average distance error of all predicted values
    \item[$\bullet$] Acc3DS: the percentage of points which statisfied $||F_{pred} - F_{gt}||_2 < 0.05m$ or $||F_{pred} - F_{gt}||_2/F_{gt} < 5\%$
    \item[$\bullet$] Acc3DR: the percentage of points which statisfied $||F_{pred} - F_{gt}||_2 < 0.1m$ or $||F_{pred} - F_{gt}||_2/F_{gt} < 10\%$
    \item[$\bullet$] Outliers3D: the percentage of points which statisfied $||F_{pred} - F_{gt}||_2 > 0.3m$ or $||F_{pred} - F_{gt}||_2/F_{gt} > 10\%$.
\end{enumerate}
As shown in~\tabref{sceneflow}, we have achieved state-of-the-art results for both self-supervised and supervised setting. Specifically, we performs better than the other methods by a large margin on the KITTI dataset, which demonstrates the superiority of our formulation in~\equref{point_wise3}.


\begin{table}[t]
\centering
\resizebox{\linewidth}{!}{
\begin{tabular}{c | c  c  c  c}
\toprule
Method & MAE(R) & MAE(t)& Error(R) &Error(t)\\
\midrule
ICP\cite{icp} & $13.719$ & $0.132$ & $27.250$ & $0.280$ \\
RPM\cite{rpm} & $9.771$ & $0.092$ & $19.551$ & $0.212$ \\
FGR\cite{fgr} & $19.266$ & $0.090$ & $30.839$ & $0.192$ \\
PointNetLK\cite{pointnetlk} & $15.931$ & $0.142$ & $29.725$ & $0.297$ \\
DCP-v2\cite{deepcp} & $6.380$ & $0.083$ & $12.607$ & $0.169$ \\
TEASER++\cite{teaser} & $4.138$ & $0.020$ & $7.144$ & $0.041$ \\
RPMNet\cite{rpmnet} & $0.893$ & $0.0087$ & $1.712$ & $0.018$ \\
\midrule
Ours & $\mathbf{0.845}$ & $\mathbf{0.0077}$ & $\mathbf{1.665}$ & $\mathbf{0.016}$ \\

\bottomrule
\end{tabular}}
\caption{Evaluation results on ModelNet40~\cite{modelnet} on partially visible setting with Gaussian noise.}
\label{tab:icp_paritial_reg}
\vspace{-5mm}
\end{table}

\begin{figure*}[t]
    \centering
    
    \rule{0.95\textwidth}{0.1pt}
    \vspace{1mm}
 
    \begin{subfigure}{0.48\columnwidth}
        \centering
        \includegraphics[width=1\columnwidth, trim={0cm 0cm 0cm 1cm}, clip]{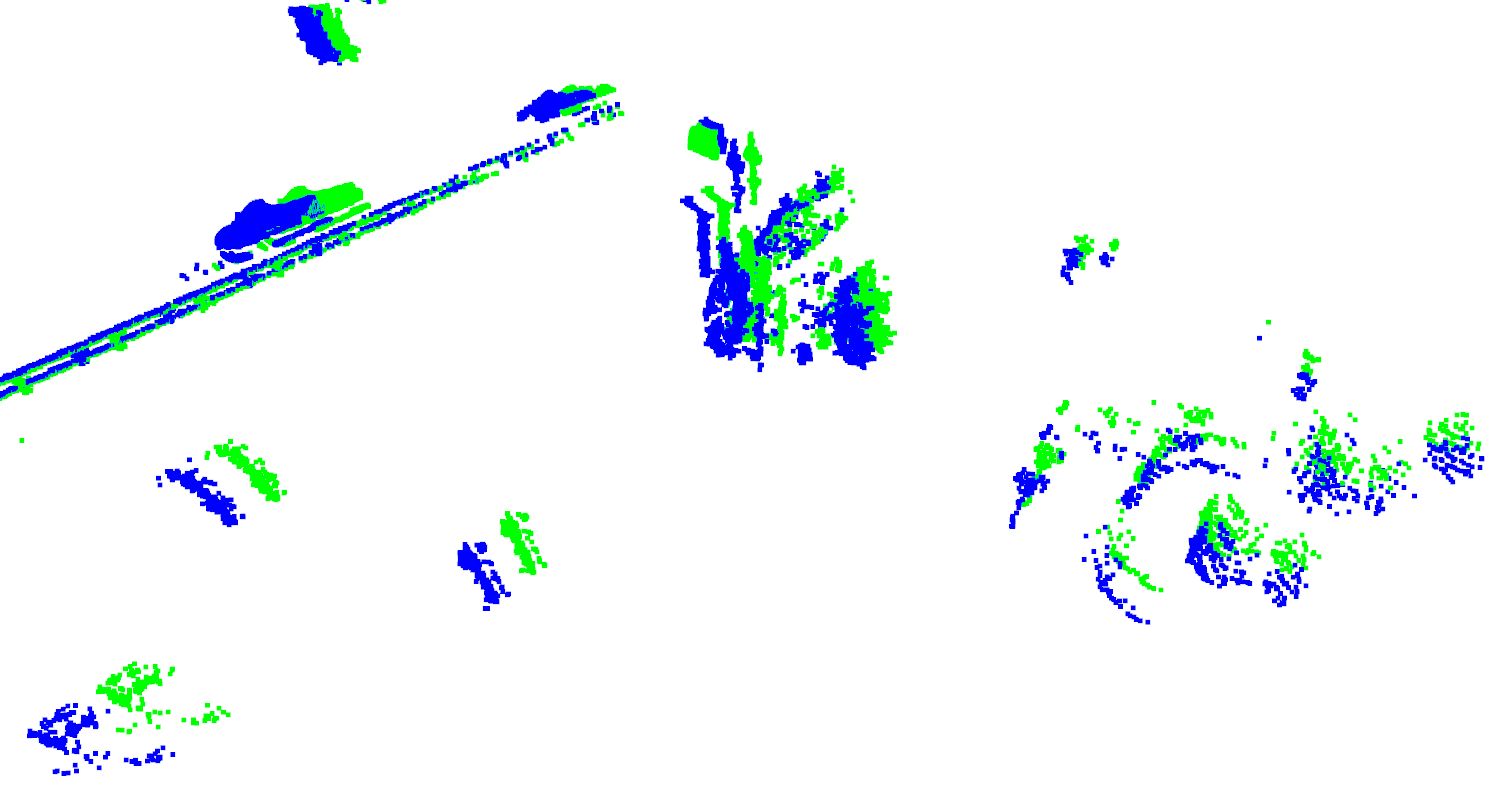}
    \end{subfigure}
    \begin{subfigure}{0.48\columnwidth}
        \centering
        \includegraphics[width=1\columnwidth, trim={0cm 0cm 0cm 1cm}, clip]{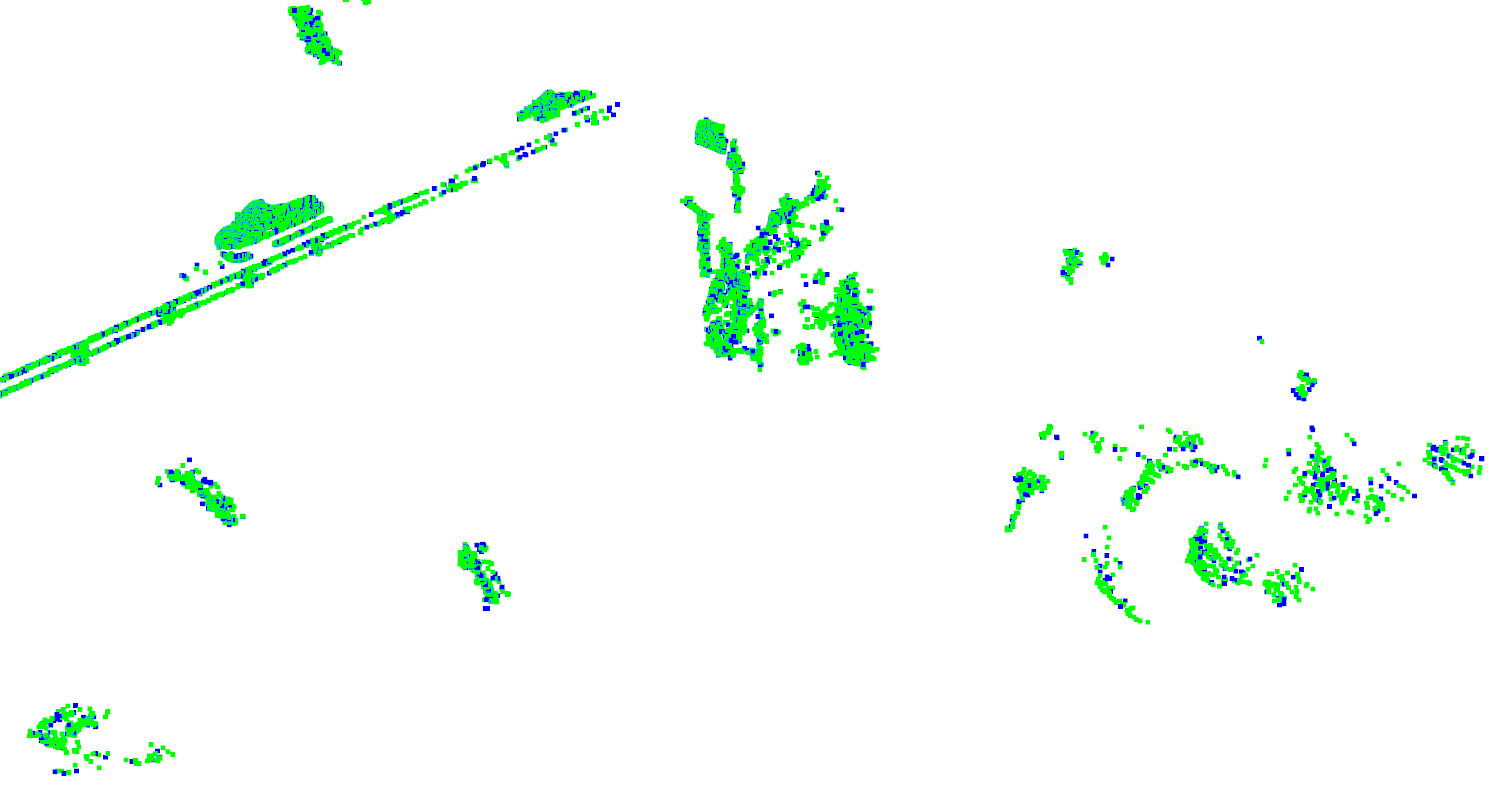}
    \end{subfigure}
    \begin{subfigure}{0.48\columnwidth}
        \centering
        \includegraphics[width=1\columnwidth, trim={0cm 0cm 0cm 1cm}, clip]{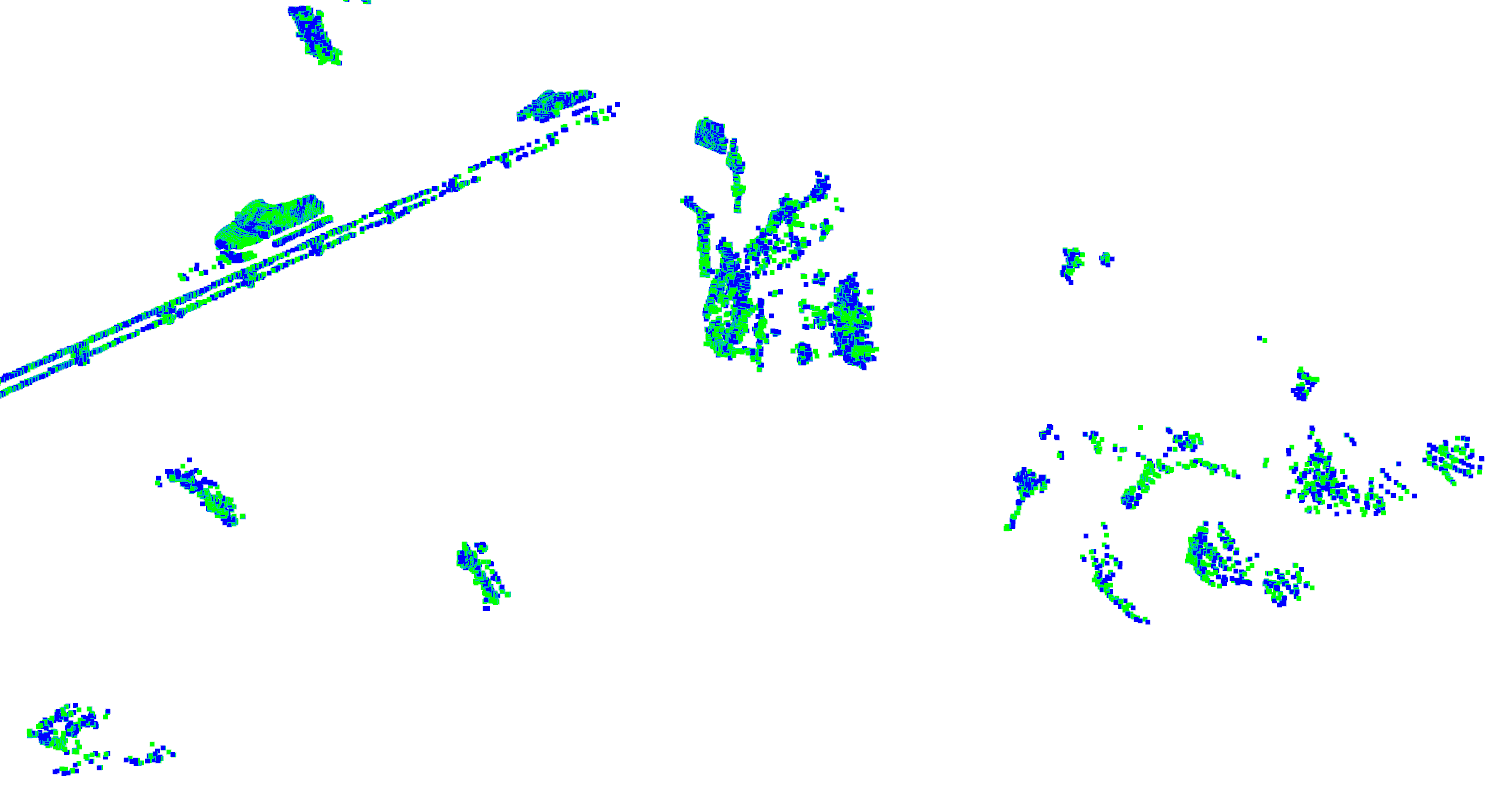}
    \end{subfigure}
    \begin{subfigure}{0.48\columnwidth}
        \centering
        \includegraphics[width=1\columnwidth, trim={0cm 0cm 0cm 1cm}, clip]{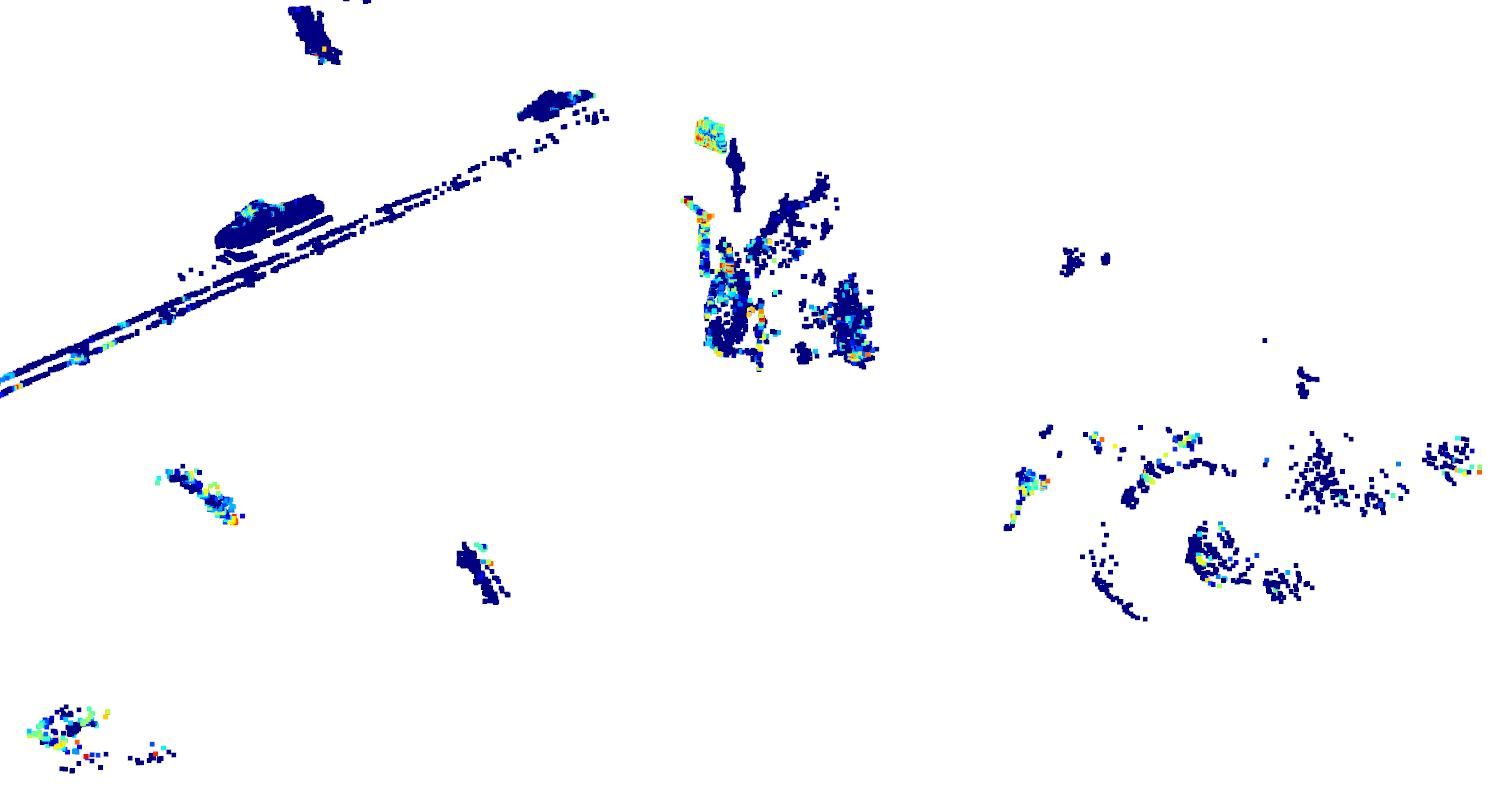}
    \end{subfigure}
    
    \vspace{1mm}
    \rule{0.95\textwidth}{0.1pt}
    \vspace{1mm}

    \begin{subfigure}{0.48\columnwidth}
        \centering
        \includegraphics[width=1\columnwidth, trim={0cm 0cm 0cm 1cm}, clip]{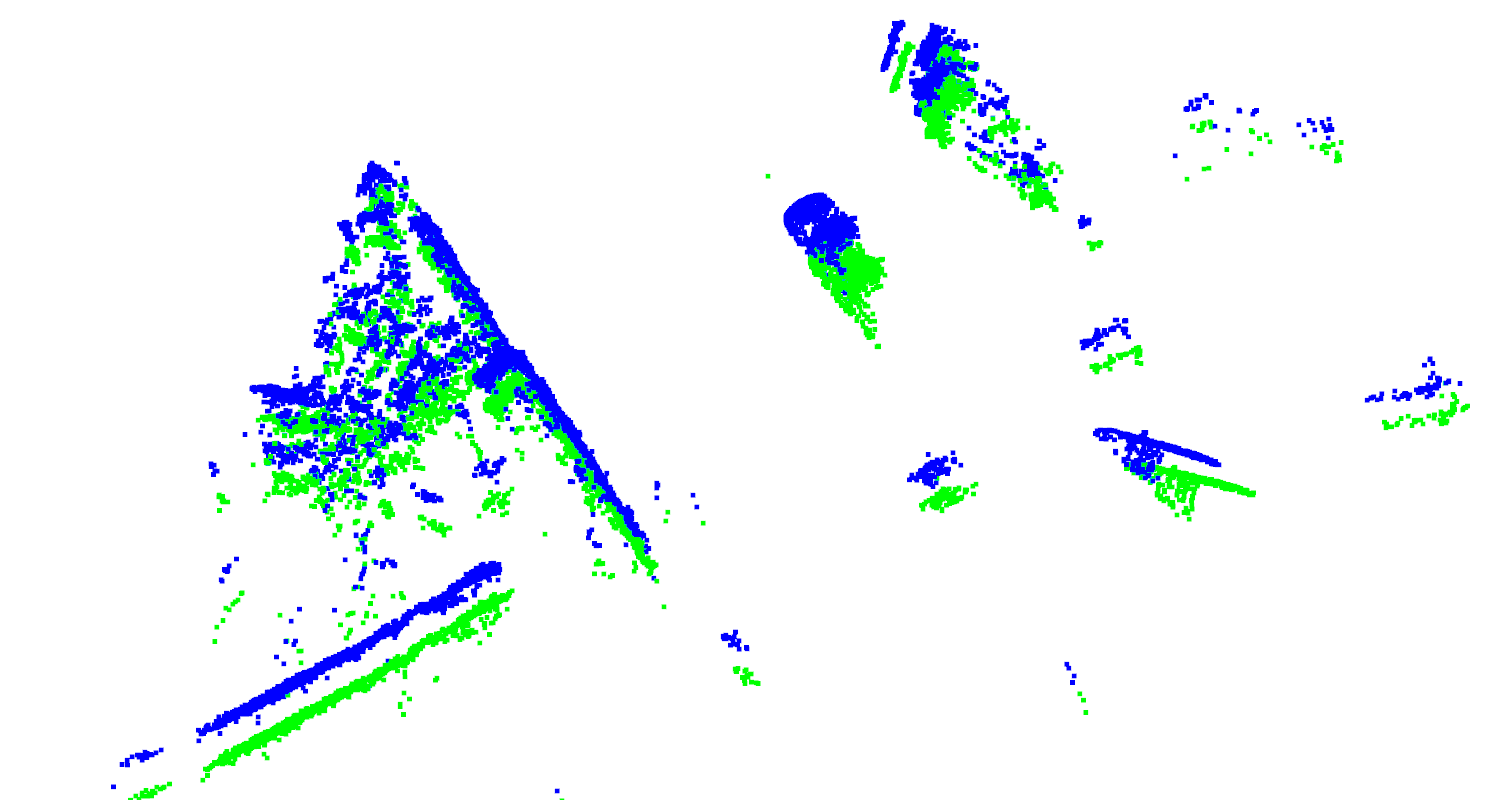}
    \end{subfigure}
    \begin{subfigure}{0.48\columnwidth}
        \centering
        \includegraphics[width=1\columnwidth, trim={0cm 0cm 0cm 1cm}, clip]{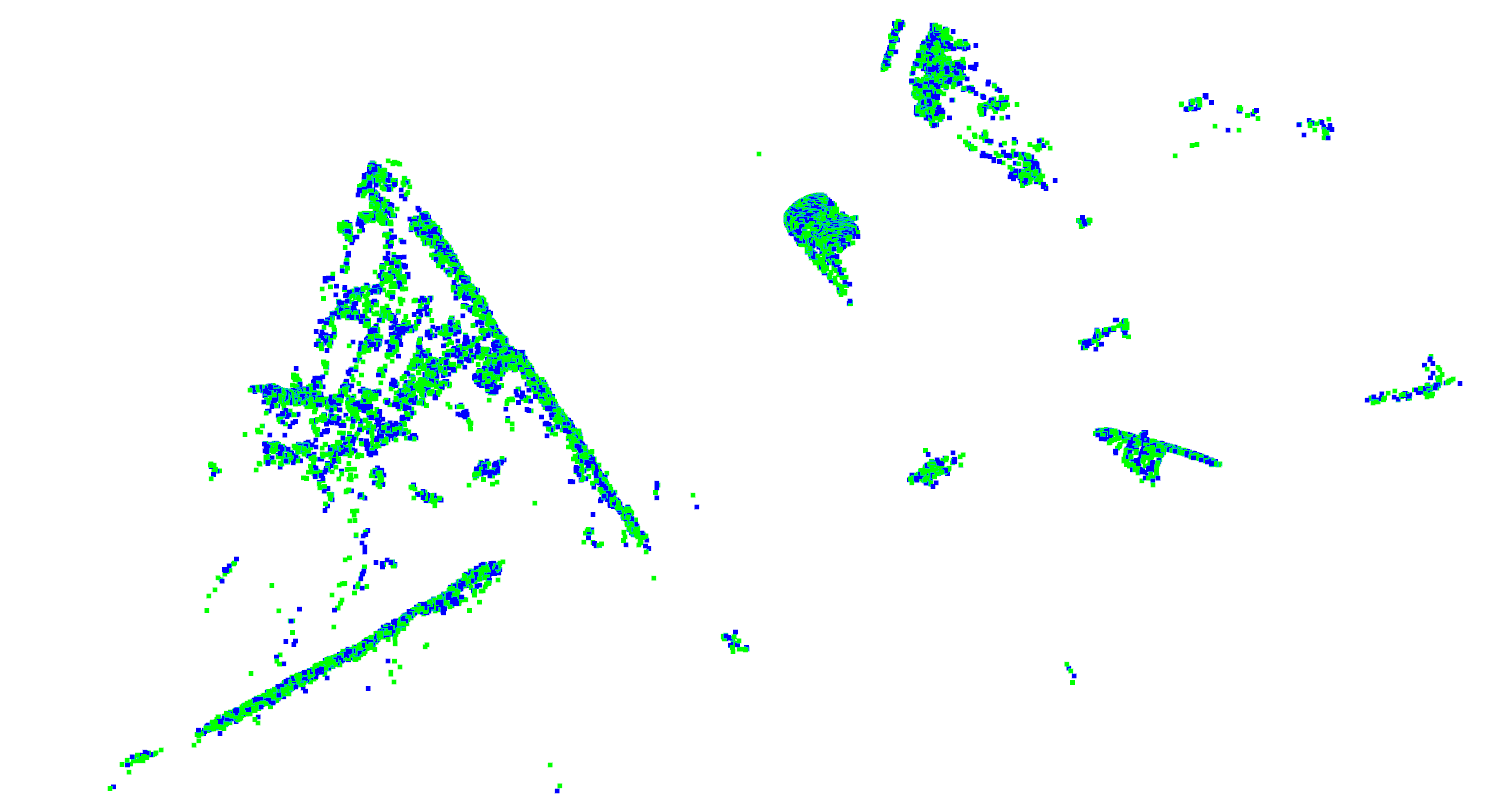}
    \end{subfigure}
    \begin{subfigure}{0.48\columnwidth}
        \centering
        \includegraphics[width=1\columnwidth, trim={0cm 0cm 0cm 1cm}, clip]{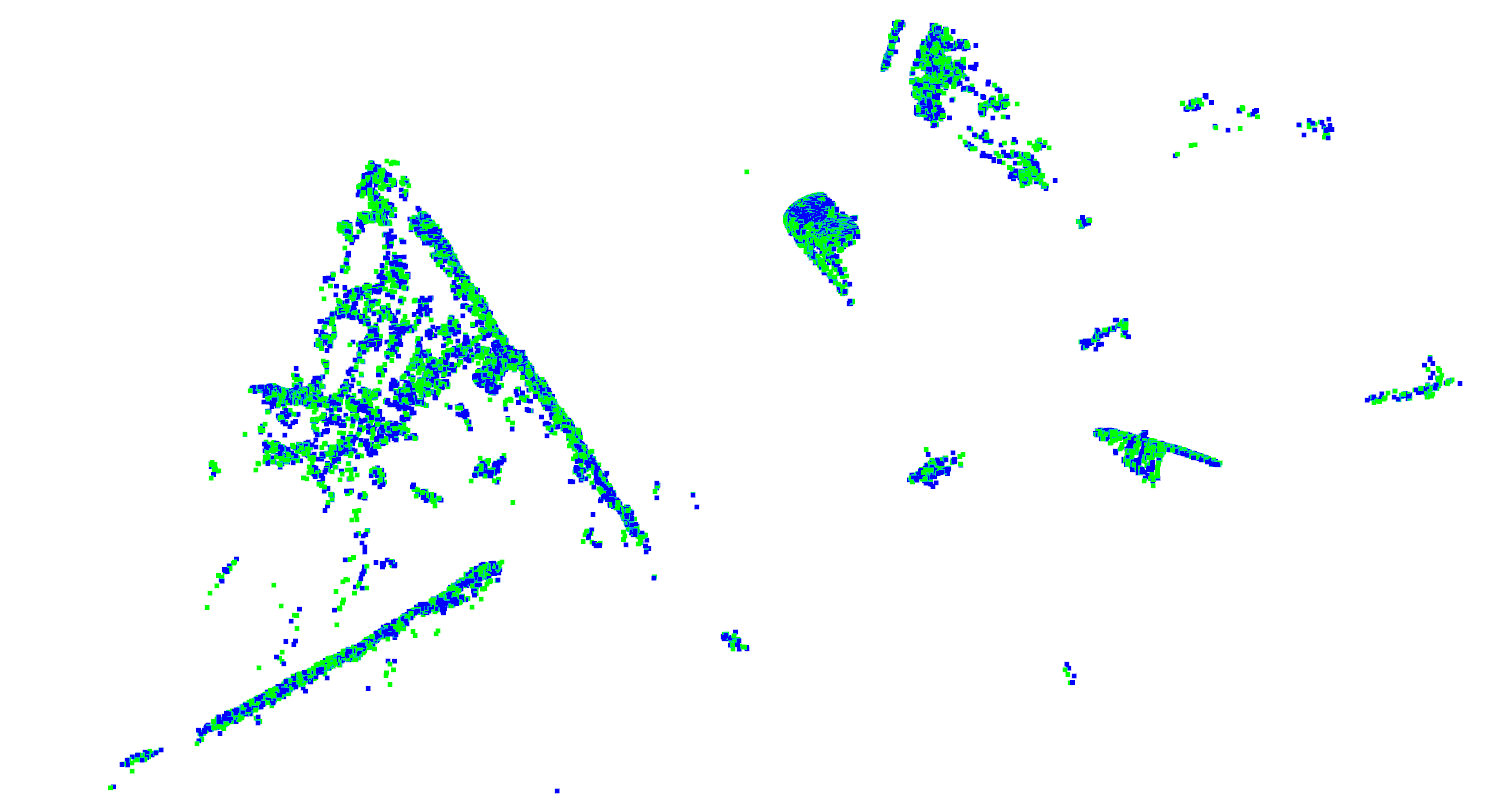}
    \end{subfigure}
    \begin{subfigure}{0.48\columnwidth}
        \centering
        \includegraphics[width=1\columnwidth, trim={0cm 0cm 0cm 1cm}, clip]{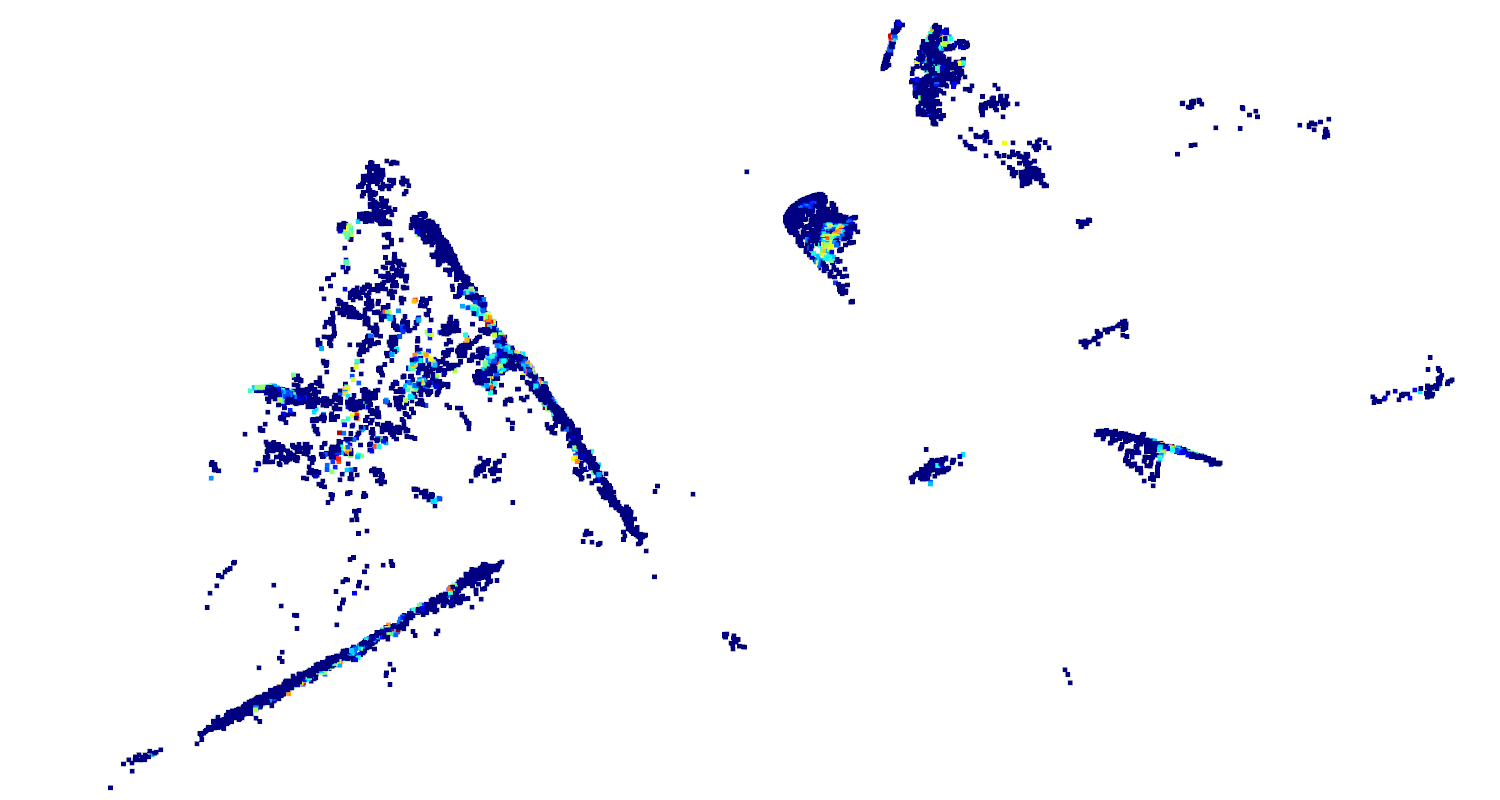}
    \end{subfigure}

    \vspace{1mm}
    \rule{0.95\textwidth}{0.1pt}

    \begin{subfigure}{0.48\columnwidth}
        \centering
        \includegraphics[width=1\columnwidth, trim={0cm 0cm 0cm 1cm}, clip]{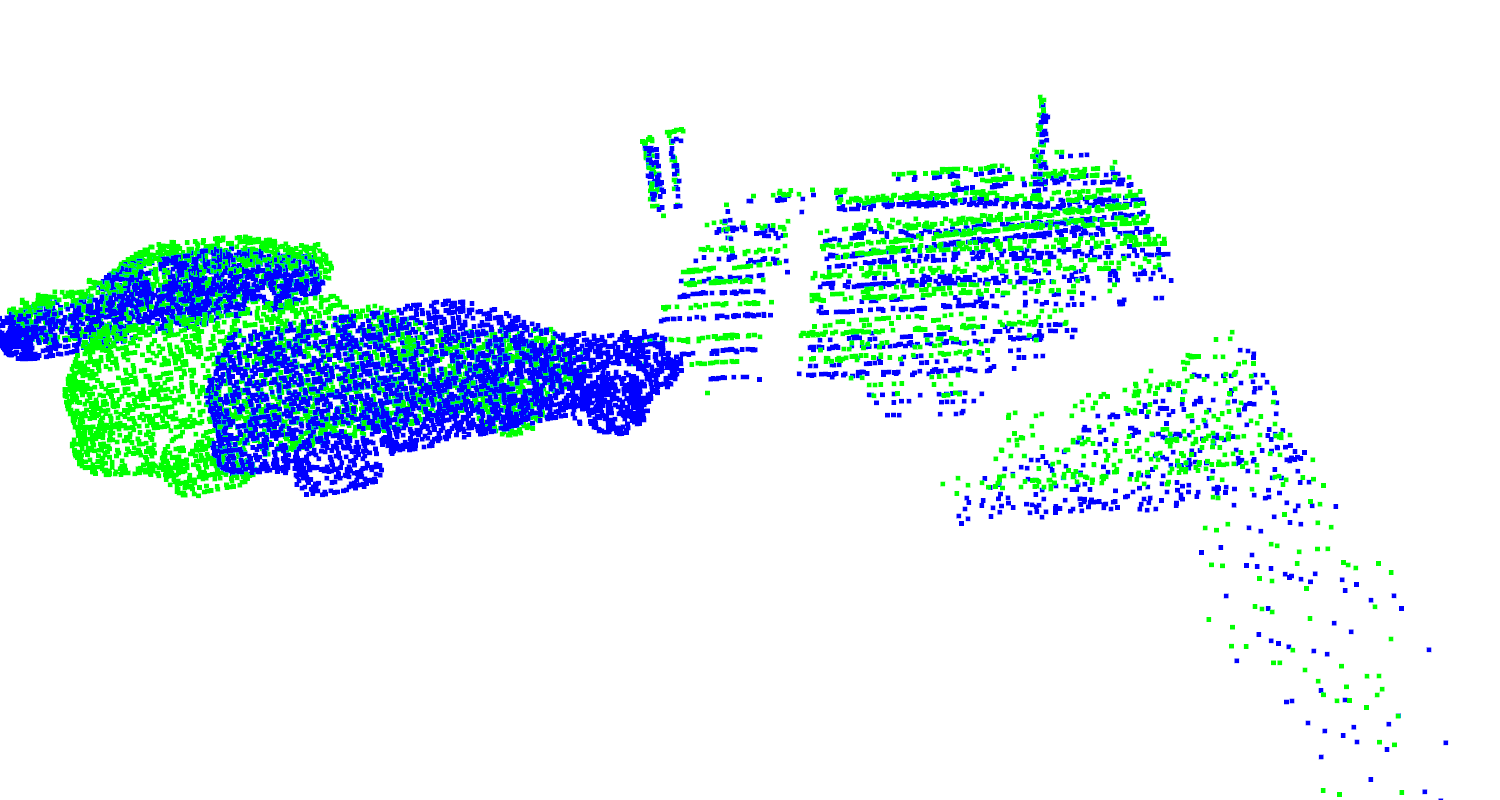}
    \end{subfigure}
    \begin{subfigure}{0.48\columnwidth}
        \centering
        \includegraphics[width=1\columnwidth, trim={0cm 0cm 0cm 1cm}, clip]{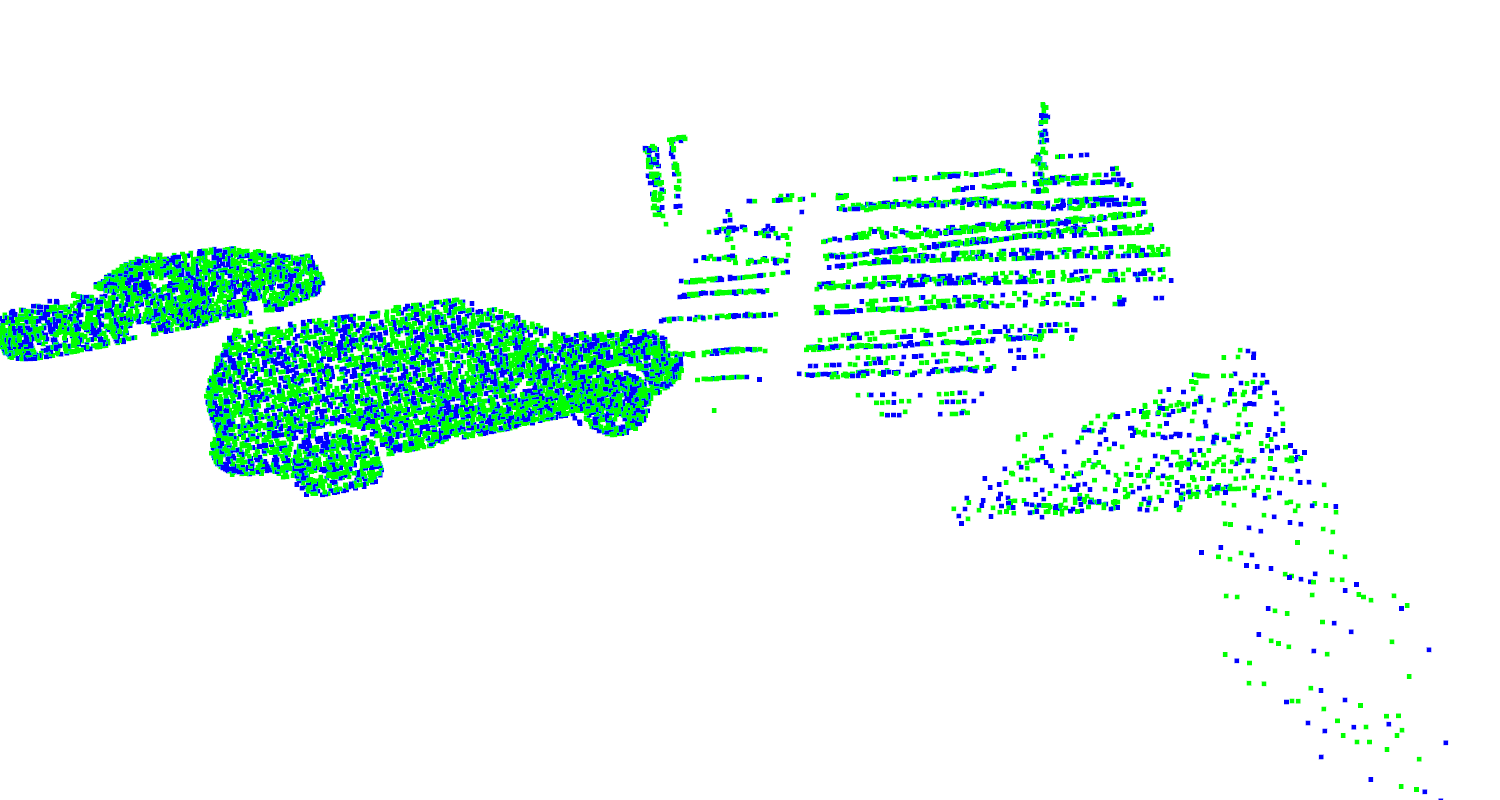}
    \end{subfigure}
    \begin{subfigure}{0.48\columnwidth}
        \centering
        \includegraphics[width=1\columnwidth, trim={0cm 0cm 0cm 1cm}, clip]{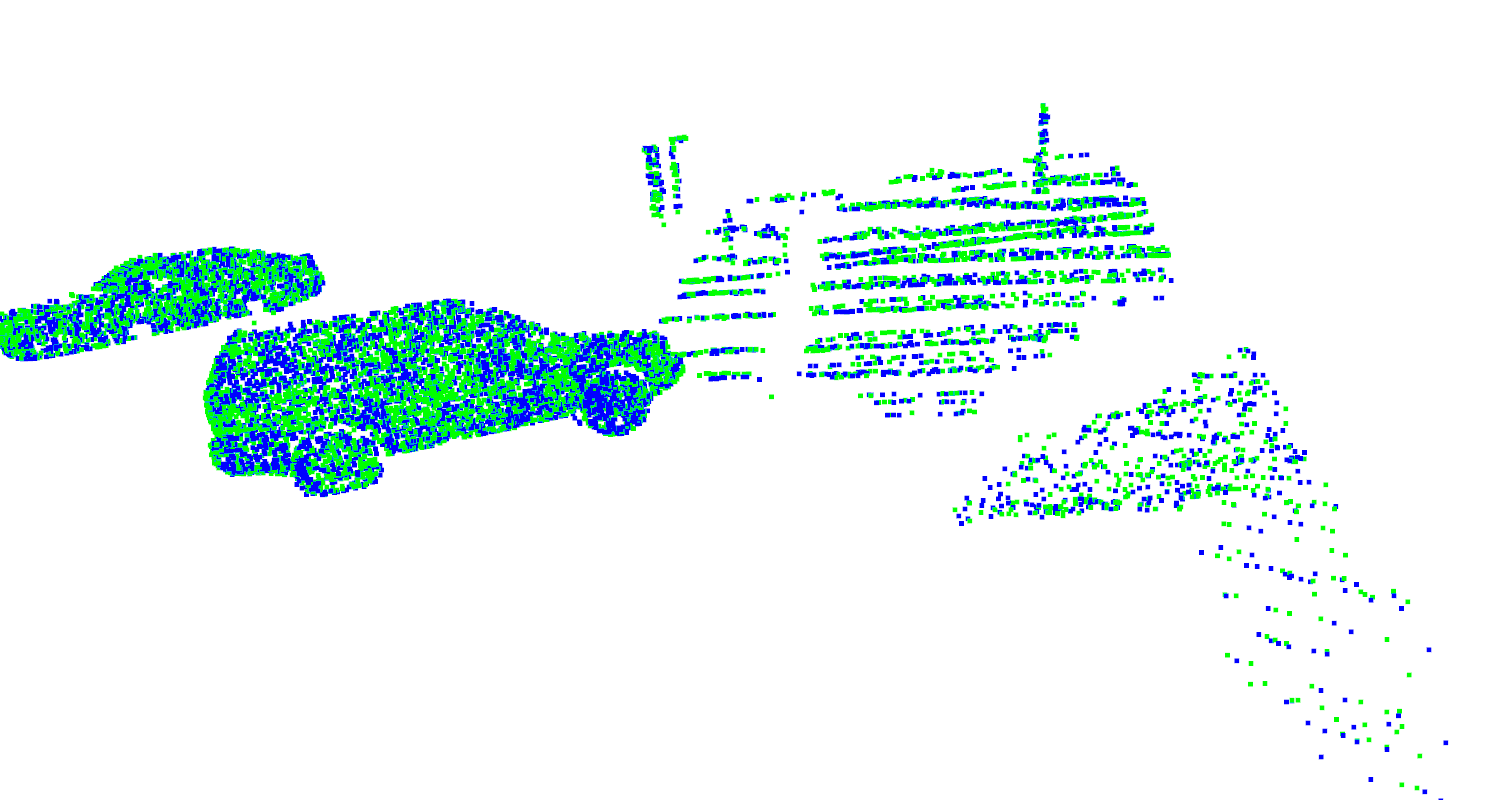}
    \end{subfigure}
    \begin{subfigure}{0.48\columnwidth}
        \centering
        \includegraphics[width=1\columnwidth, trim={0cm 0cm 0cm 1cm}, clip]{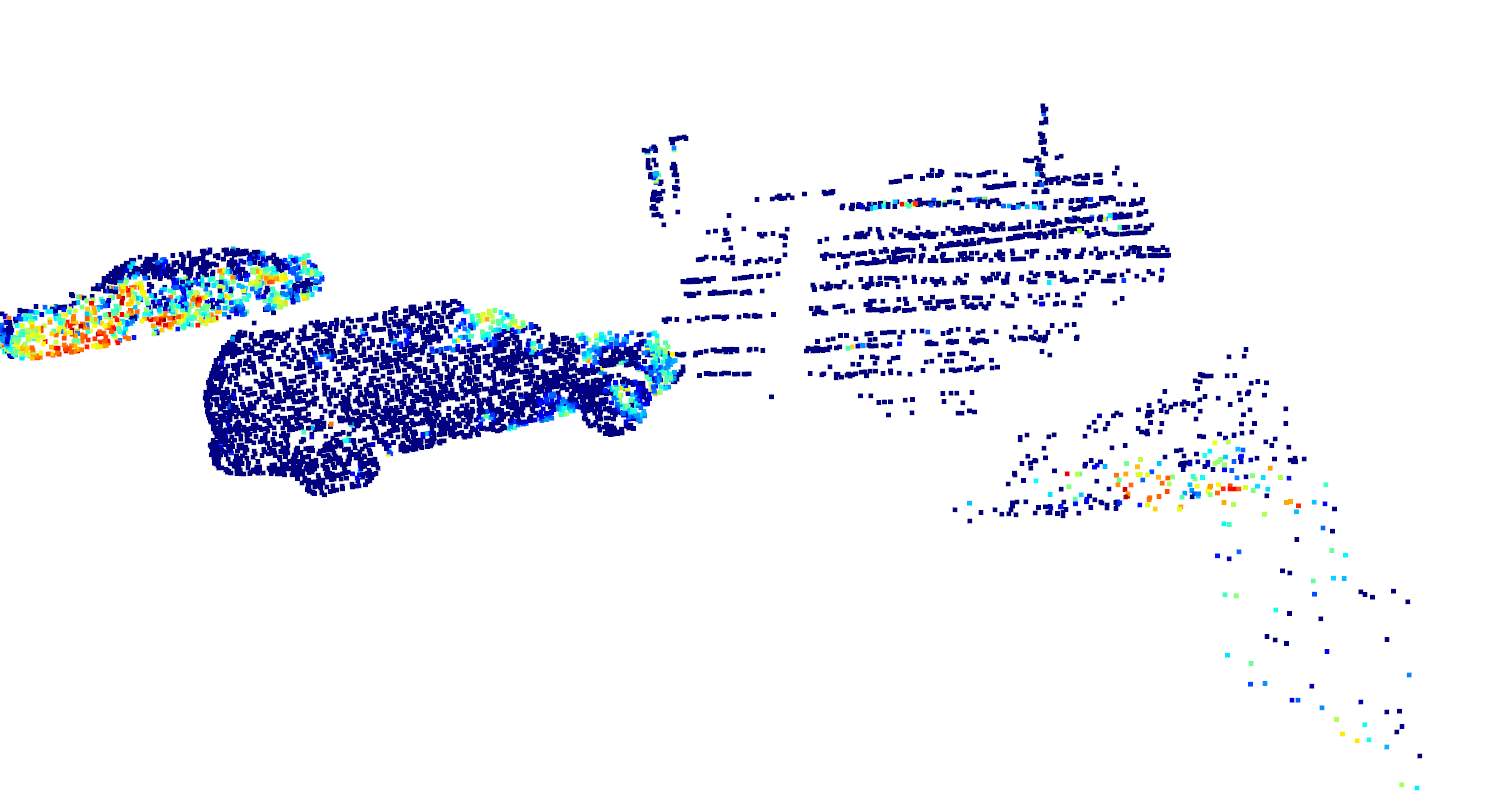}
    \end{subfigure}

    \rule{0.95\textwidth}{0.1pt}

    \begin{subfigure}{0.48\columnwidth}
        \centering
        \includegraphics[width=1\columnwidth, trim={0cm 0cm 0cm 1cm}, clip]{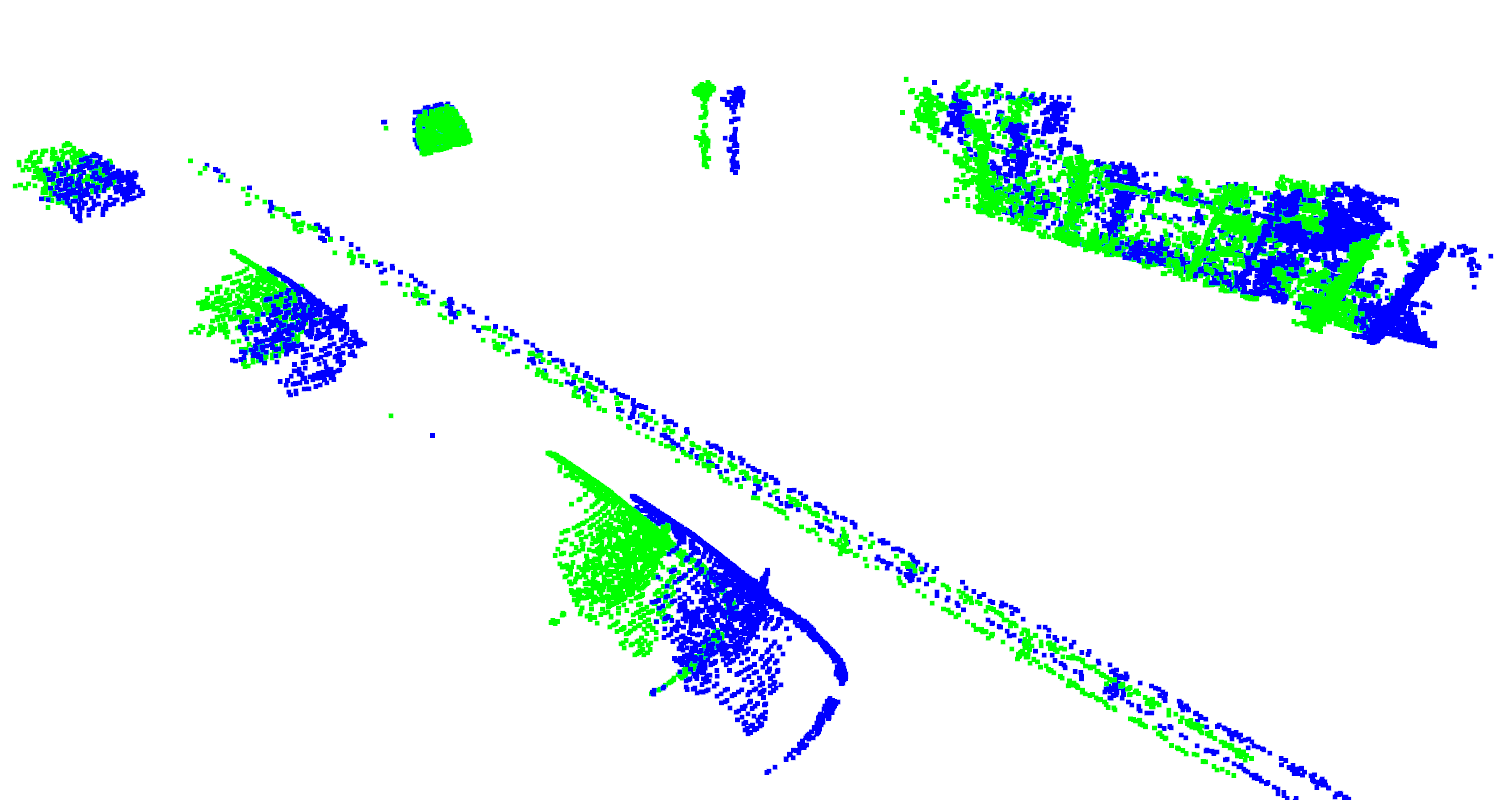}
    \end{subfigure}
    \begin{subfigure}{0.48\columnwidth}
        \centering
        \includegraphics[width=1\columnwidth, trim={0cm 0cm 0cm 1cm}, clip]{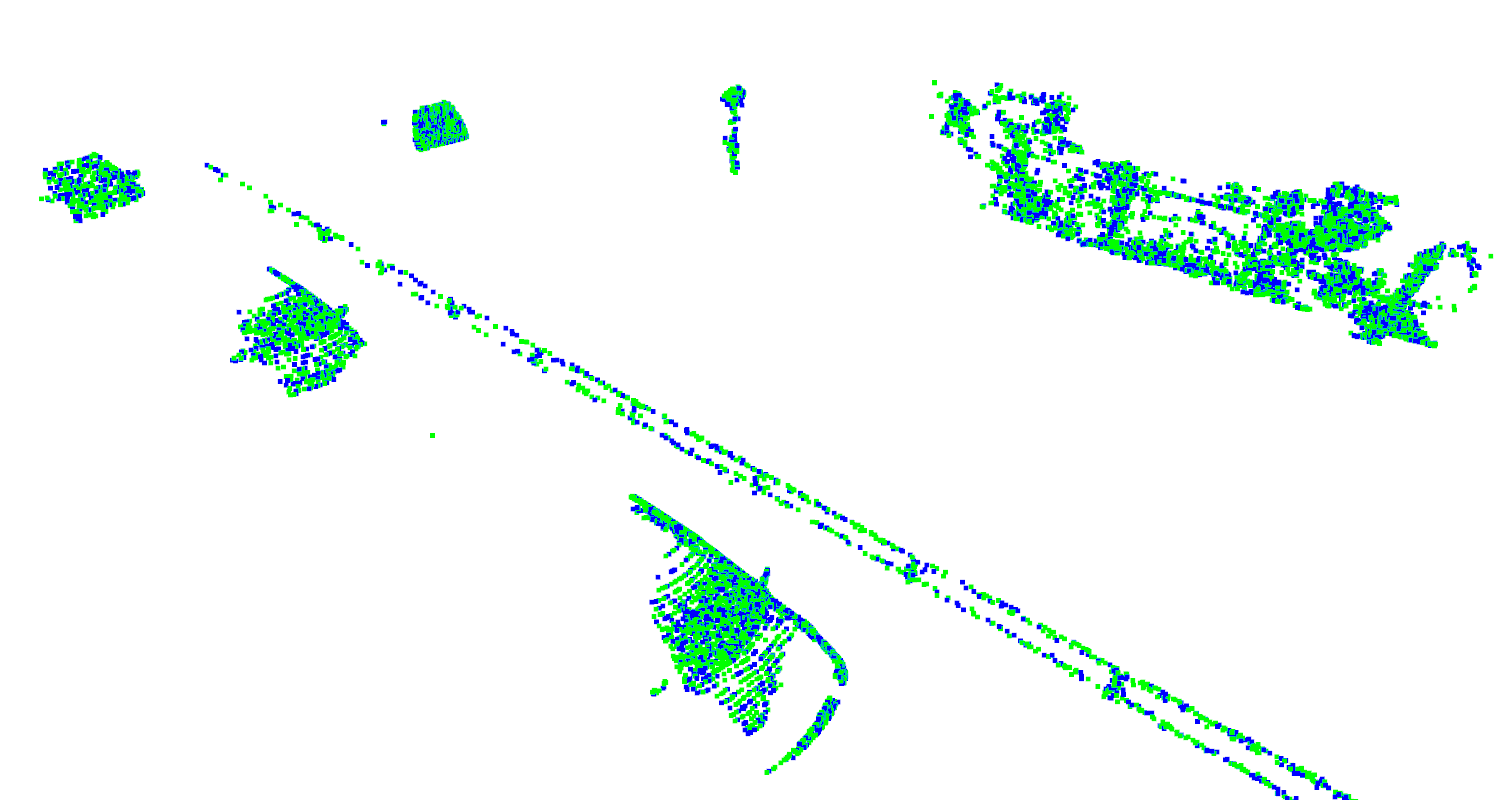}
    \end{subfigure}
    \begin{subfigure}{0.48\columnwidth}
        \centering
        \includegraphics[width=1\columnwidth, trim={0cm 0cm 0cm 1cm}, clip]{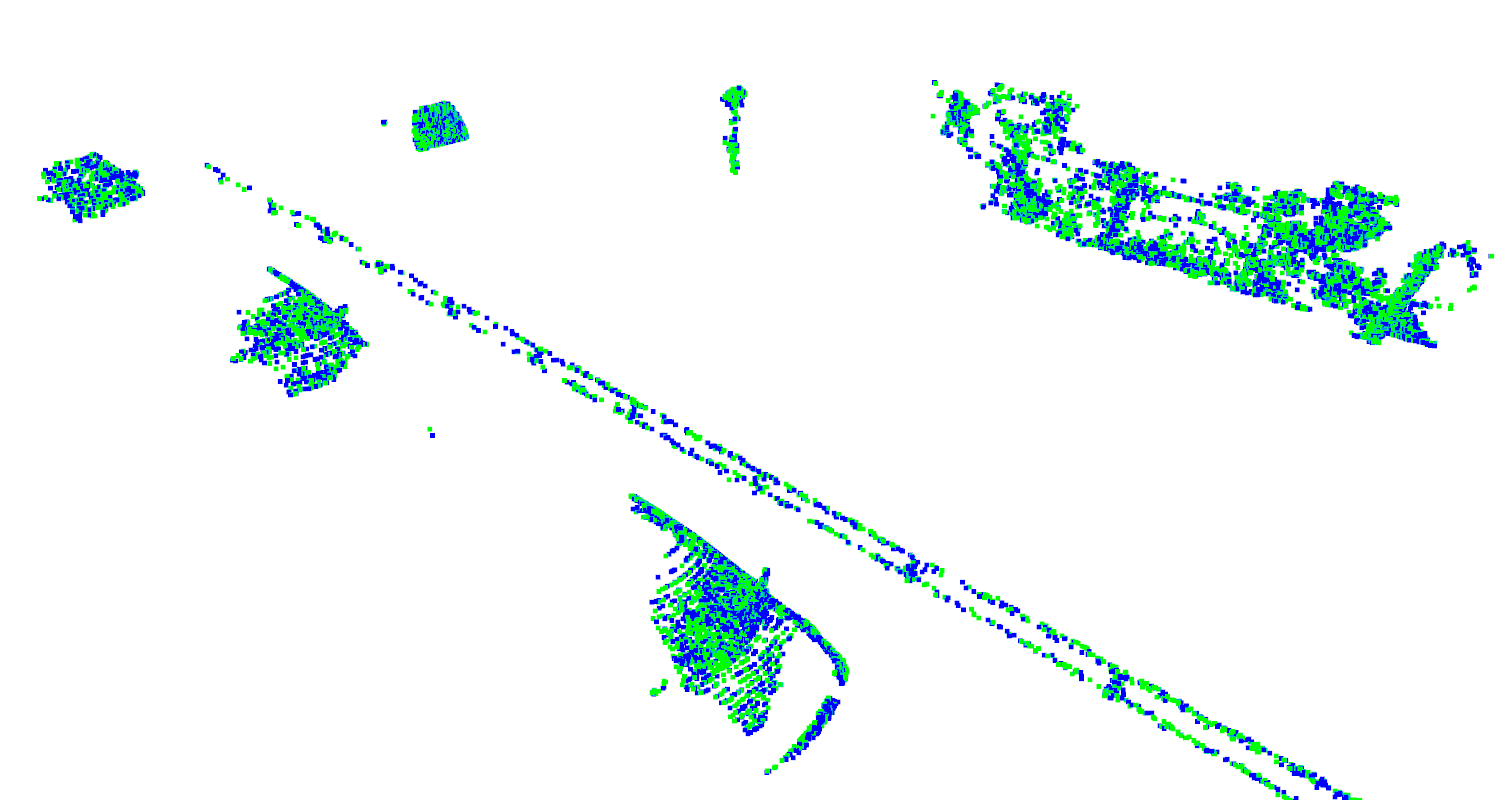}
    \end{subfigure}
    \begin{subfigure}{0.48\columnwidth}
        \centering
        \includegraphics[width=1\columnwidth, trim={0cm 0cm 0cm 1cm}, clip]{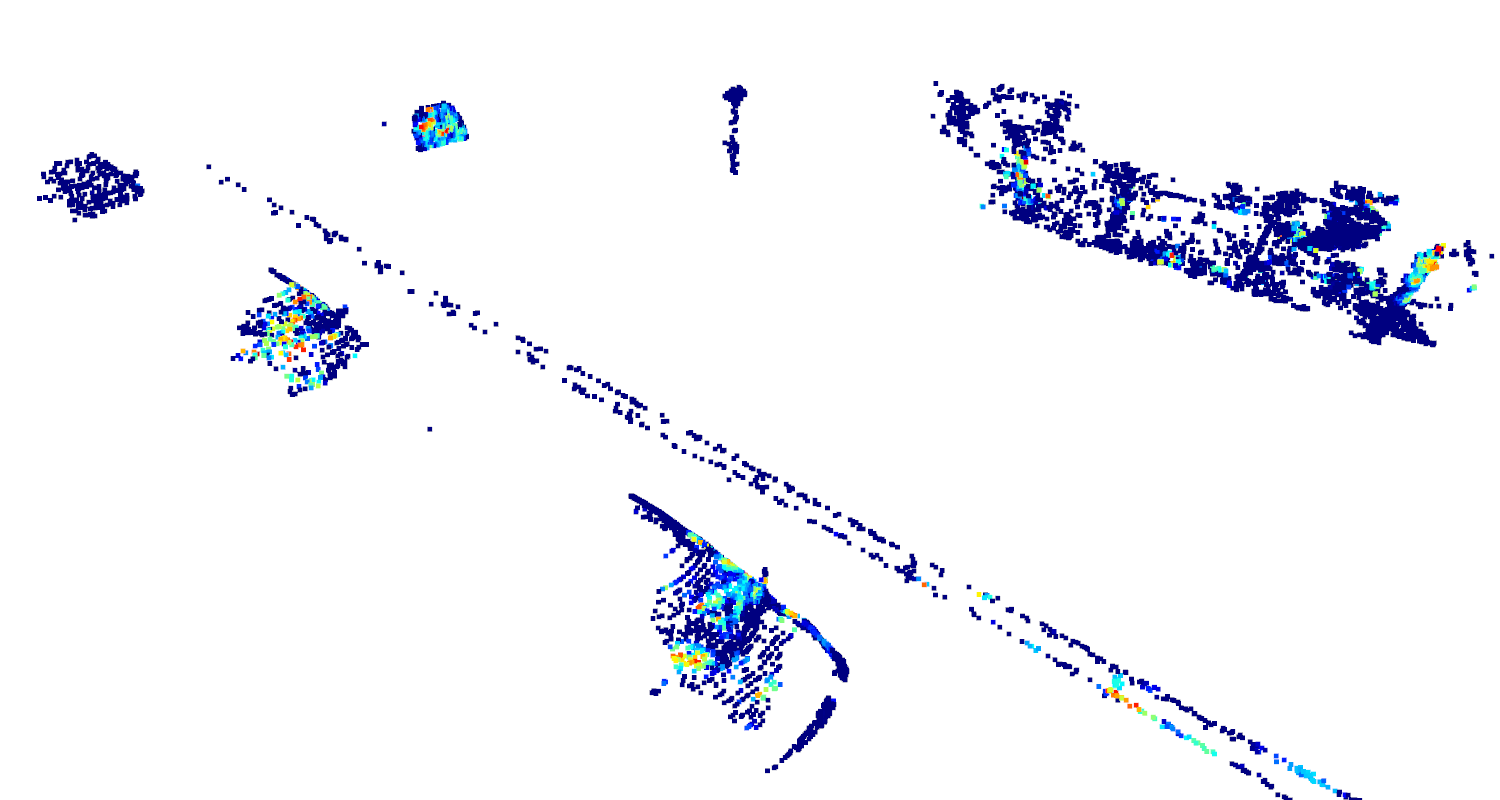}
    \end{subfigure}

    \rule{0.95\textwidth}{0.1pt}
    
    \vspace{2mm}
    
    \begin{subfigure}{0.48\columnwidth}
        \centering
        \includegraphics[width=1\columnwidth, trim={0cm 0cm 0cm 1cm}, clip]{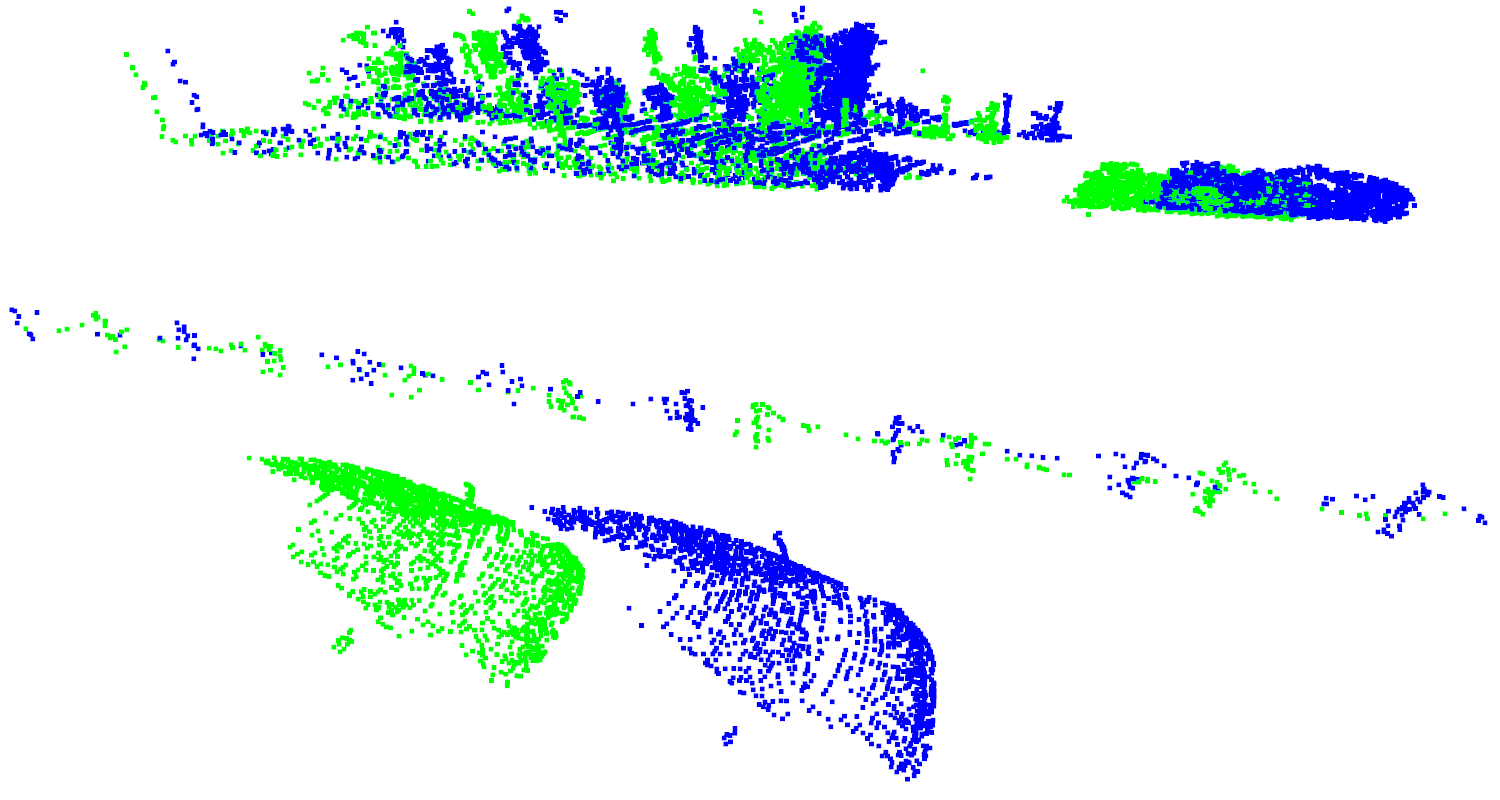}
        \caption*{PC1 and PC2}
    \end{subfigure}
    \begin{subfigure}{0.48\columnwidth}
        \centering
        \includegraphics[width=1\columnwidth, trim={0cm 0cm 0cm 1cm}, clip]{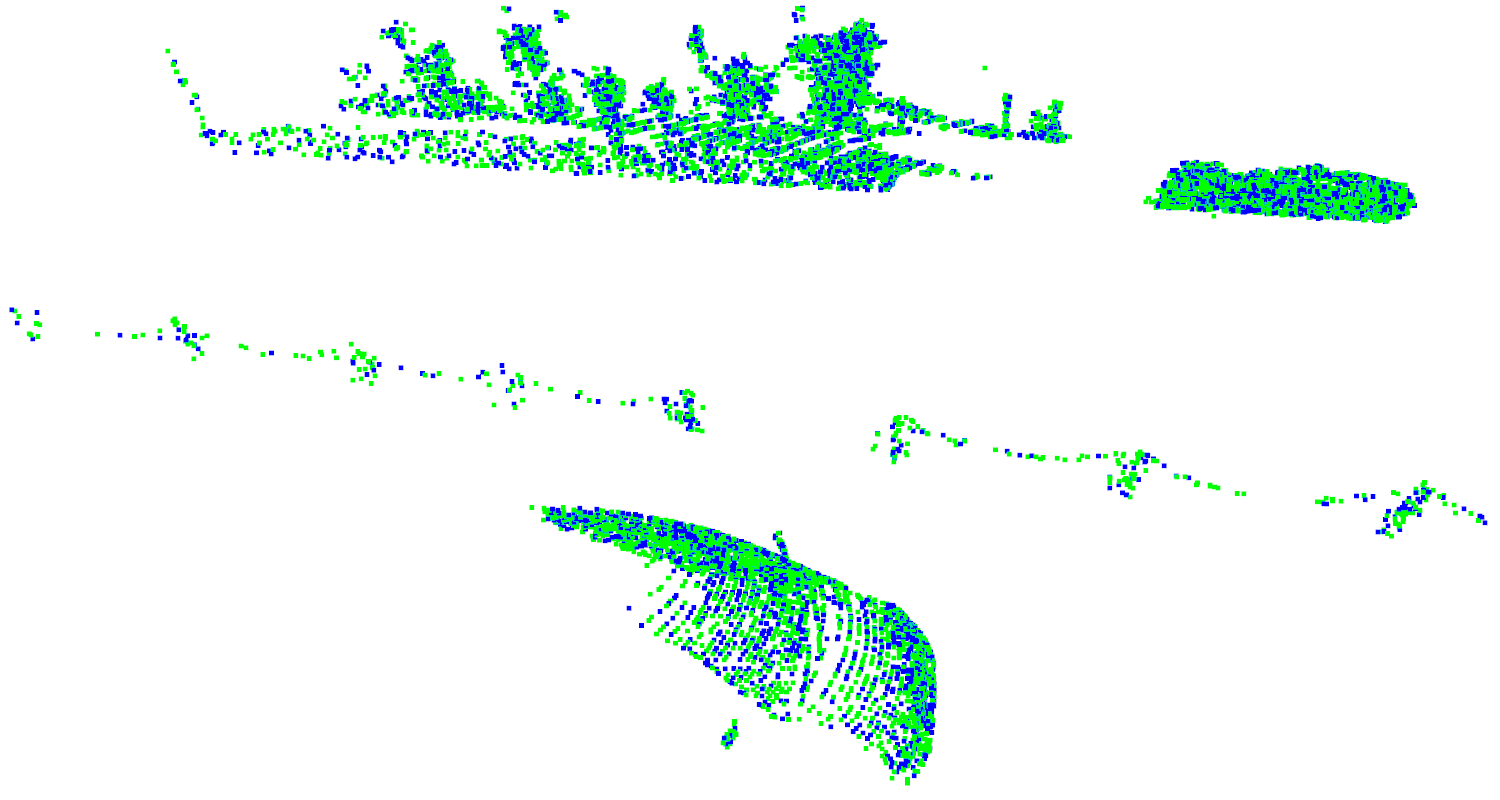}
        \caption*{PC1+GT and PC2}
    \end{subfigure}
    \begin{subfigure}{0.48\columnwidth}
        \centering
        \includegraphics[width=1\columnwidth, trim={0cm 0cm 0cm 1cm}, clip]{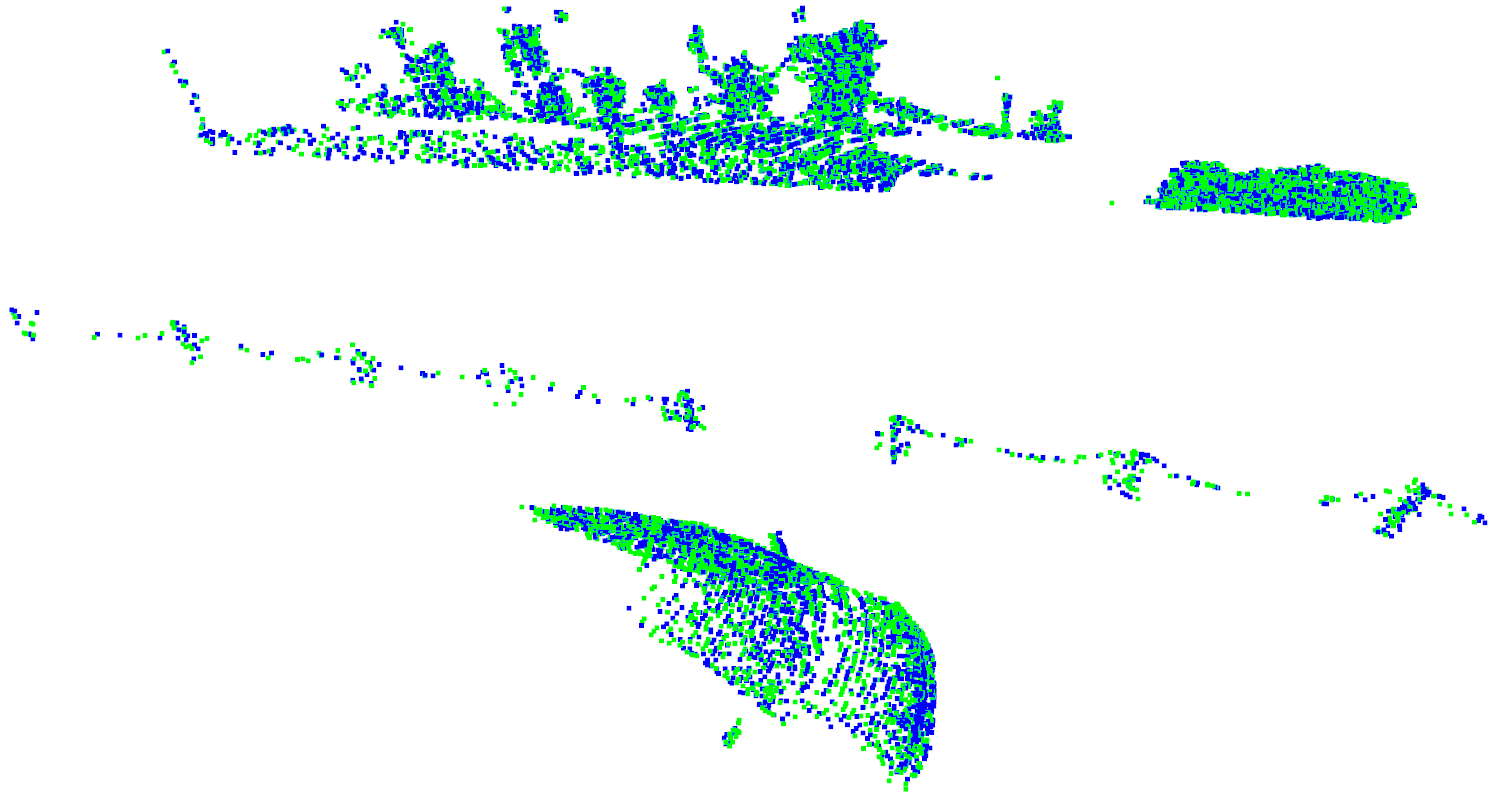}
        \caption*{PC1+Pred and PC2}
    \end{subfigure}
    \begin{subfigure}{0.48\columnwidth}
        \centering
        \includegraphics[width=1\columnwidth, trim={0cm 0cm 0cm 1cm}, clip]{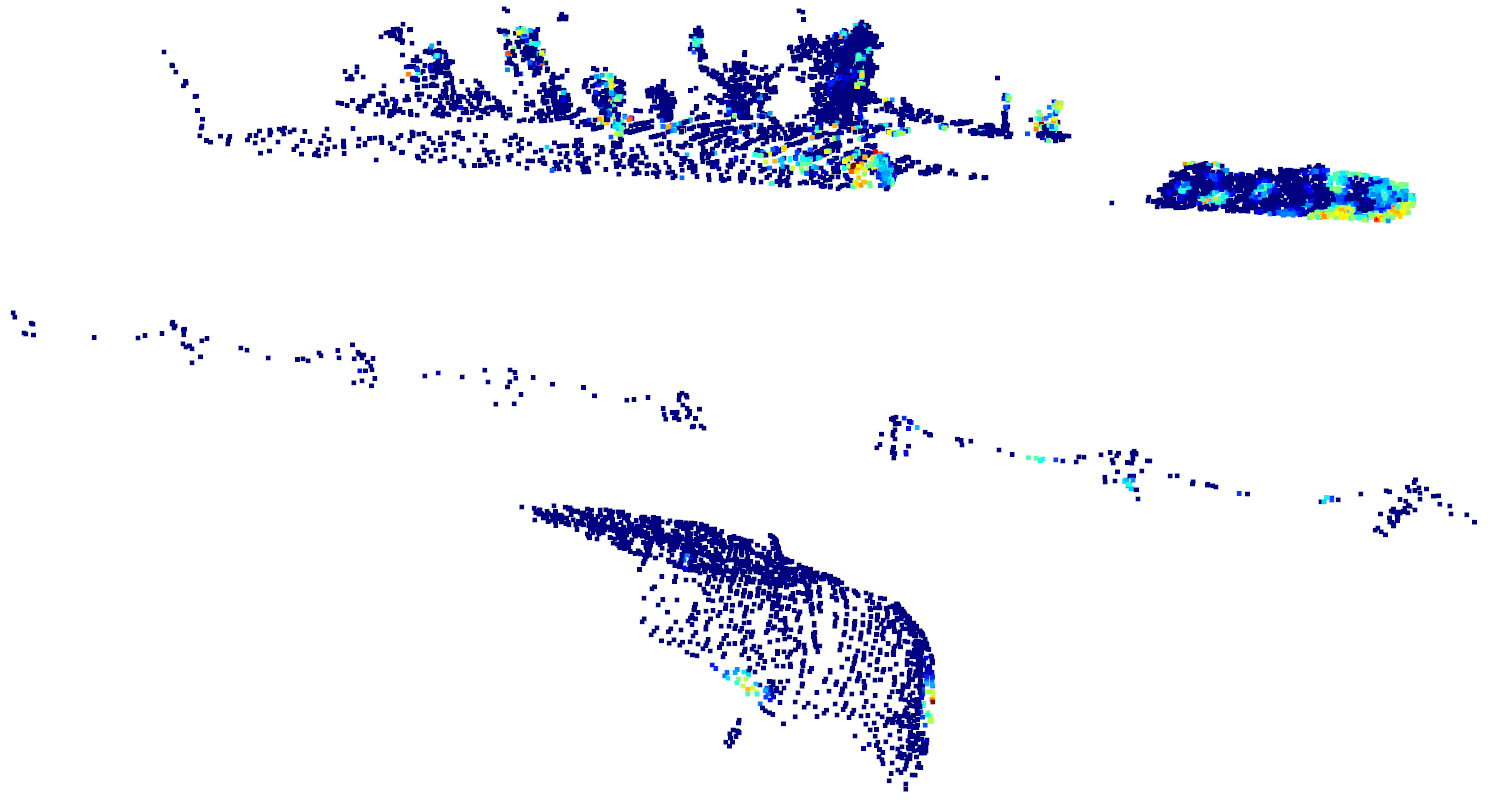}
        \caption*{Error}
    \end{subfigure}
    
    \caption{Qualitative results on the KITTI~\cite{kitti2015} scene flow dataset.}
    \label{fig:res-kitti}
\end{figure*}

\begin{table*}[t]
    \centering
    \begin{tabular}{c | c  c  c  c | c c c c}
    \toprule
    \multirow{2}{*}{Iterations} 
    & \multicolumn{4}{c|}{KITTI} & \multicolumn{4}{c}{FlyingThings3D} \\
    & EPE3D$\downarrow$ & Acc3DS$\uparrow$ & AccDR$\uparrow$ & Outliers3D$\downarrow$ & EPE3D$\downarrow$ & Acc3DS$\uparrow$ & AccDR$\uparrow$ & Outliers3D$\downarrow$ \\
    \midrule
    0  & $0.1696$ & $0.2744$ & $0.5967$ & $0.5671$ & $0.1277$  & $0.1809$  & $0.6127$  & $0.7428$  \\
    3  & $0.0660$  & $0.7513$ & $0.8877$ & $0.1893$ & $0.0460$  & $0.80720$ & $0.9517$  & $0.2476$  \\
    7  & $0.0570$ & $0.7979$ & $0.9154$ & $0.1566$ & $0.0403$  & $0.8567$  & $0.9635$  & $0.1976$  \\
    10 & $0.0521$ & $0.8245$ & $0.9299$ & $0.1385$ & $0.0409$  & $0.8532$  & $0.9631$  & $0.2035$ \\    
    14 & $0.0481$ & $0.8491$ & $0.9448$ & $0.1228$ & $0.0429$  & $0.8368$  & $0.9606$  & $0.2227$ \\
    \bottomrule
    \end{tabular}
    \caption{Ablation studies of iterations.}
    \label{tab:ablation_iter}
    \vspace{-5mm}
\end{table*}

\begin{figure*}[ht]
    \centering
 
    \begin{subfigure}{0.68\columnwidth}
        \centering
        \includegraphics[width=1\columnwidth, trim={0cm 0cm 0cm 1cm}, clip]{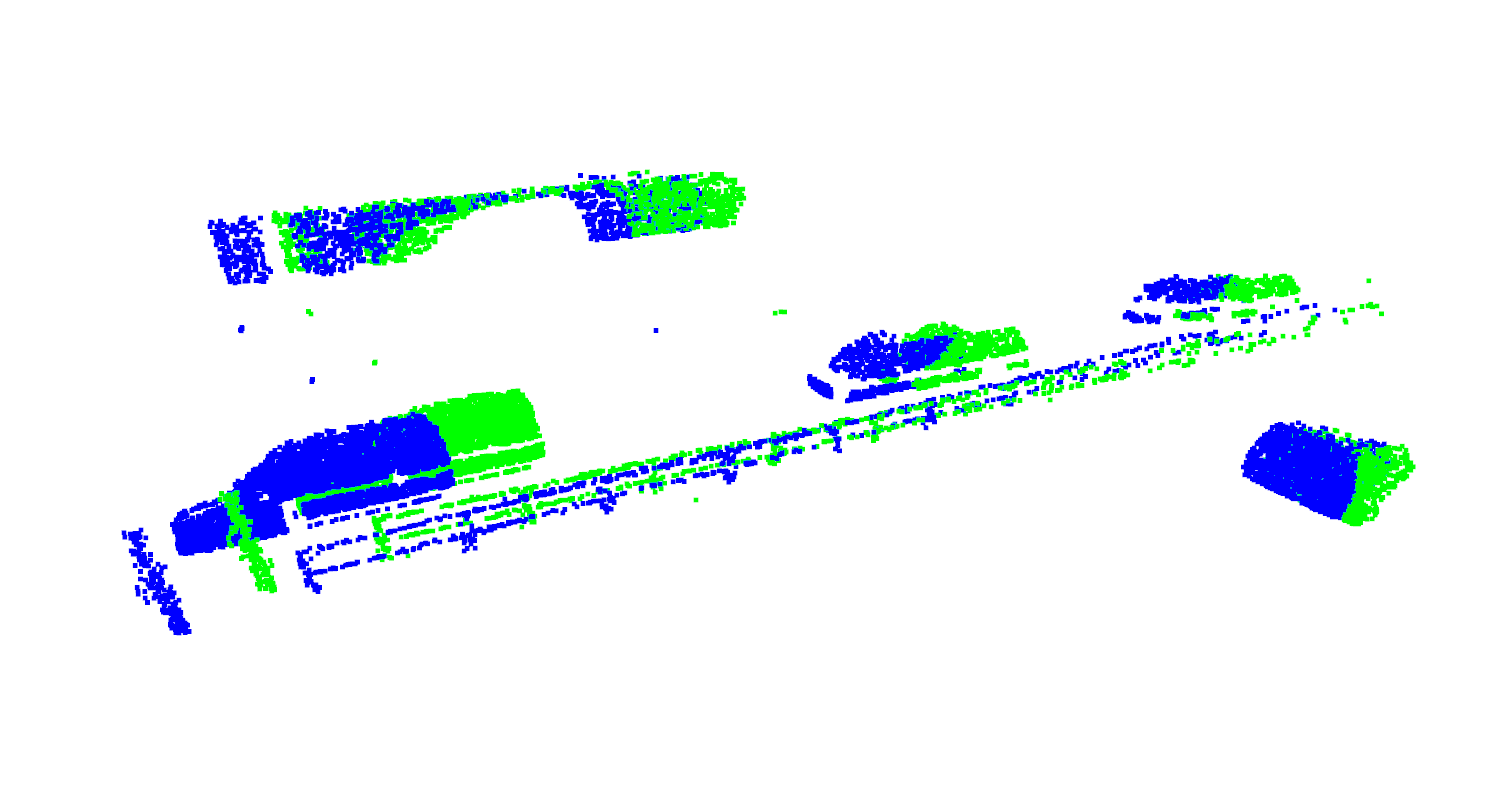}
        \caption*{Input}
    \end{subfigure}
    \begin{subfigure}{0.68\columnwidth}
        \centering
        \includegraphics[width=1\columnwidth, trim={0cm 0cm 0cm 1cm}, clip]{fig2/sceneflow/kitti/iters/big_v1/0_341_000023_img_it0}
        \caption*{iter=0}
    \end{subfigure}
    \begin{subfigure}{0.68\columnwidth}
        \centering
        \includegraphics[width=1\columnwidth, trim={0cm 0cm 0cm 1cm}, clip]{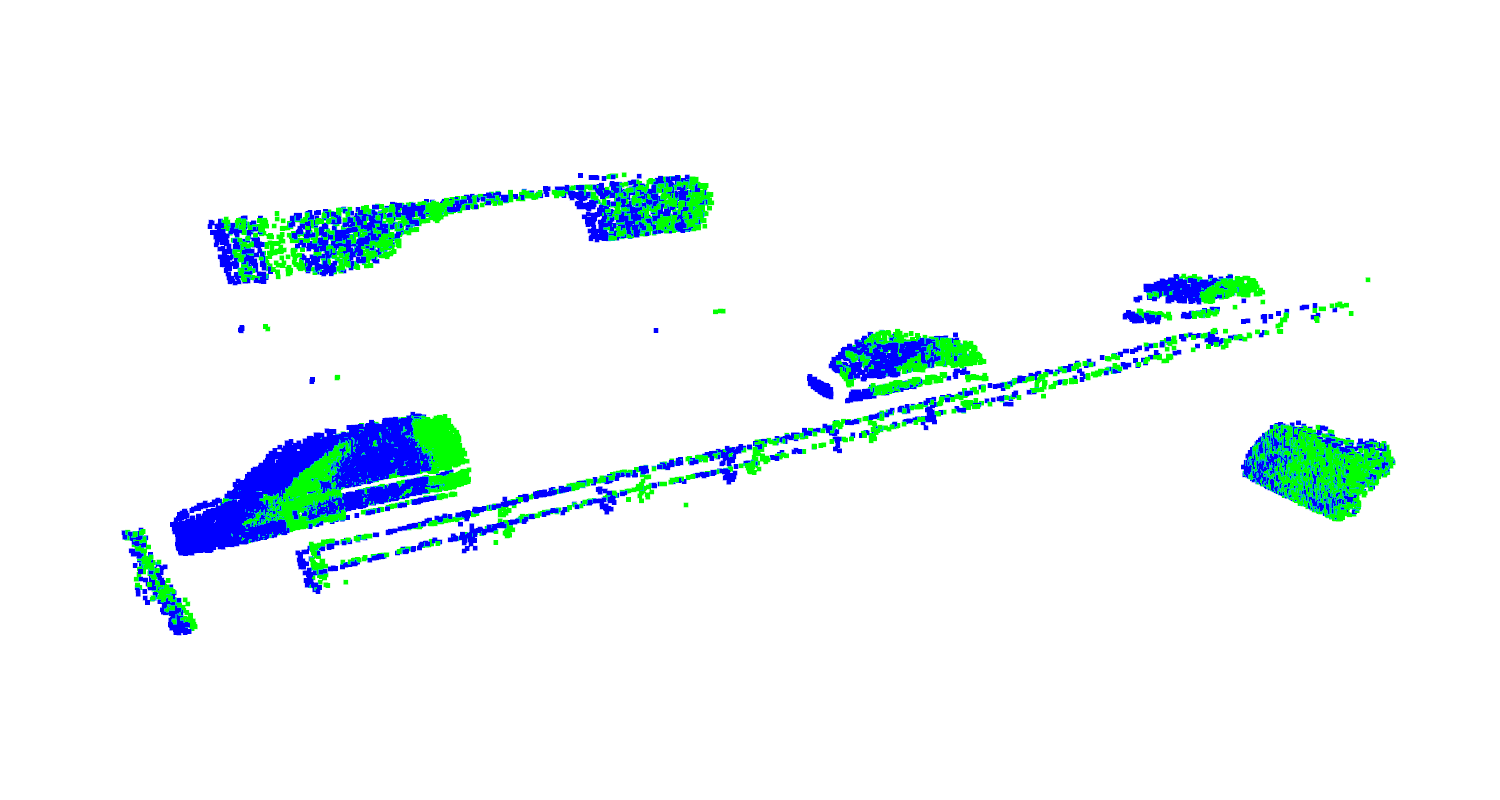}
        \caption*{iter=1}
    \end{subfigure}

    \begin{subfigure}{0.68\columnwidth}
        \centering
        \includegraphics[width=1\columnwidth, trim={0cm 0cm 0cm 1cm}, clip]{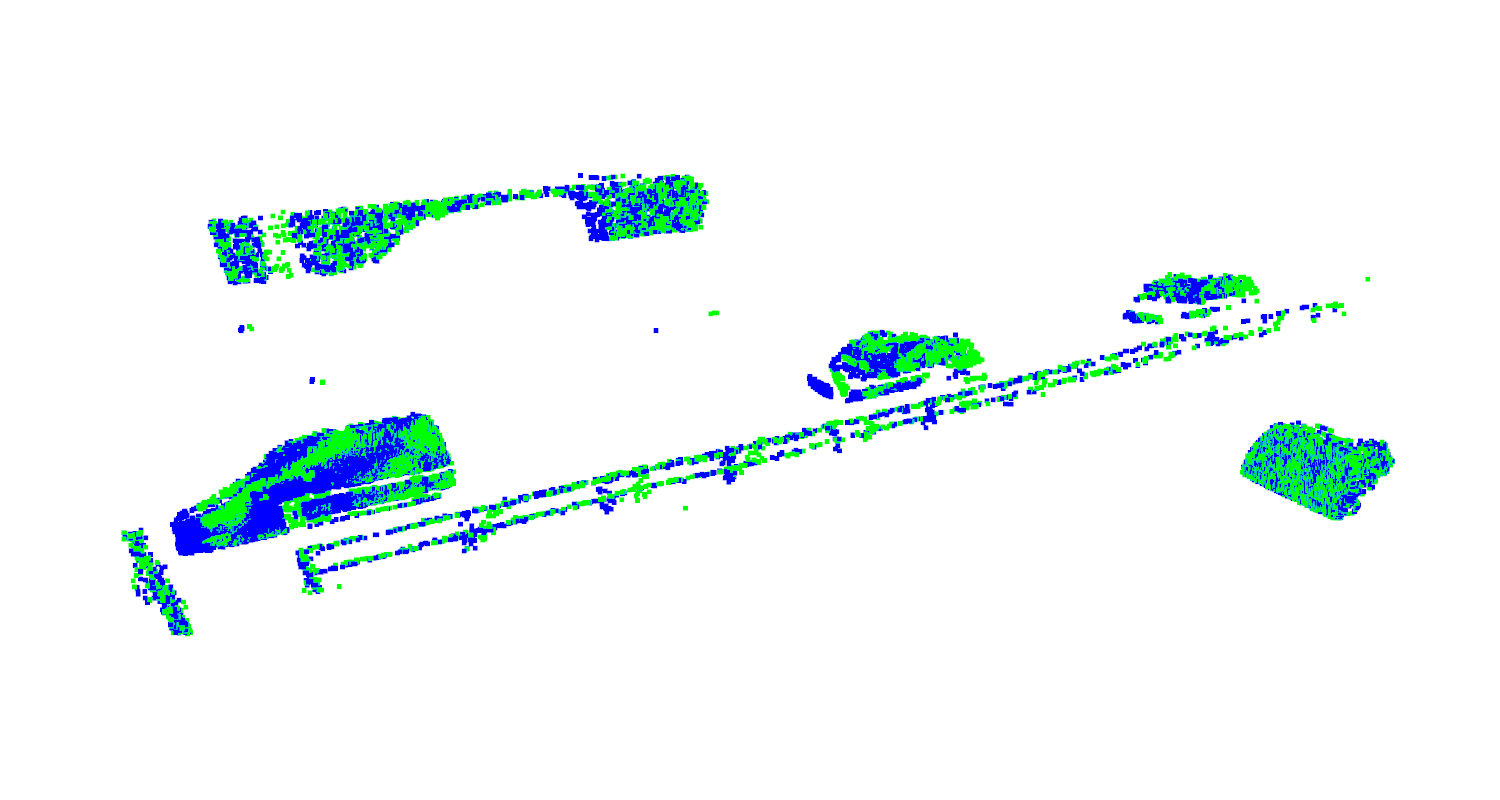}
        \caption*{iter=3}
    \end{subfigure}
    \begin{subfigure}{0.68\columnwidth}
        \centering
        \includegraphics[width=1\columnwidth, trim={0cm 0cm 0cm 1cm}, clip]{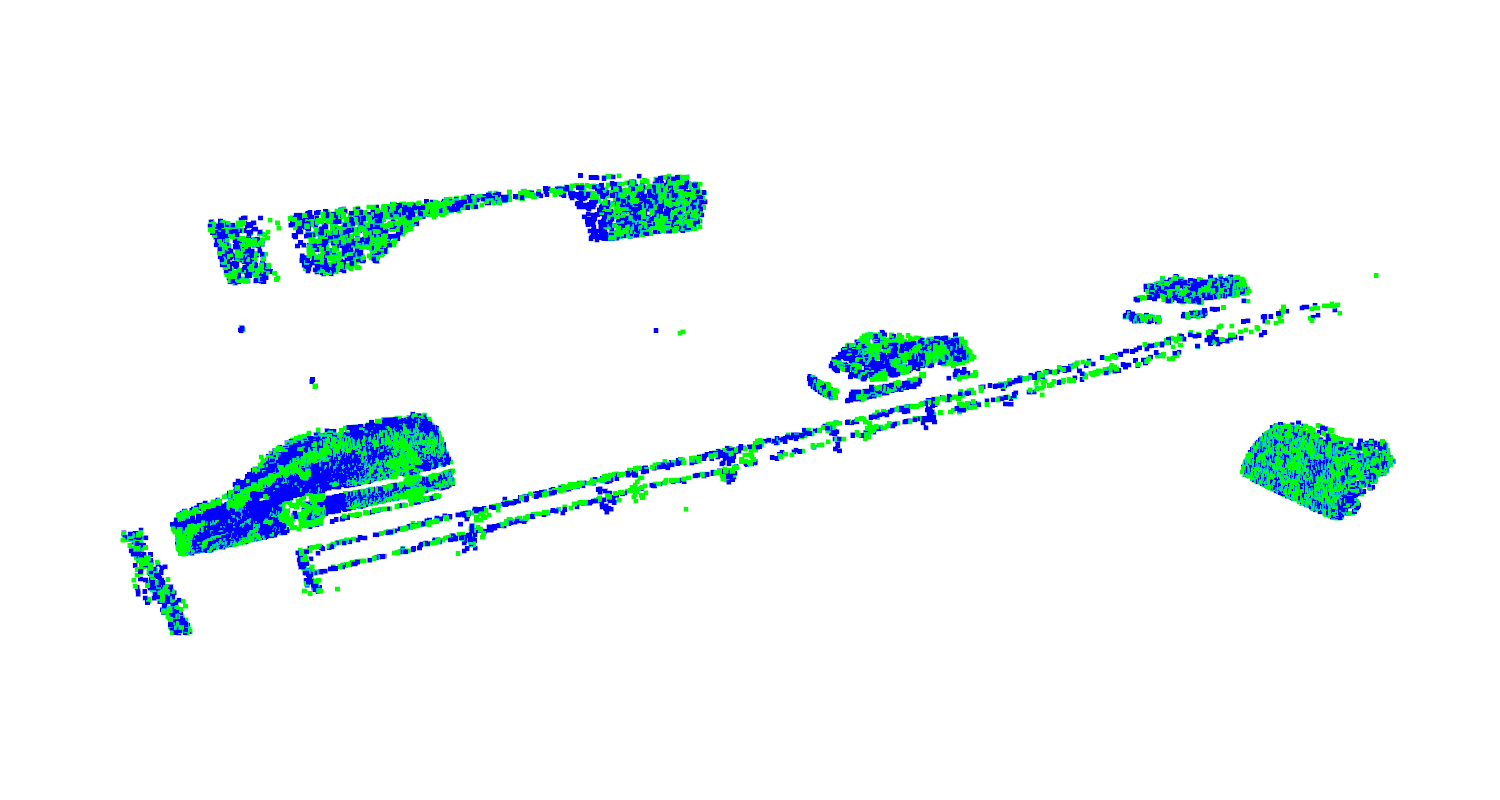}
        \caption*{iter=9}
    \end{subfigure}
    \begin{subfigure}{0.68\columnwidth}
        \centering
        \includegraphics[width=1\columnwidth, trim={0cm 0cm 0cm 1cm}, clip]{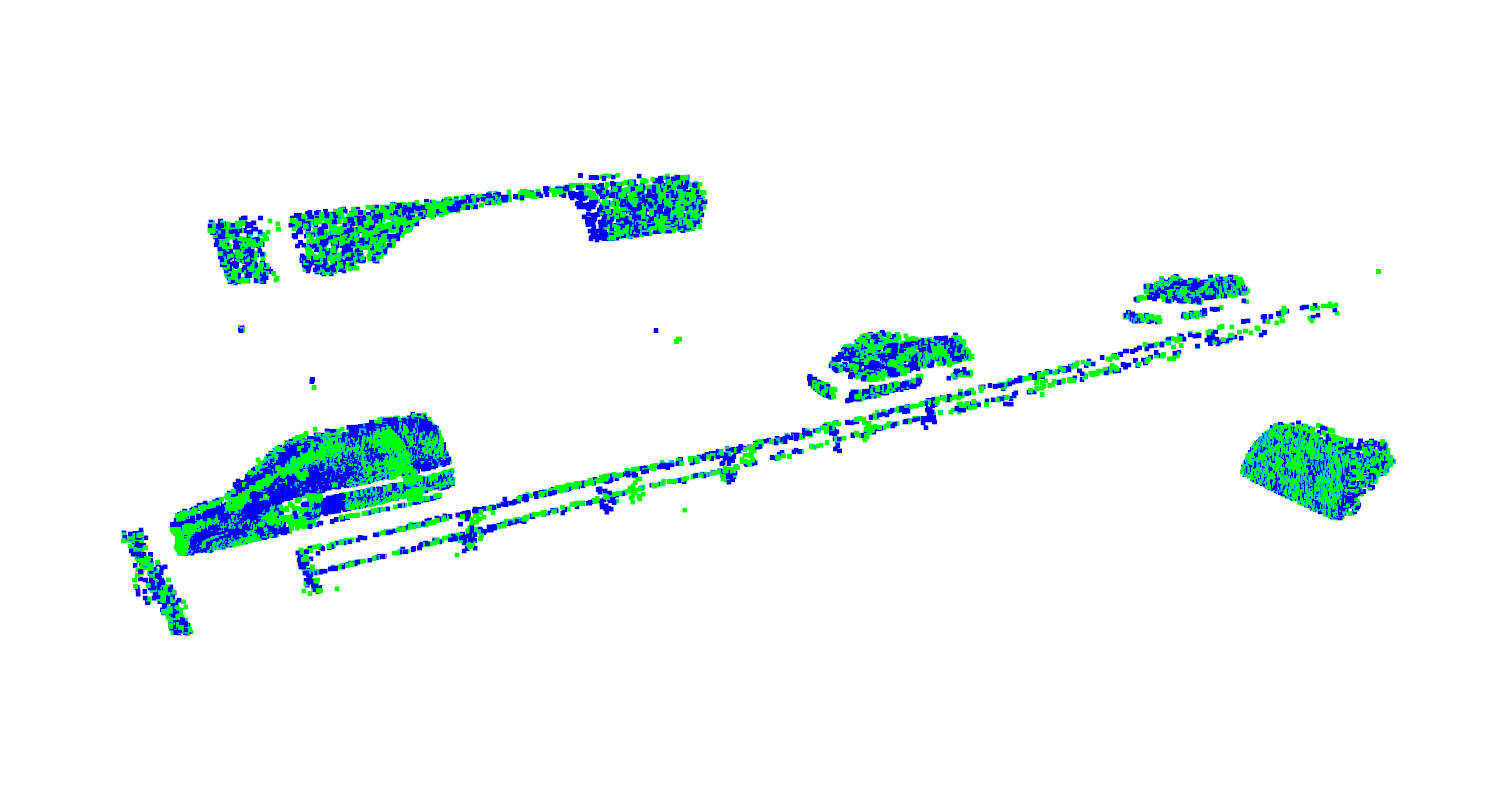}
        \caption*{iter=14}
    \end{subfigure}

    \caption{Visualization of different iterations. As the number of iterations increases, two point clouds from KITTI\cite{kitti2015} scene flow dataset are gradually aligned together. }
    \label{fig:iterations}
    \vspace{-5mm}
\end{figure*}

\subsection{Point Cloud Registration}

\paragraph{Dataset} We follow previous methods~\cite{rpmnet} to evaluate on the ModelNet40~\cite{modelnet} dataset which contains 40 object categories of CAD models. Each CAD model is sampled into 2048 points and normalized into a unit sphere as in~\cite{rpmnet}. 
In order to generate training pairs with ground-truth rigid transformations, we follow RPMNet\cite{rpmnet} to first synthetic transformation and then transform an existing source point cloud to the target point cloud.
The transformation is synthesized by randomly sampling the rotation vector between $[0^\circ, 45^\circ]$ and the translation vector between $[-0.5, 0.5]$. To simulate the condition of partial-to-partial registration, $30\%$ of the points are further removed according to a random direction~\cite{rpmnet}.  
We use the first 20 categories for training and validation and the remained 20 categories for evaluation.

\paragraph{Training Details}
Compared with the FlyingThings3D, the ModelNet40 is a relatively small scale dataset, so we train our model for point cloud registration in a single stage, where the batch size is 8 on each GPU, and alternate between the~\equref{point_wise3} and the~\equref{recurrent} for 7 times. The learning rate also adopts the same decay strategy as above and is decayed every 100 epochs, while the model is trained 600 epochs.

\paragraph{Quantitative Comparisons} The quantitative comparisons follow previous works and use both the isotropic metrics $Error(R)$ and $Error(t)$~\cite{rpmnet} as well as the anisotropic metric $MAE(R)$ and $MAE(t)$~\cite{deepcp}:
\begin{enumerate}
    \vspace{-2mm}
    \item [$\bullet$] $Error(R)$:  $arcos\frac{tr({R_{gt}}^{-1}R_{pred}-1)}{2}$
    \vspace{-2mm}
    \item [$\bullet$] $Error(t)$:  $||R_{gt}^{-1}t-t_{gt}||_2$
    \vspace{-2mm}
    \item [$\bullet$] $MAE(R)$:  $||Euler(R_{gt})- Euler(R_{pred})||_1$
    \vspace{-2mm}
    \item [$\bullet$] $MAE(t)$:  $||R_{gt}^{-1}t-t_{gt}||_1$.
    \vspace{-2mm}
\end{enumerate}
As shown in \tabref{icp_paritial_reg}, our method outperforms previous methods on both isotropic metrics and anisotropic metrics.

\subsection{Ablation Studies}
In order to further analyze the effectiveness of each component, we conduct ablation studies on 3D flow estimation and evaluate different variations of our model.

\paragraph{Point-wise Optimization}
The first question is whether we need to further update the transformation after the recurrent regularization, \ie, can we remove the point-wise optimization~\equref{point_wise3} and only preserve the recurrent regularization~\equref{recurrent} to answer this question, we exclude~\equref{point_wise3} for both training and inference, which performs significantly worse as shown in the first row of \tabref{ablation_cost_feats} and indicate that the point-wise optimization is necessary. We also explore different alternative options for point-wise optimization. First, we only optimize the feature metric distance and do not constrain the difference between the auxiliary variable $\mathcal{Z}$ and $\mathcal{X}$. As shown in the second row of \tabref{ablation_cost_feats}, the result is even worse than not using~\equref{point_wise3}, which is because the feature distance only is ambiguous in some regions such as planar areas. Second, we use~\equref{point_wise1} directly instead of the soften version~\equref{point_wise3}. We train the model with Gumbel Softmax and inference with argmax. The Gumbel Softmax performs worse than the soften version in~\equref{point_wise3}, which is because the gradient back-propagated by the~\equref{point_wise3} is less noisy for training. Additionally, we also evaluate~\equref{point_wise2} using bilateral weight, which performs worse than~\equref{point_wise3}. Therefore, we use~\equref{point_wise3} as the default setting to minimize~\equref{point_wise1} point-wisely.

\begin{table}[t]
    \small
    \centering
    \resizebox{\linewidth}{!}{
    \begin{tabular}{c | c  c  c  c}
    \toprule
    Point-wise Optimization & EPE3D$\downarrow$  & Acc3DS$\uparrow$  &  AccDR$\uparrow$  & Outliers3D$\downarrow$ \\
    \midrule
    W/O & $0.0514$ & $0.7612$ & $0.9308$ & $0.2880$ \\
    Feature Only& $0.0445$ & $0.8327$ & $0.9614$ & $0.2318$ \\
    Gumbel Softmax& $0.0451$ & $0.8262$ & $0.9604$ & $0.2370$ \\
    Bilateral Weight& $0.0441$ & $0.8361$ & $0.9555$ & $0.2305$ \\
    Ours & $0.0403$ & $0.8567$ & $0.9635$ & $0.1976$ \\
    \bottomrule
    \end{tabular}}
    \caption{Ablation studies of point-wise optimization module. We show some alternative ways.}
    \label{tab:ablation_cost_feats}
    \vspace{-6mm}
\end{table}

\paragraph{Recurrent Regularization}
The second question is whether we need to solve~\equref{reg_term} with a recurrent network, \ie, can we use standard point convolution? We replace the GRU~\cite{GRU} block with three $set\_conv$~\cite{pointnet++} layers to keep similar model parameters for fair comparison, which increases the EPE3D from $0.0405$ to $0.0491$ as shown in \tabref{ablation_main}. It indicates that the historical information passed by the recurrent unit regularizes the solution better. We also replace the GRU~\cite{GRU} unit with LSTM~\cite{lstm}, which has slight improvement in~\tabref{ablation_main}. Therefore, we use GRU~\cite{GRU} as our default recurrent unit because of its simpler formulation.

\begin{table}[t]
    \small
    \centering
    \resizebox{\linewidth}{!}{
    \begin{tabular}{c | c  c  c  c}
    \toprule
    Recurrent Reg. & EPE3D$\downarrow$ & Acc3DS$\uparrow$ & AccDR$\uparrow$ & Outliers3D$\downarrow$ \\
    \midrule
    Set\_Conv~\cite{pointnet++} & $0.0491$ & $0.7682$ & $0.9408$ & $0.2823$ \\
    GRU~\cite{GRU} &$0.0403$ & $0.8567$ & $0.9635$ & $0.1976$ \\    
    LSTM~\cite{lstm} &$0.0402$ & $0.8578$ & $0.9642$ & $0.1991$ \\
    \bottomrule
    \end{tabular}}
    \caption{Ablation studies of recurrent regularization module. We use GRU~\cite{GRU} as our default recurrent unit because of the simple formulation.}
    \label{tab:ablation_main}
    \vspace{-6mm}
\end{table}

\paragraph{Number of Iterations:} 
During training, the final iteration number of~\equref{point_wise3} and~\equref{recurrent} is 7. While during inference, we vary the iteration number from 3 to 14 to investigate how the performance changes along with the iteration number.
As shown in \tabref{ablation_iter}, for the KITTI~\cite{kitti2015} dataset, all the metrics are gradually improved along with the increasing iteration number. However, for the FlyingThings3D\cite{ft3d} dataset, the performance reaches the peak when the iteration number is exactly the same as training, and gradually degrades afterward. It is because FlyingThings3D~\cite{ft3d} is more challenging because of its larger motion and more complicated occlusion than KITTI~\cite{kitti2015}. Therefore, we set the iteration number to 7 for Flyingthings3D~\cite{ft3d} and 14 for KITTI~\cite{kitti2015} as the default setting.

\vspace{-3mm}
\section{Conclusion}
\vspace{-2mm}

We propose a recurrent framework for the 3D motion estimation. 
To address the irregularity of the point cloud, we optimize the 3D flows on a point-wise cost.
Then a recurrent network globally regularize the flow to output an accurate 3D motion estimation.
We generalize our method to both the 3D scene flow estimation task and the point cloud registration task.
The experiments demonstrate that our method outperforms previous methods
and achieves a new state-of-the-art performance across all metrics.



{\small
\bibliographystyle{ieee_fullname}
\bibliography{reference}
}

\end{document}


\title{Appendix for RCP: Recurrent Closest Point for Point Cloud}

\maketitle

\appendix

\section{More Visualization Results}

\begin{figure*}[bp]
    \centering

    \rule{0.95\textwidth}{0.1pt}
    \vspace{1mm}
 
    \begin{subfigure}{0.48\columnwidth}
        \centering
        \includegraphics[width=1\columnwidth, trim={0cm 0cm 0cm 1cm}, clip]{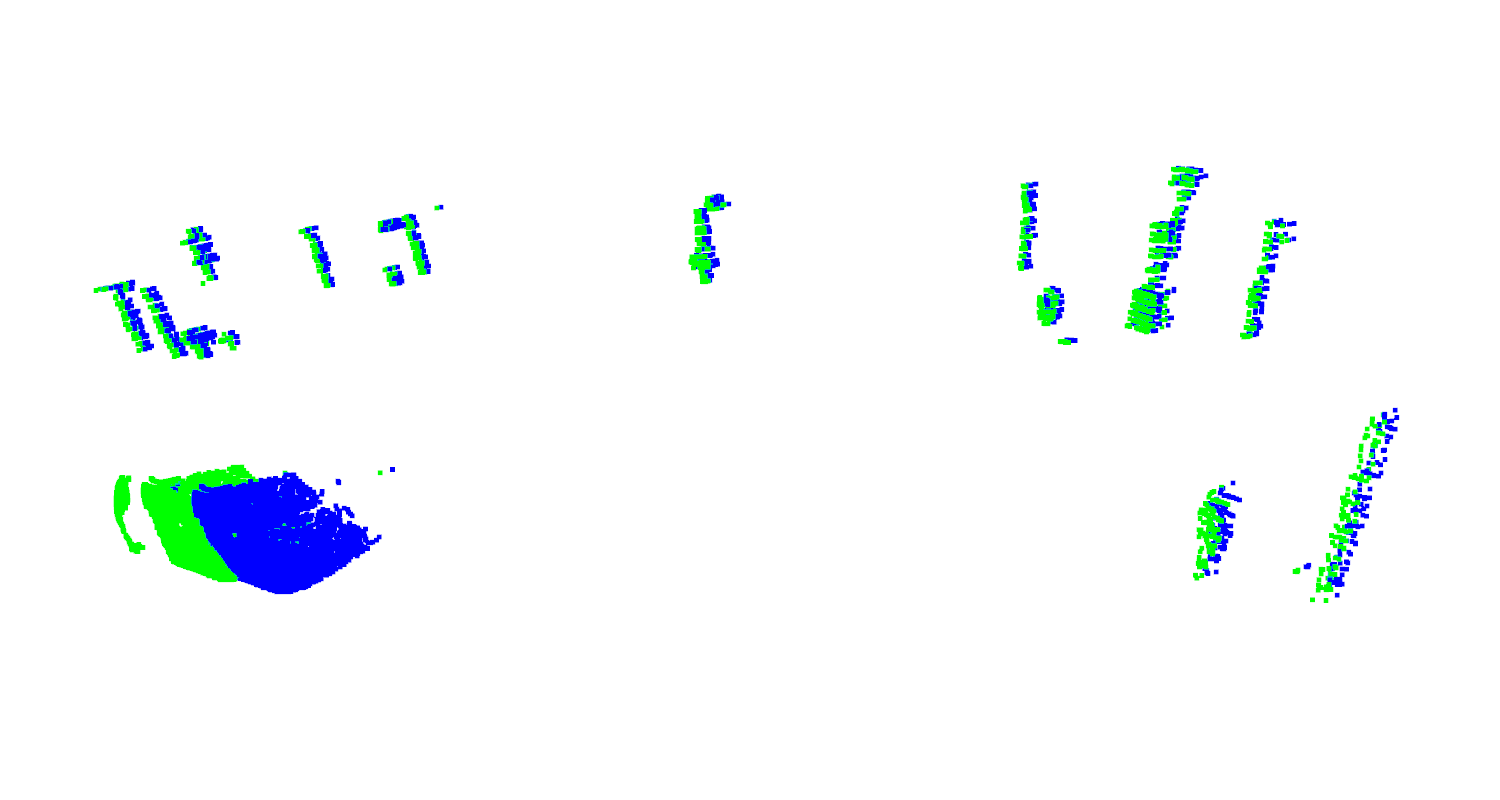}
    \end{subfigure}
    \begin{subfigure}{0.48\columnwidth}
        \centering
        \includegraphics[width=1\columnwidth, trim={0cm 0cm 0cm 1cm}, clip]{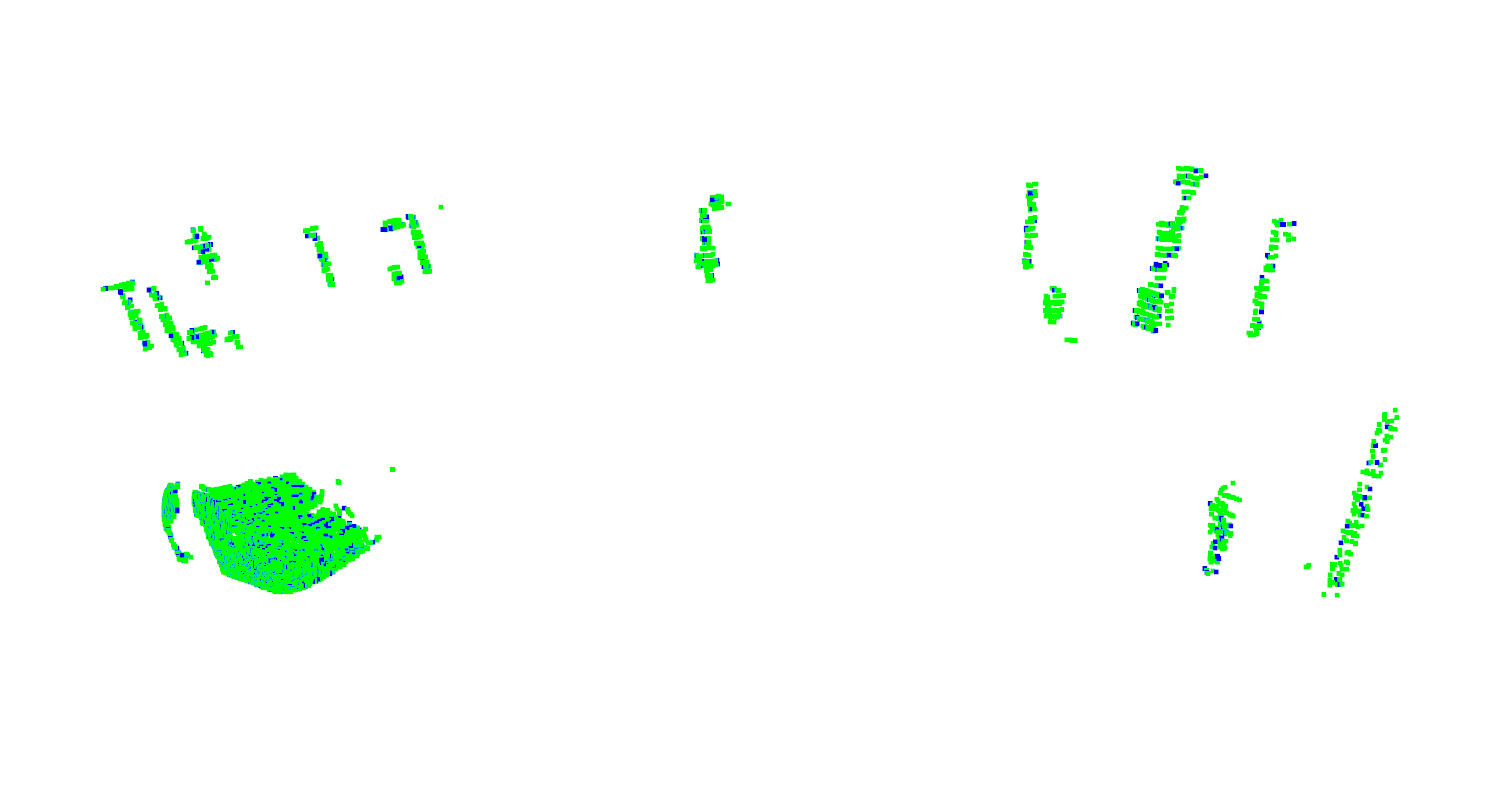}
    \end{subfigure}
    \begin{subfigure}{0.48\columnwidth}
        \centering
        \includegraphics[width=1\columnwidth, trim={0cm 0cm 0cm 1cm}, clip]{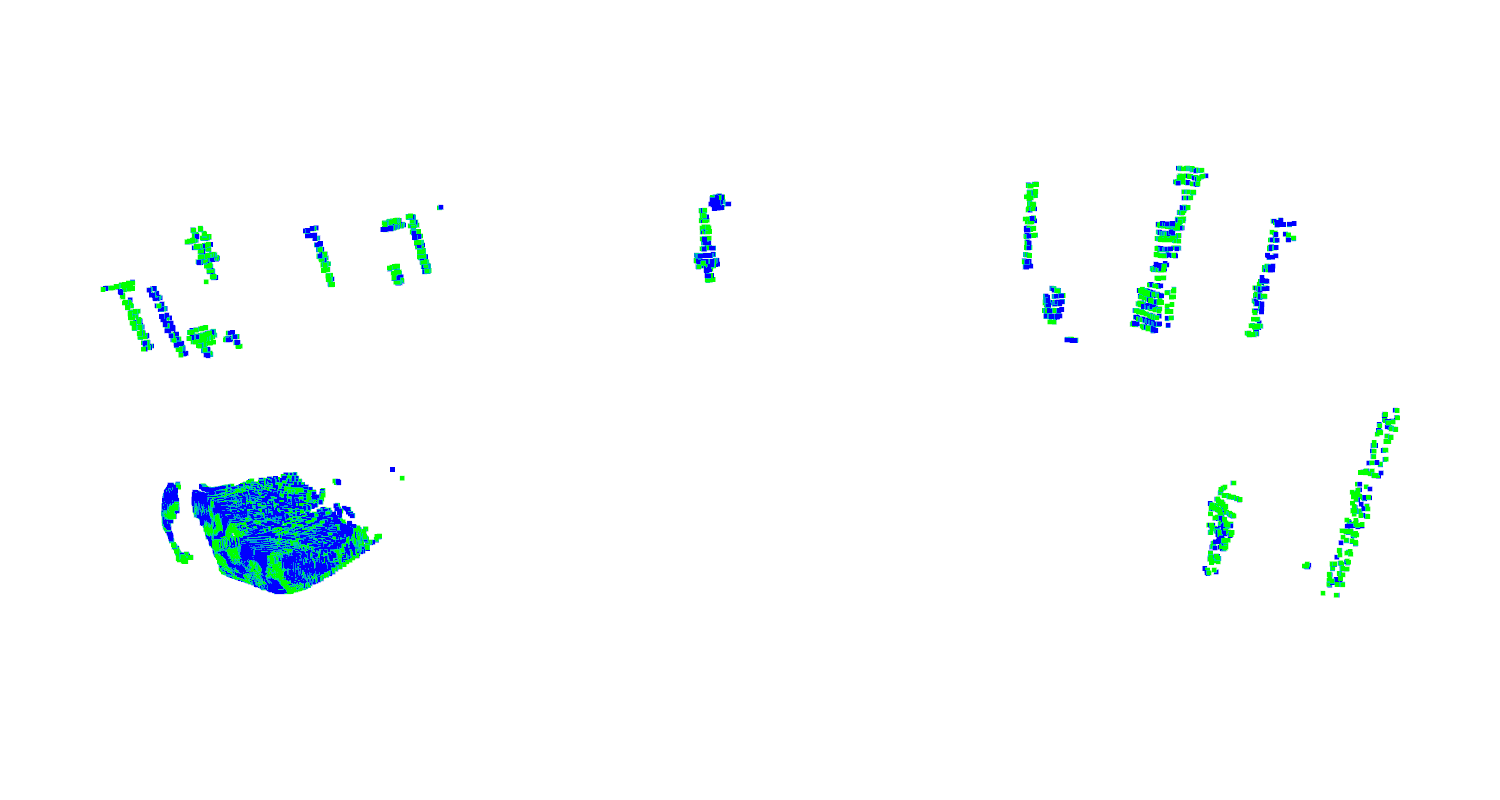}
    \end{subfigure}
    \begin{subfigure}{0.48\columnwidth}
        \centering
        \includegraphics[width=1\columnwidth, trim={0cm 0cm 0cm 1cm}, clip]{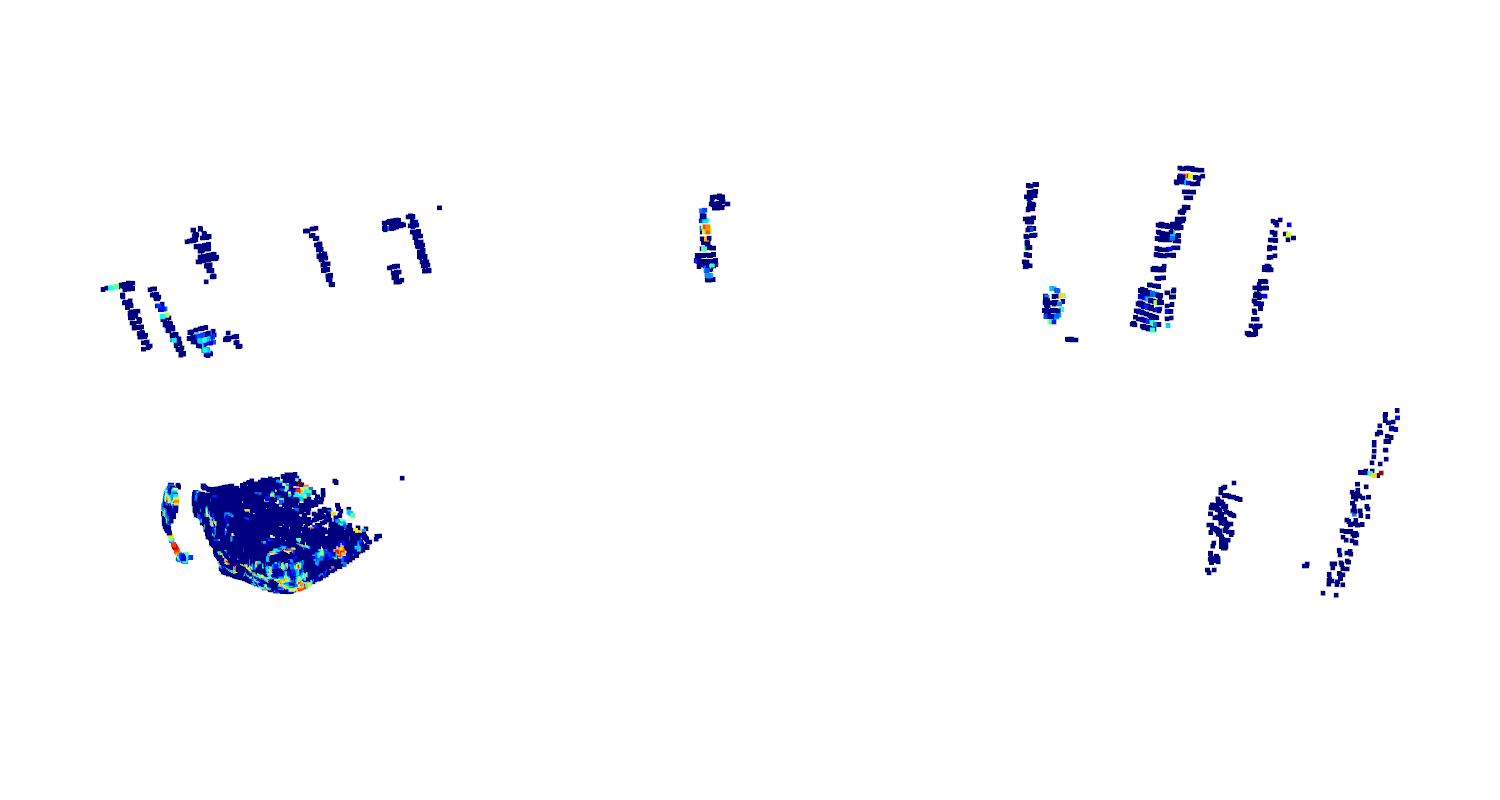}
    \end{subfigure}
    
    \rule{0.95\textwidth}{0.1pt}

    \begin{subfigure}{0.48\columnwidth}
        \centering
        \includegraphics[width=1\columnwidth, trim={0cm 0cm 0cm 1cm}, clip]{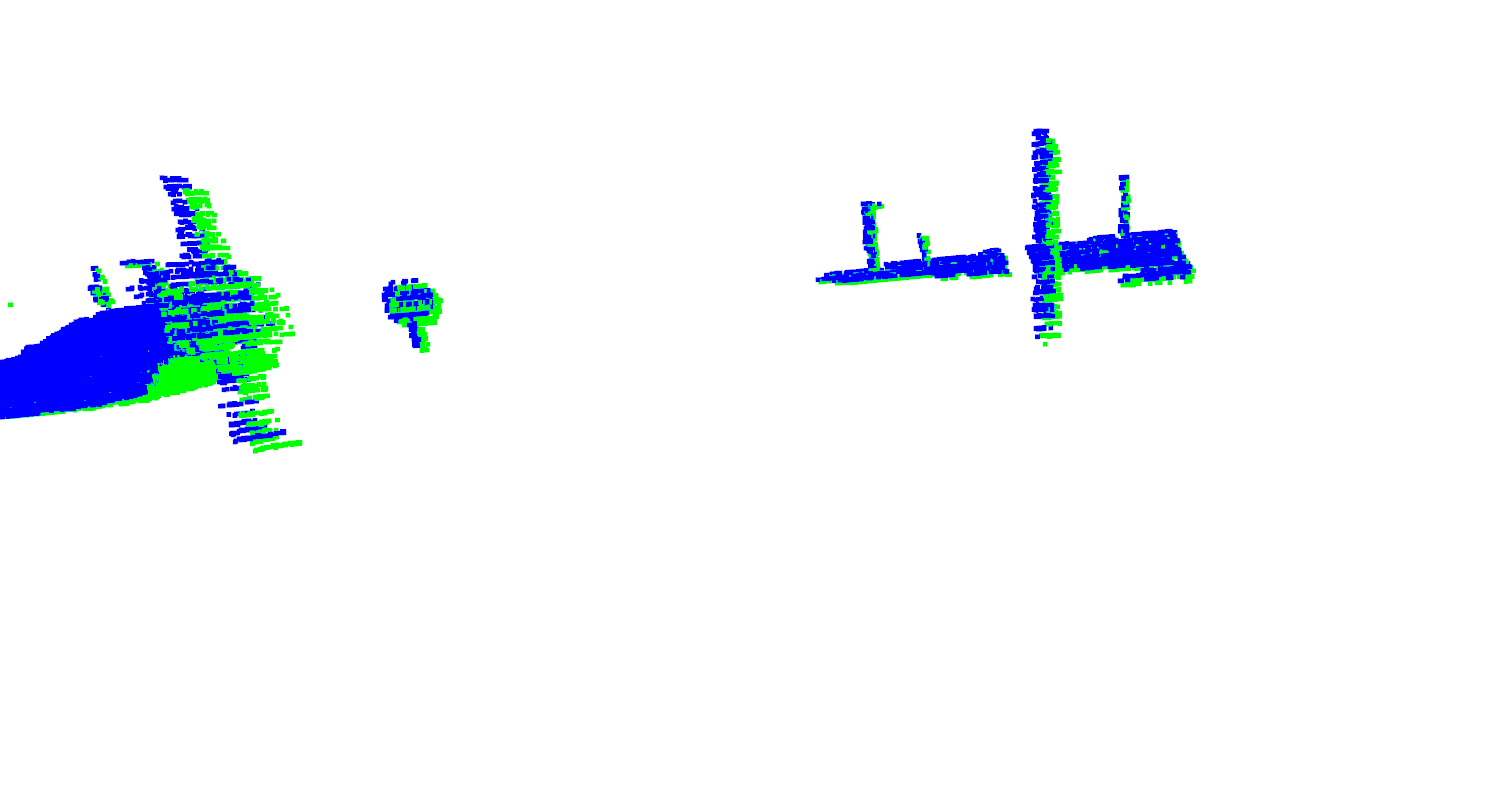}
    \end{subfigure}
    \begin{subfigure}{0.48\columnwidth}
        \centering
        \includegraphics[width=1\columnwidth, trim={0cm 0cm 0cm 1cm}, clip]{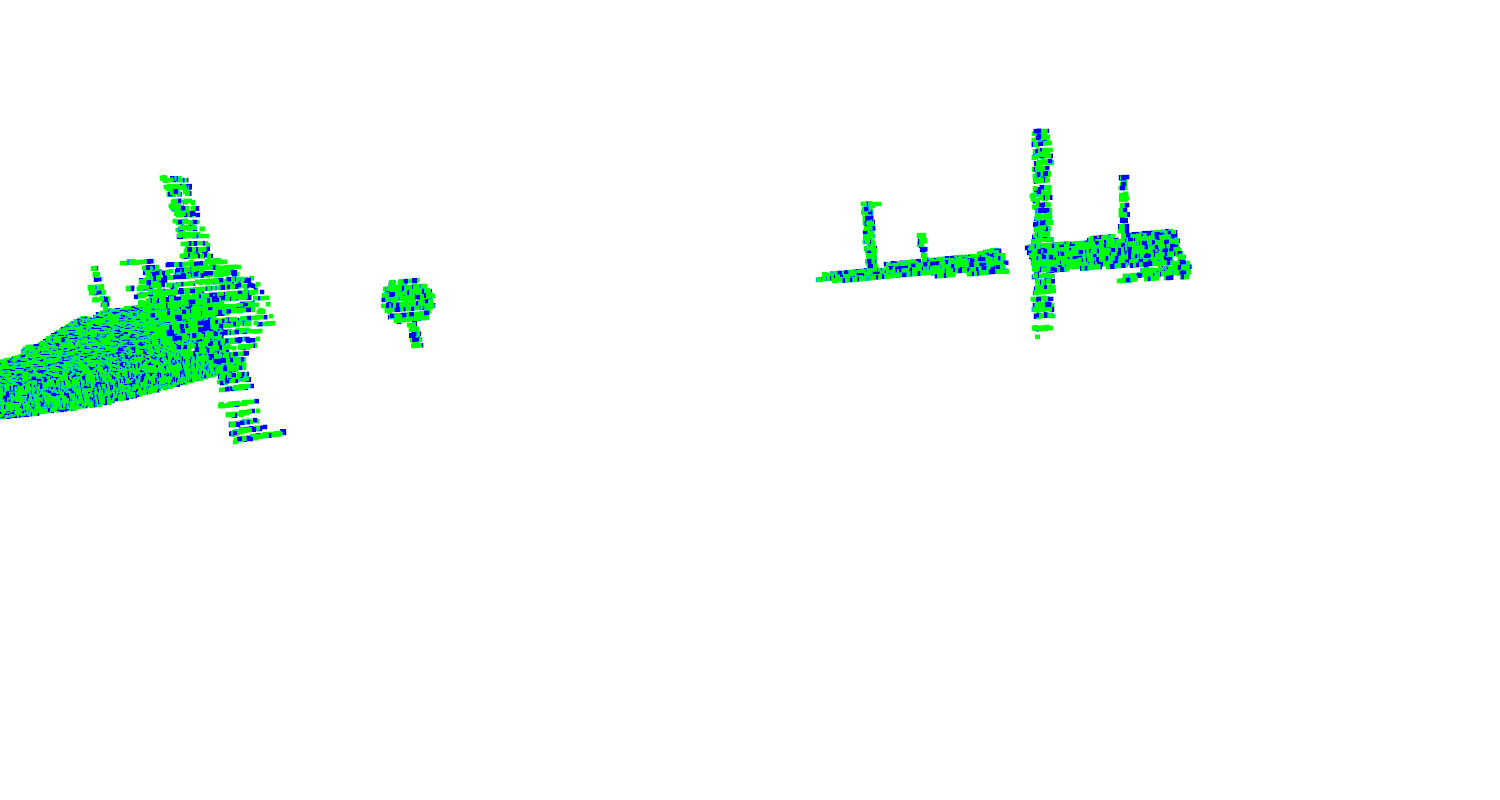}
    \end{subfigure}
    \begin{subfigure}{0.48\columnwidth}
        \centering
        \includegraphics[width=1\columnwidth, trim={0cm 0cm 0cm 1cm}, clip]{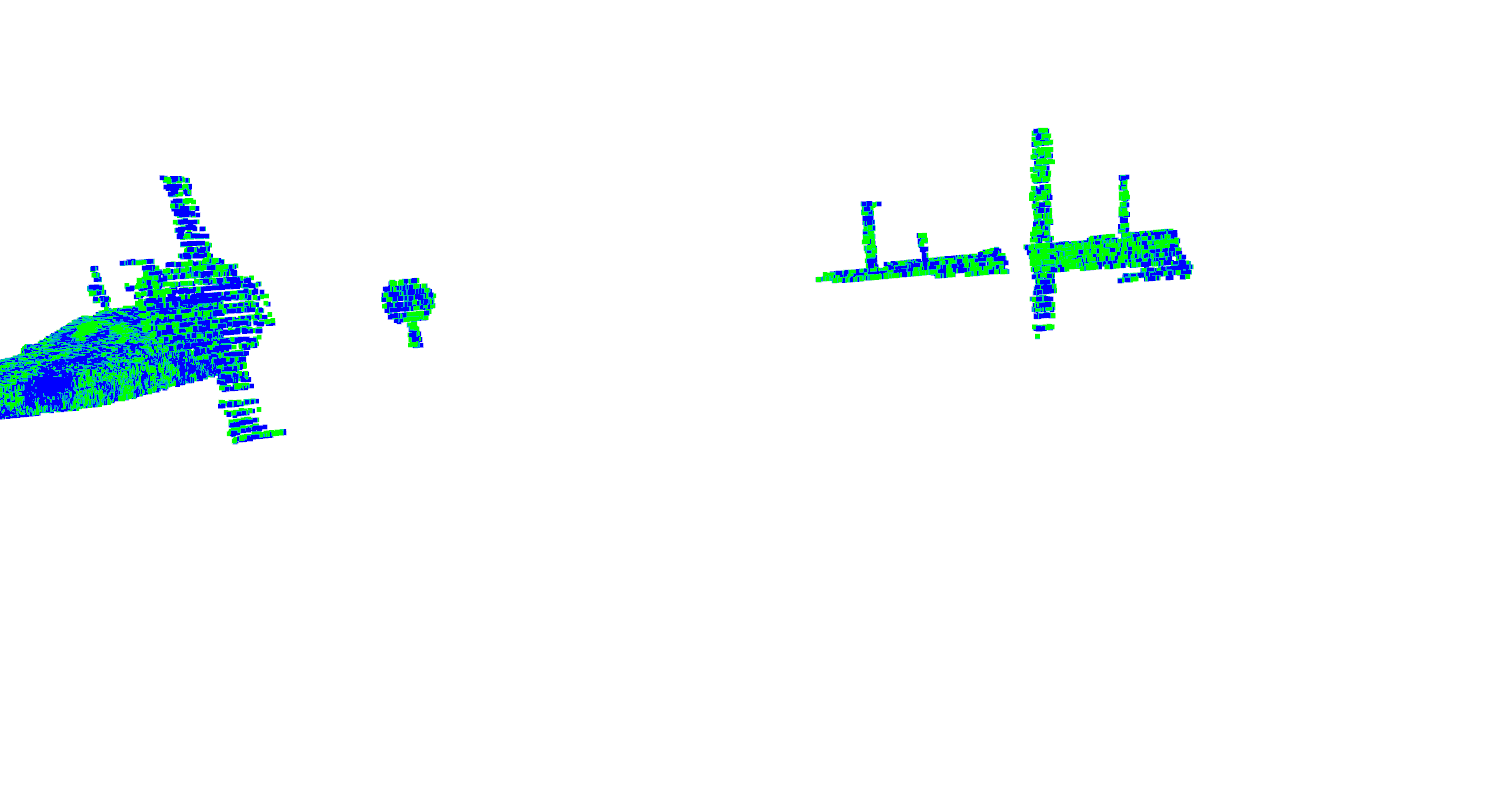}
    \end{subfigure}
    \begin{subfigure}{0.48\columnwidth}
        \centering
        \includegraphics[width=1\columnwidth, trim={0cm 0cm 0cm 1cm}, clip]{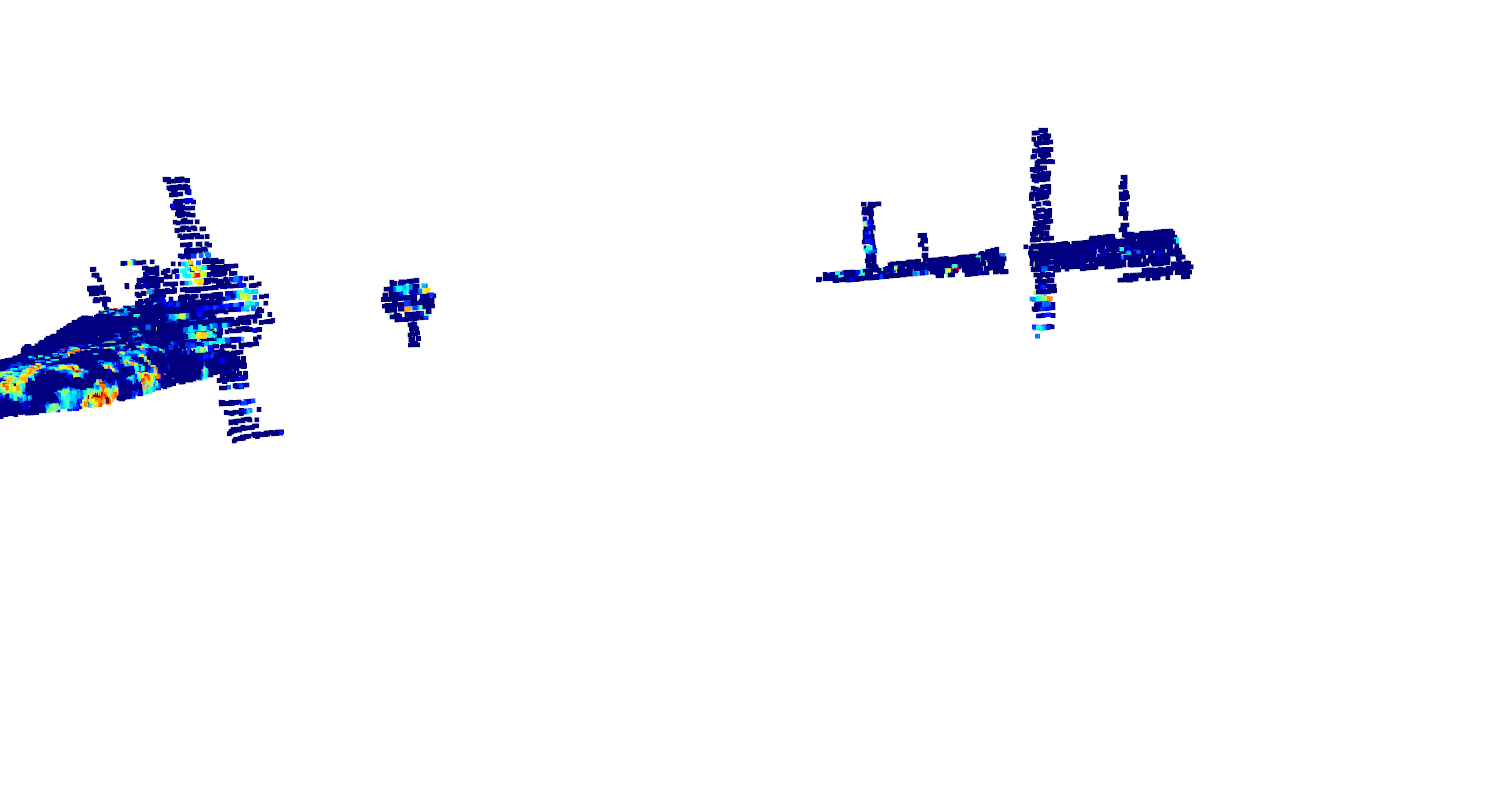}
    \end{subfigure}

    \rule{0.95\textwidth}{0.1pt}

    \begin{subfigure}{0.48\columnwidth}
        \centering
        \includegraphics[width=1\columnwidth, trim={0cm 0cm 0cm 1cm}, clip]{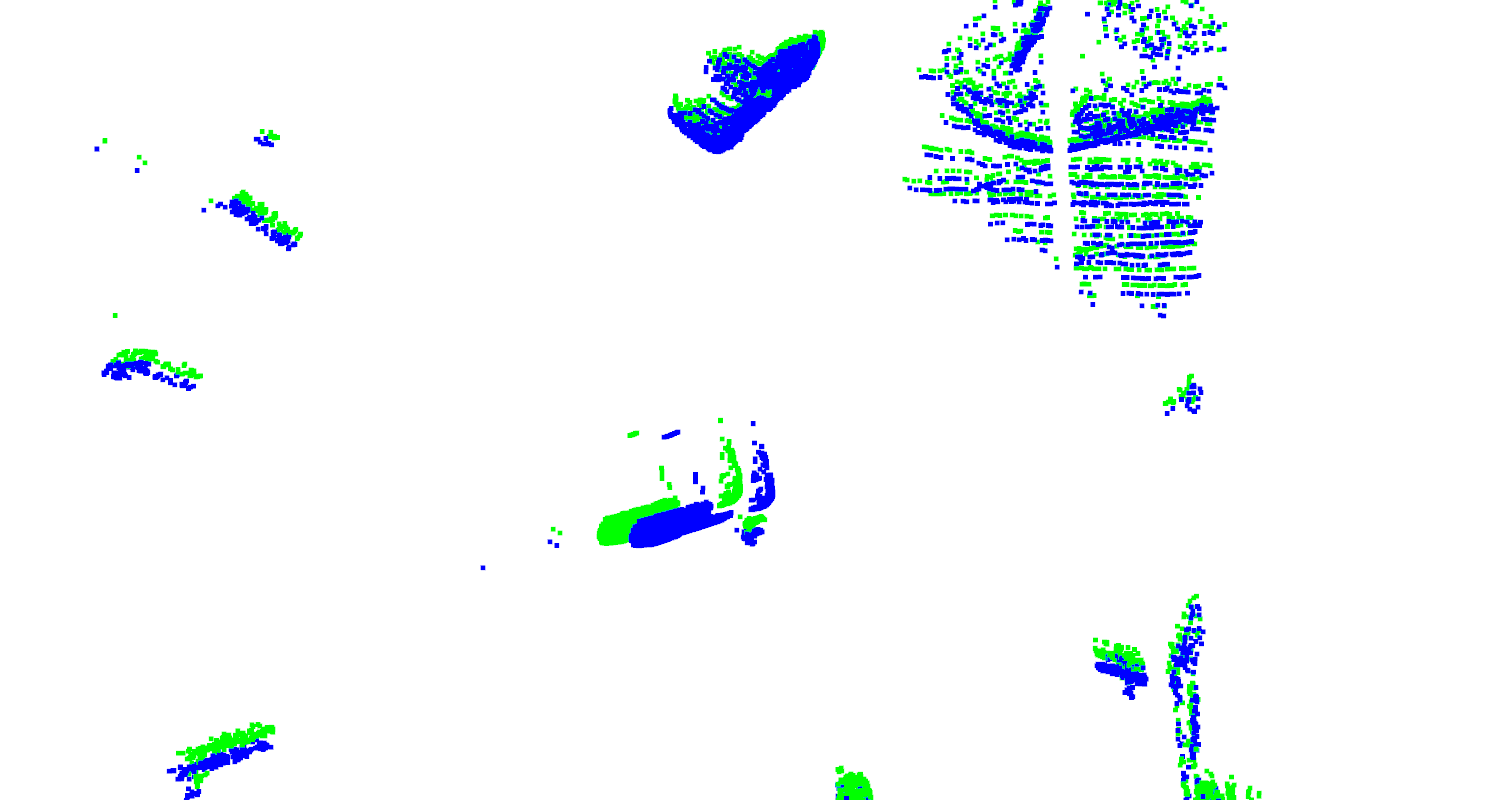}
    \end{subfigure}
    \begin{subfigure}{0.48\columnwidth}
        \centering
        \includegraphics[width=1\columnwidth, trim={0cm 0cm 0cm 1cm}, clip]{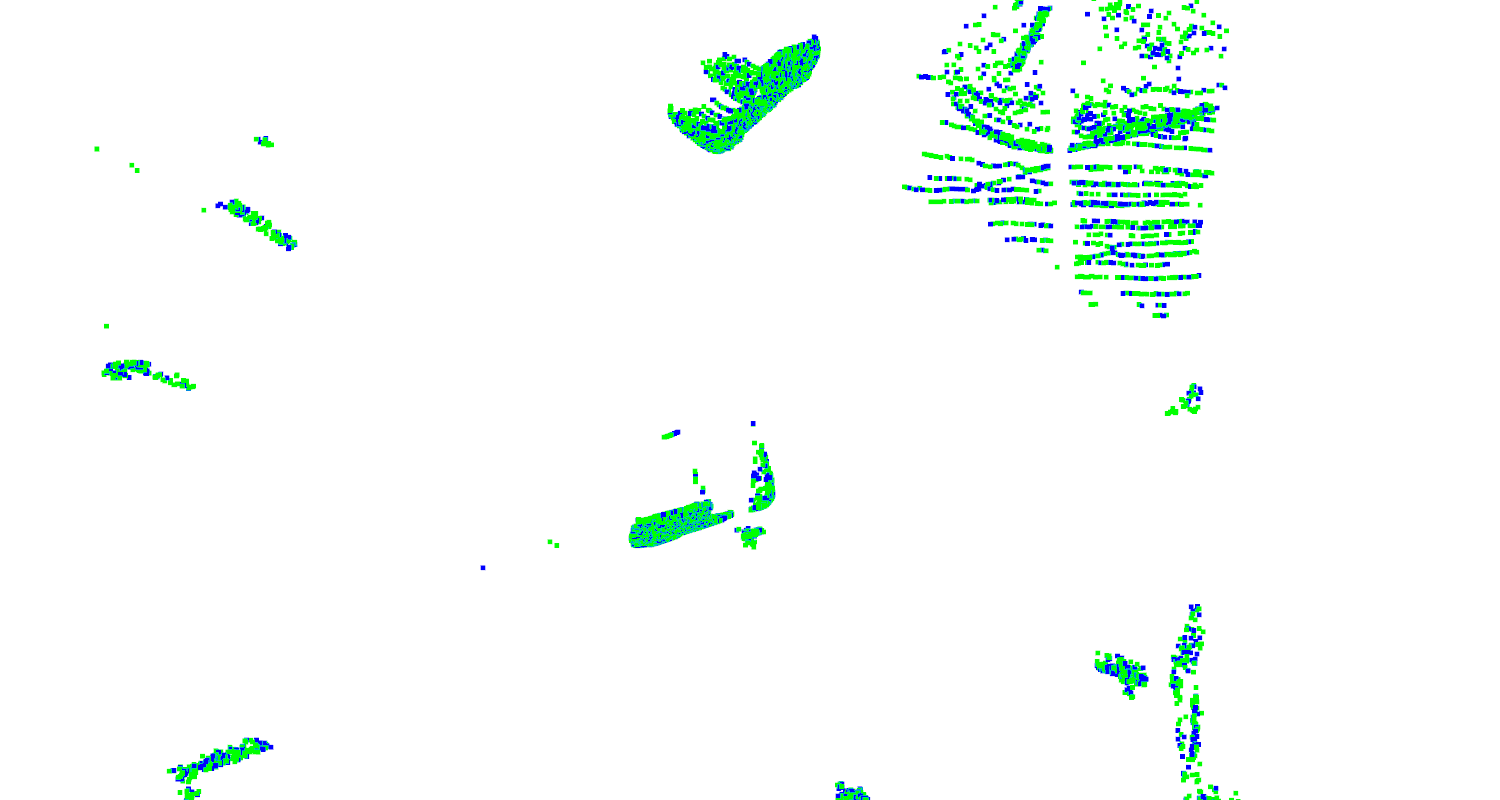}
    \end{subfigure}
    \begin{subfigure}{0.48\columnwidth}
        \centering
        \includegraphics[width=1\columnwidth, trim={0cm 0cm 0cm 1cm}, clip]{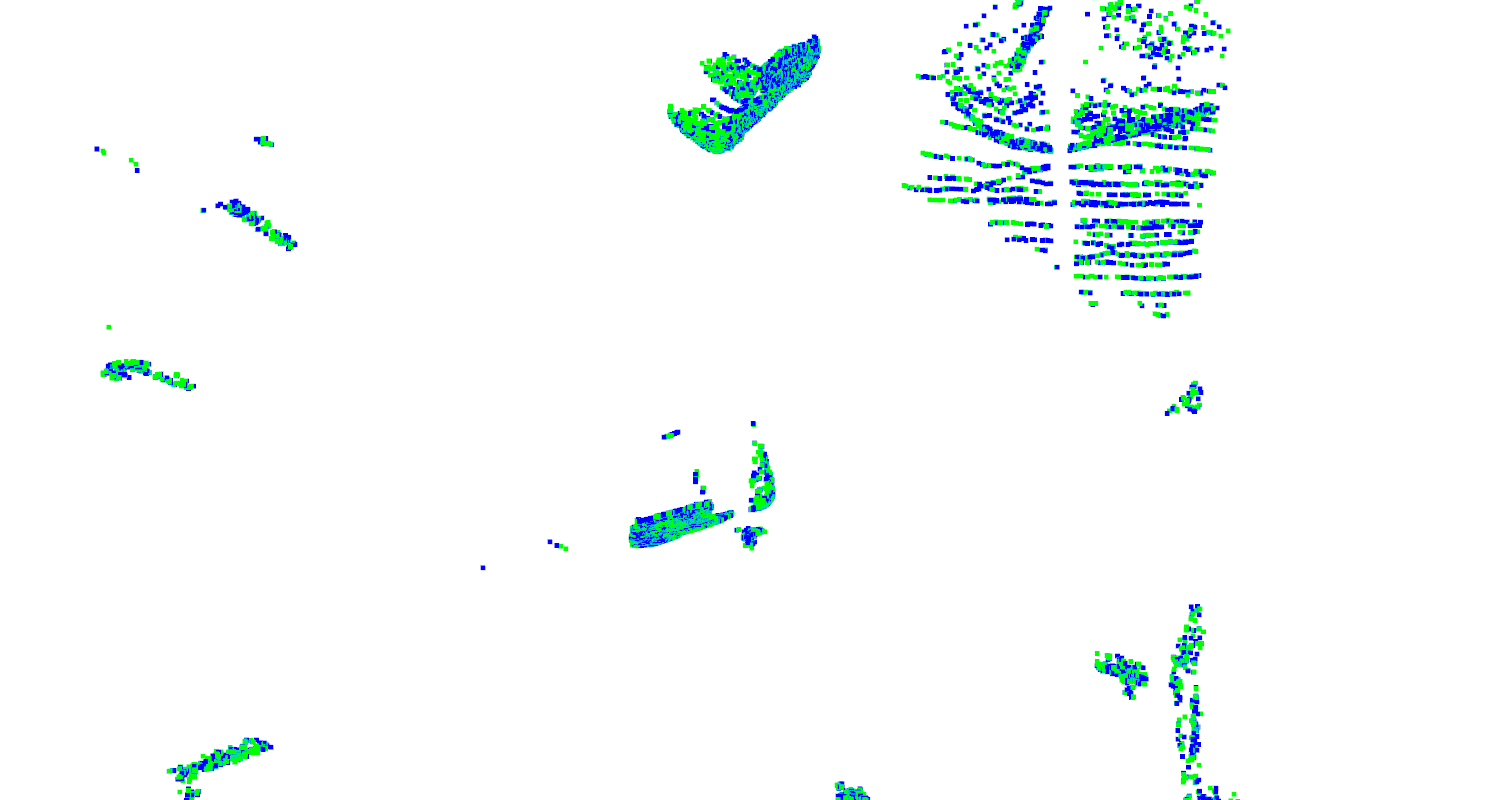}
    \end{subfigure}
    \begin{subfigure}{0.48\columnwidth}
        \centering
        \includegraphics[width=1\columnwidth, trim={0cm 0cm 0cm 1cm}, clip]{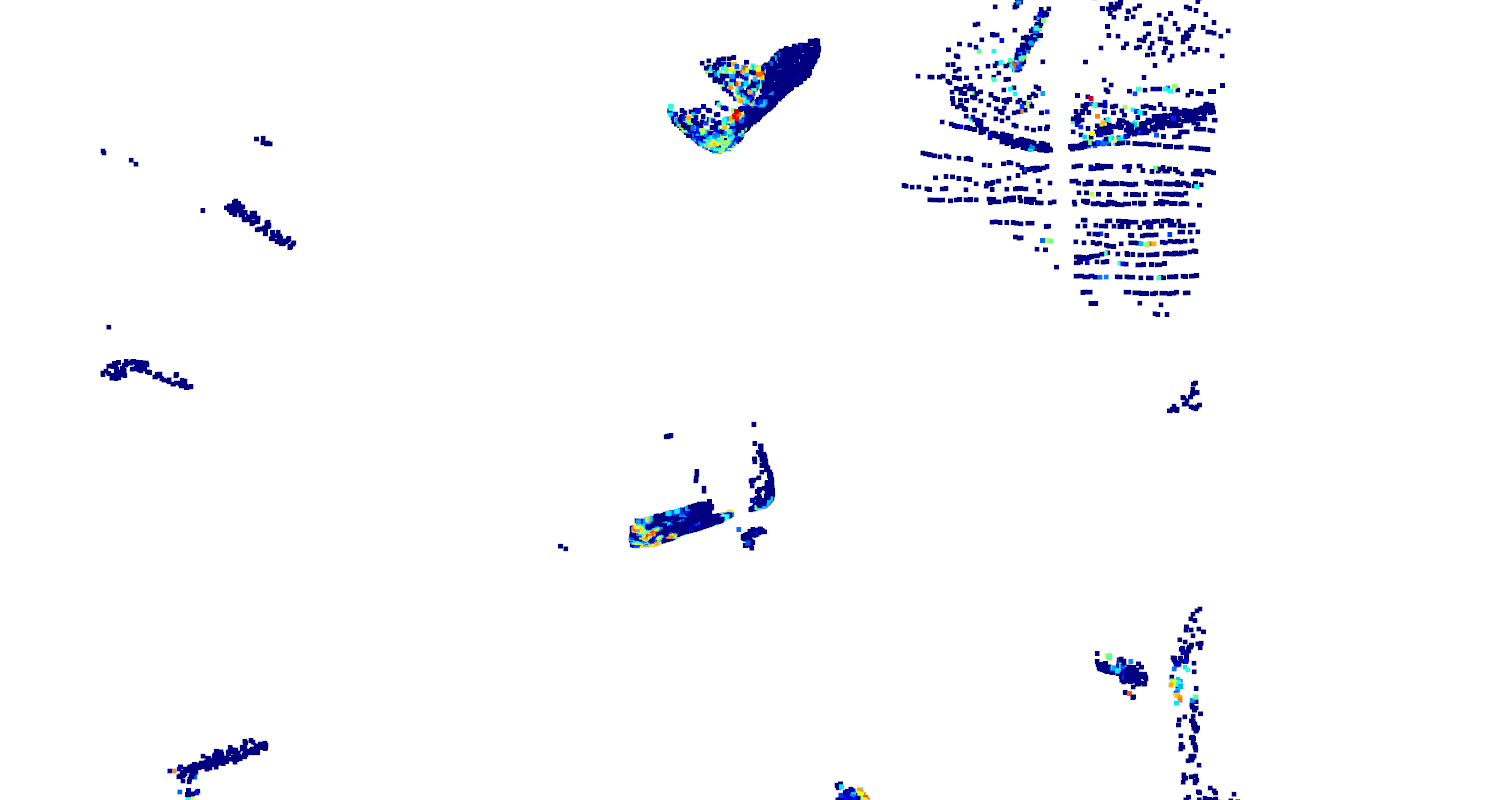}
    \end{subfigure}

    \rule{0.95\textwidth}{0.1pt}

    \begin{subfigure}{0.48\columnwidth}
        \centering
        \includegraphics[width=1\columnwidth, trim={0cm 0cm 0cm 1cm}, clip]{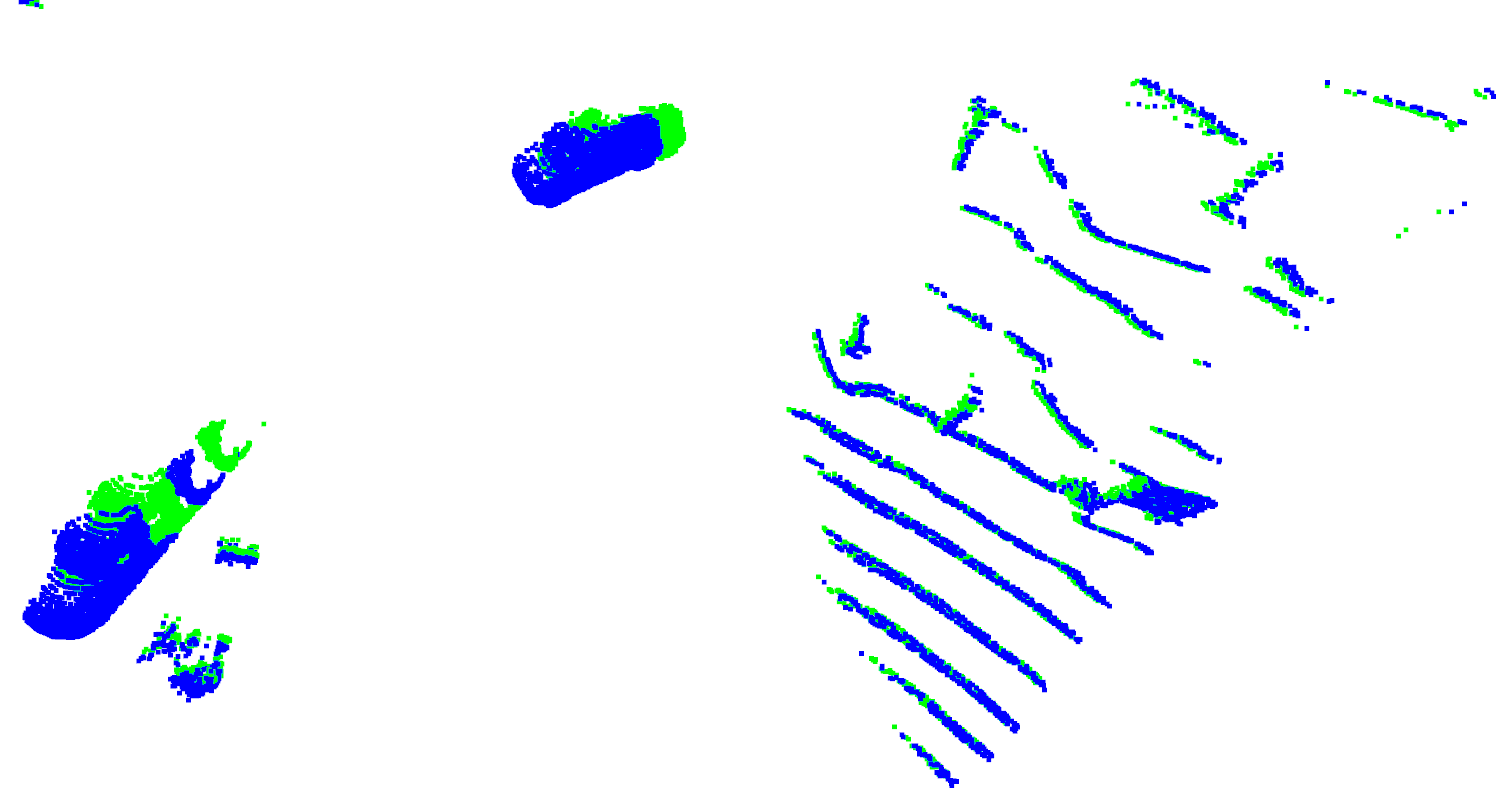}
    \end{subfigure}
    \begin{subfigure}{0.48\columnwidth}
        \centering
        \includegraphics[width=1\columnwidth, trim={0cm 0cm 0cm 1cm}, clip]{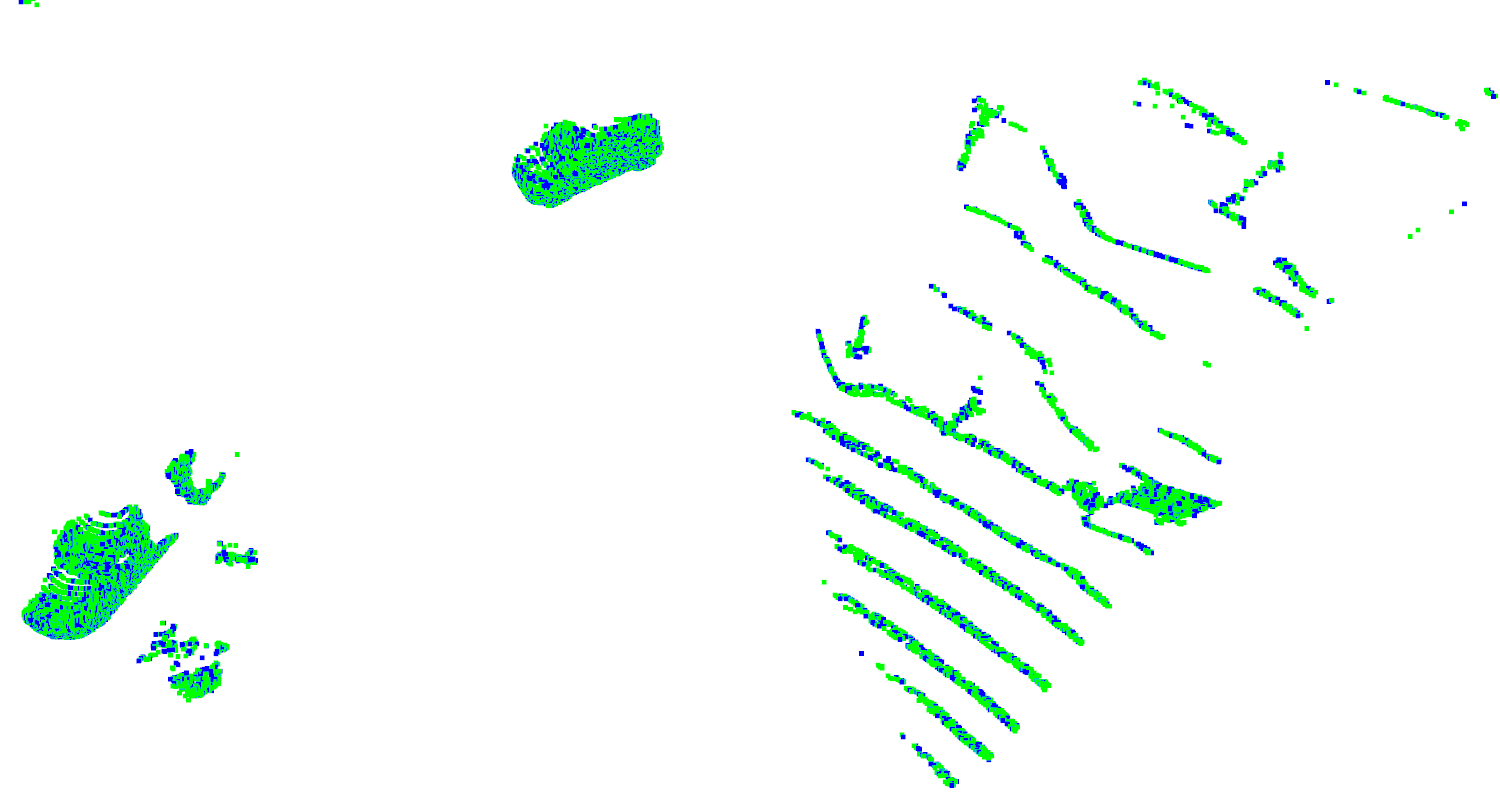}
    \end{subfigure}
    \begin{subfigure}{0.48\columnwidth}
        \centering
        \includegraphics[width=1\columnwidth, trim={0cm 0cm 0cm 1cm}, clip]{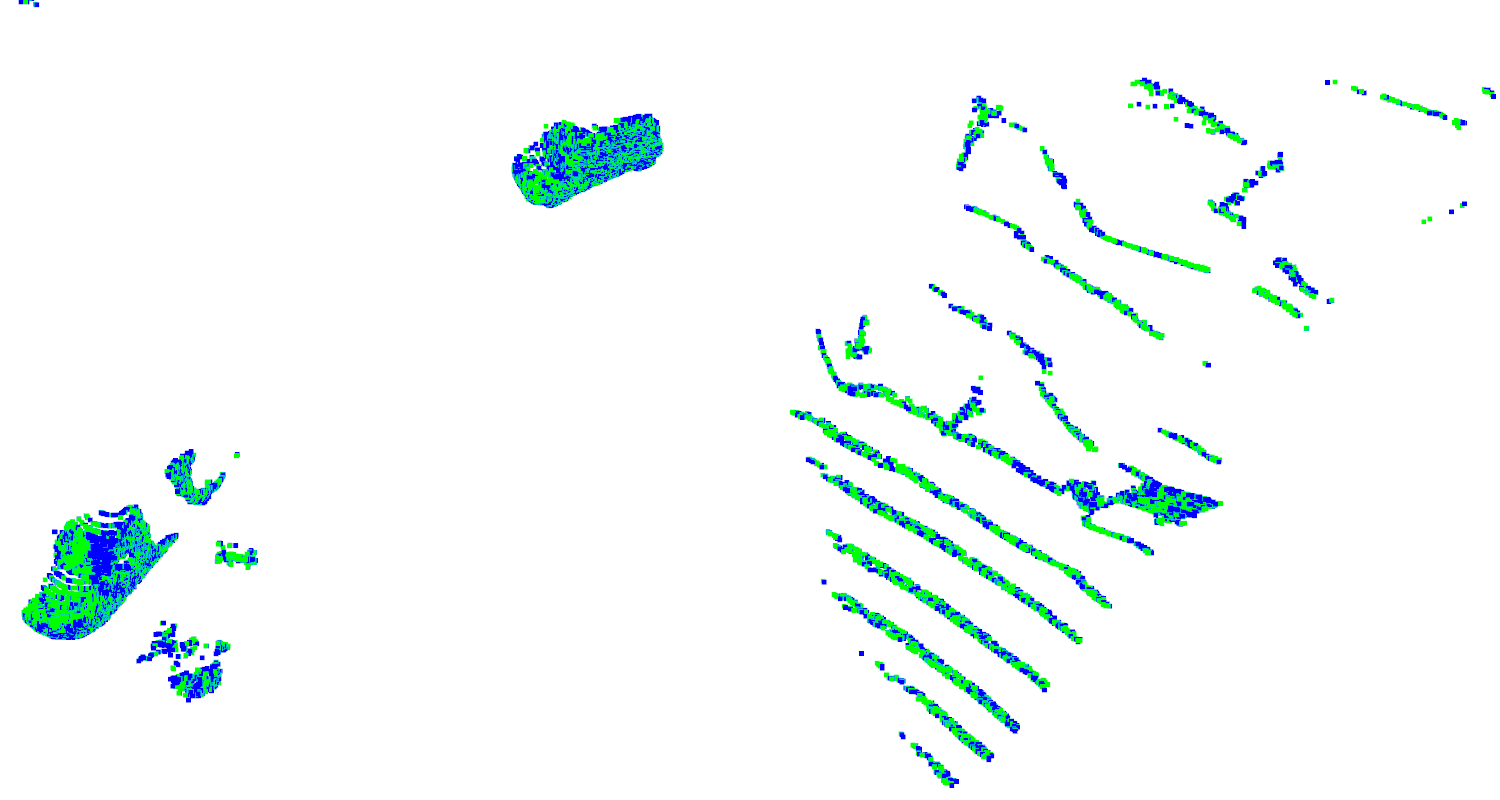}
    \end{subfigure}
    \begin{subfigure}{0.48\columnwidth}
        \centering
        \includegraphics[width=1\columnwidth, trim={0cm 0cm 0cm 1cm}, clip]{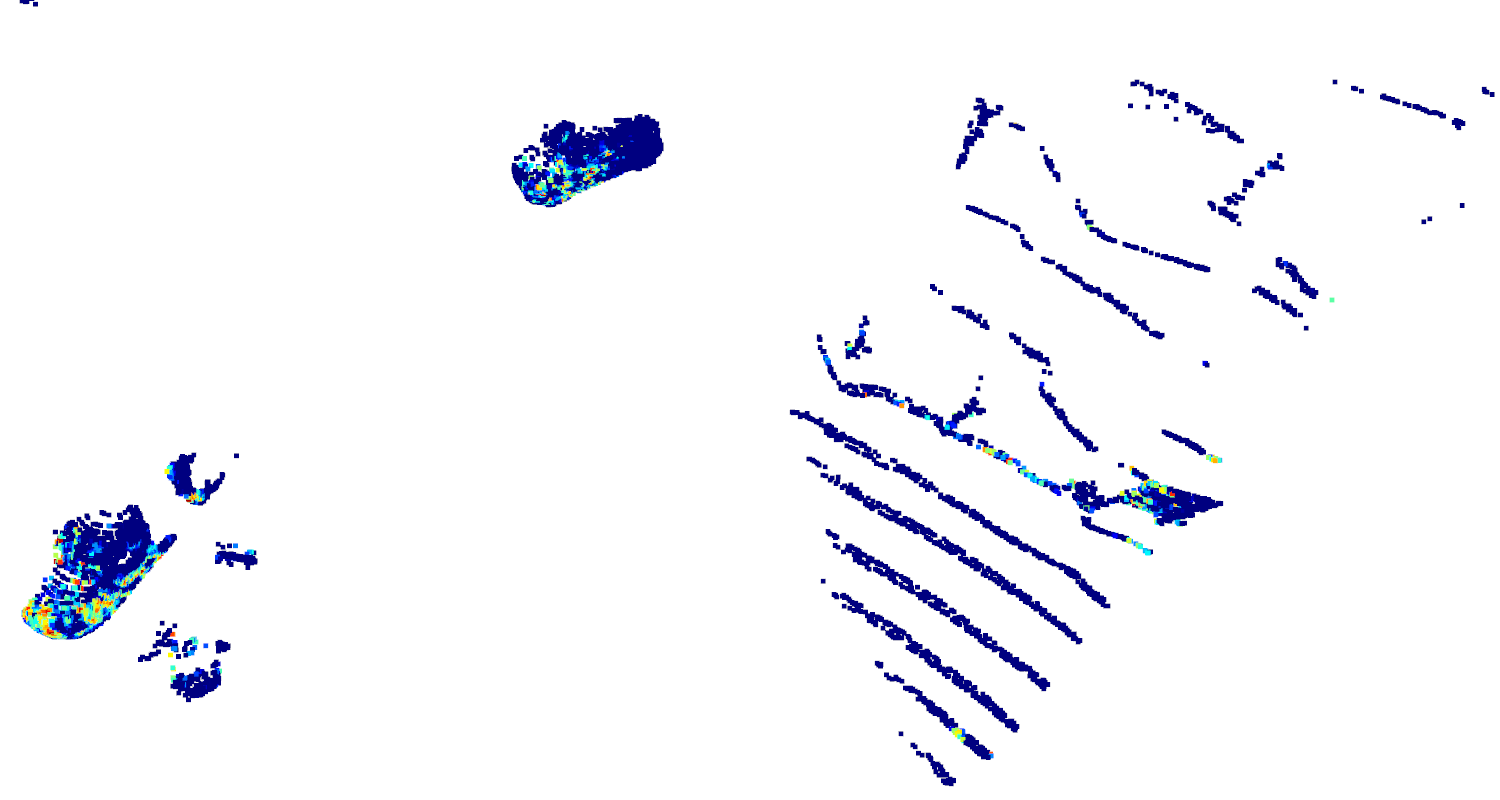}
    \end{subfigure}



    \rule{0.95\textwidth}{0.1pt}

    \begin{subfigure}{0.48\columnwidth}
        \centering
        \includegraphics[width=1\columnwidth, trim={0cm 0cm 0cm 1cm}, clip]{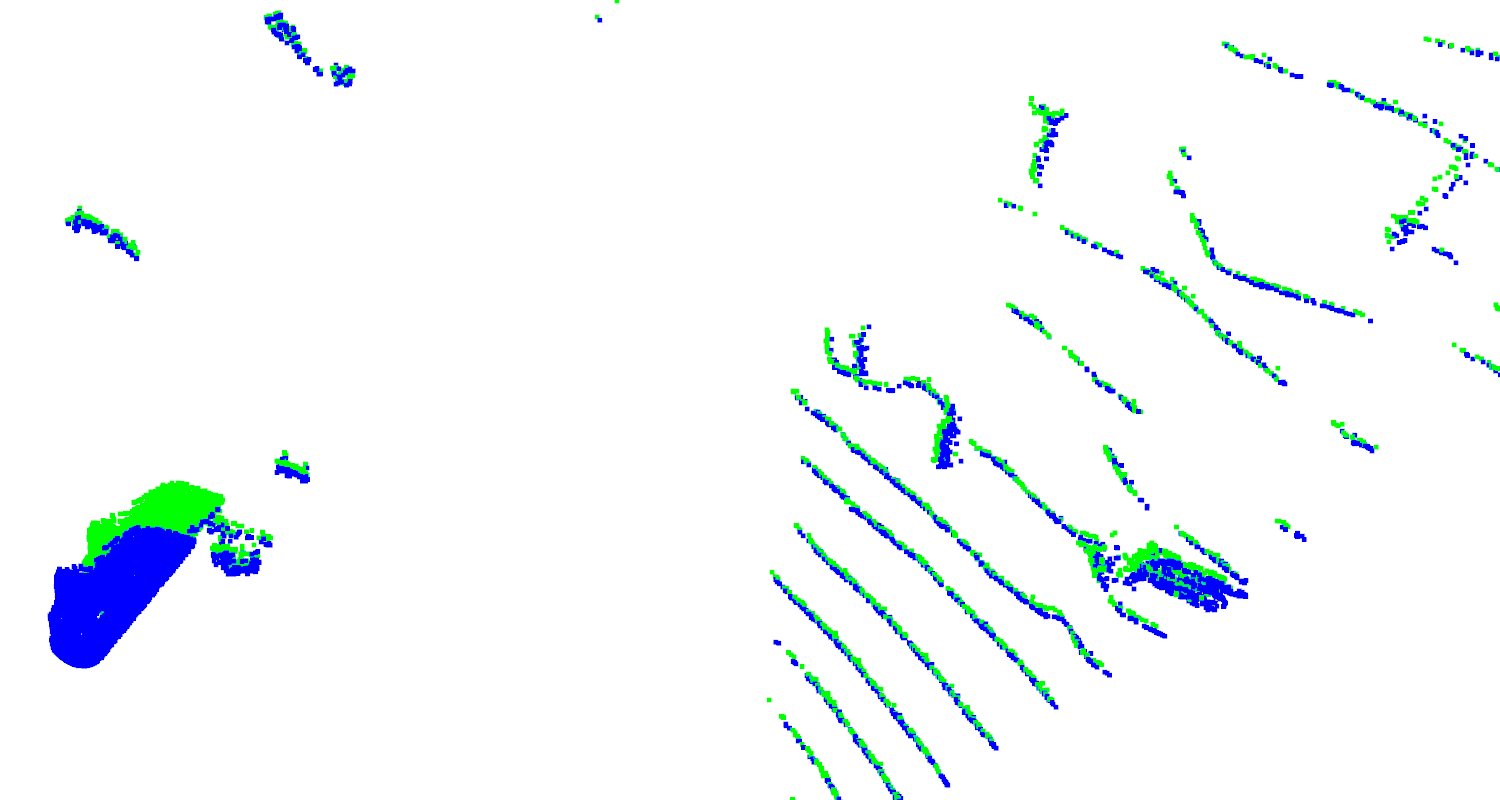}
    \end{subfigure}
    \begin{subfigure}{0.48\columnwidth}
        \centering
        \includegraphics[width=1\columnwidth, trim={0cm 0cm 0cm 1cm}, clip]{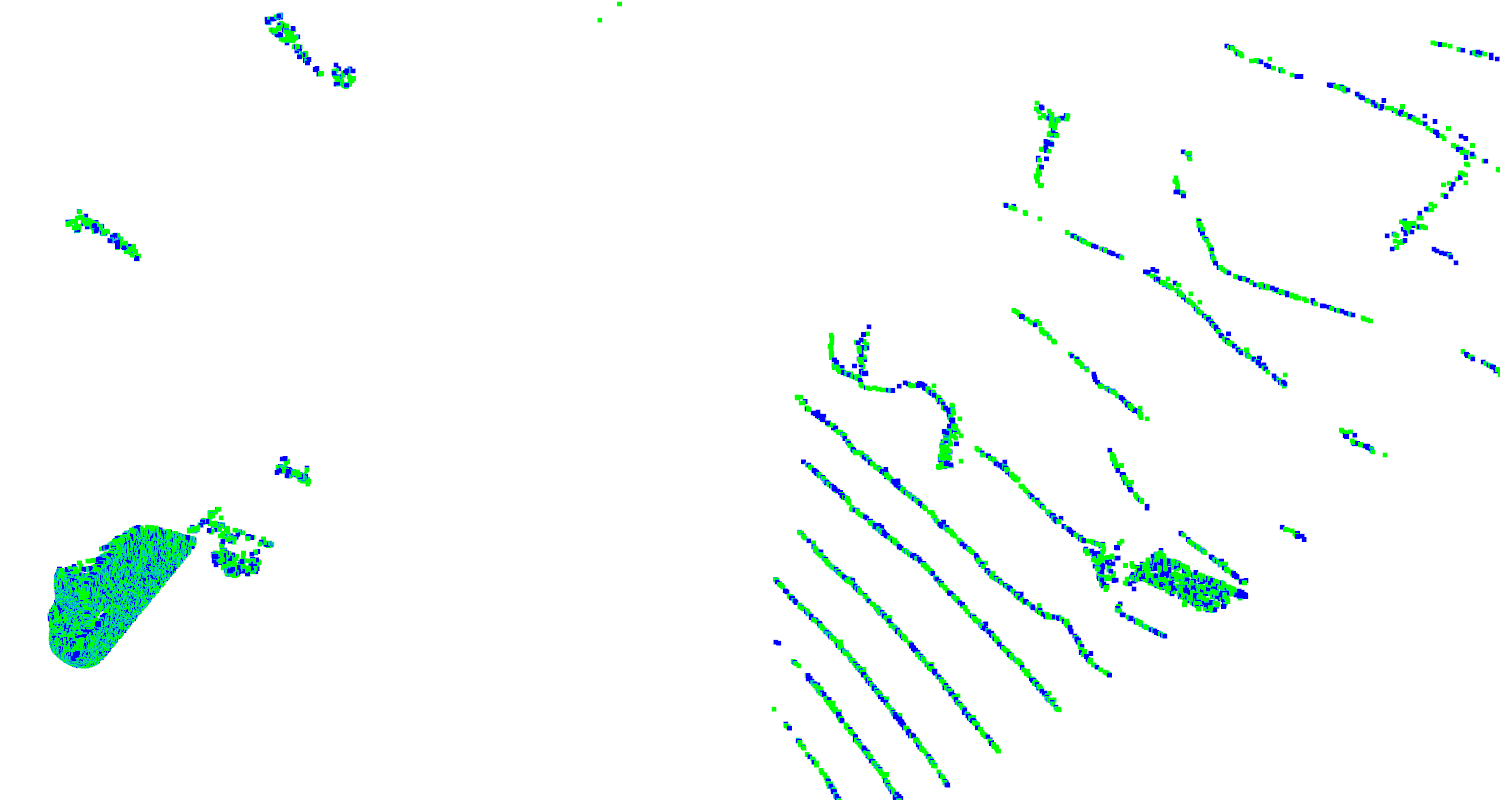}
    \end{subfigure}
    \begin{subfigure}{0.48\columnwidth}
        \centering
        \includegraphics[width=1\columnwidth, trim={0cm 0cm 0cm 1cm}, clip]{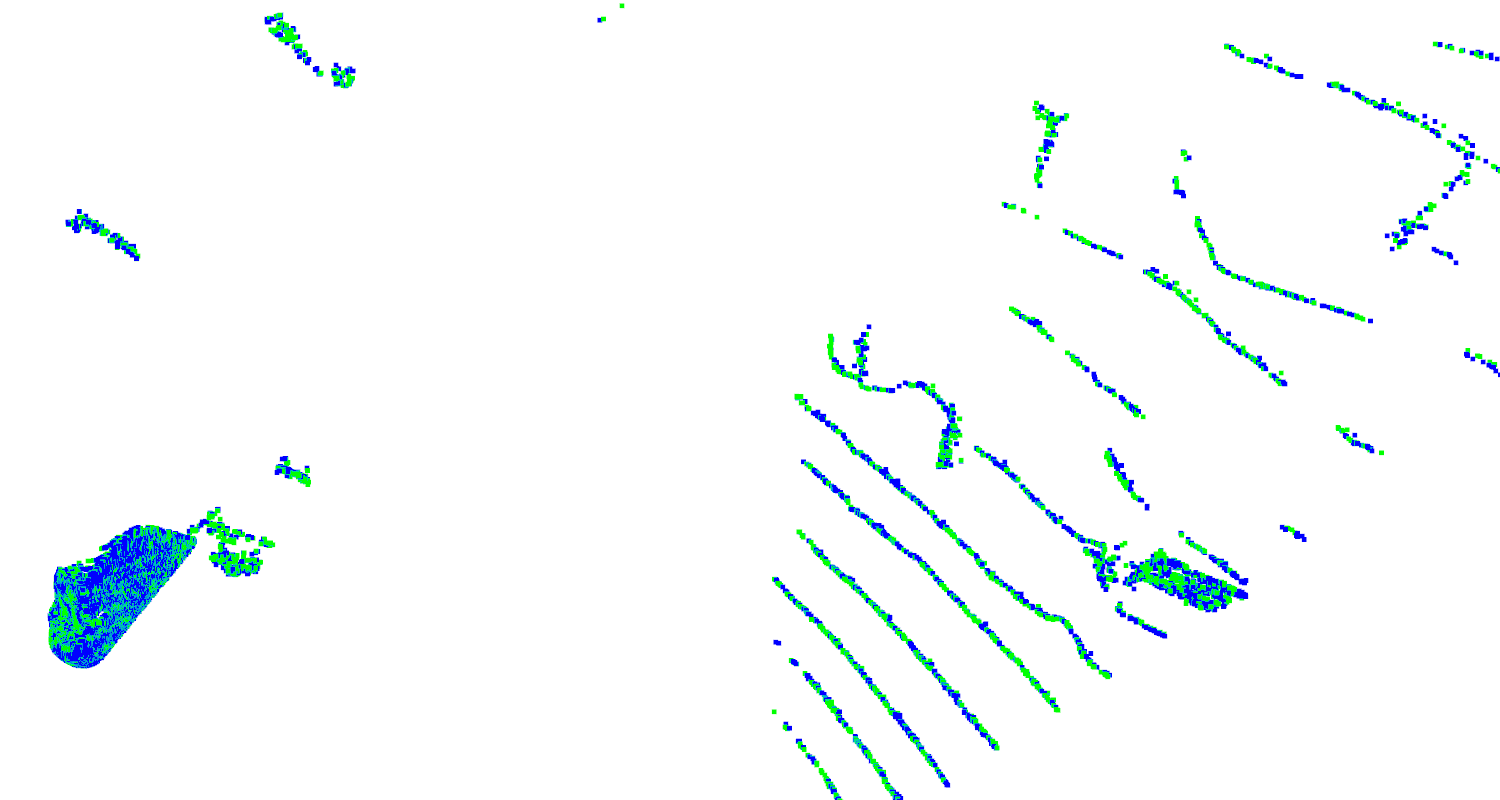}
    \end{subfigure}
    \begin{subfigure}{0.48\columnwidth}
        \centering
        \includegraphics[width=1\columnwidth, trim={0cm 0cm 0cm 1cm}, clip]{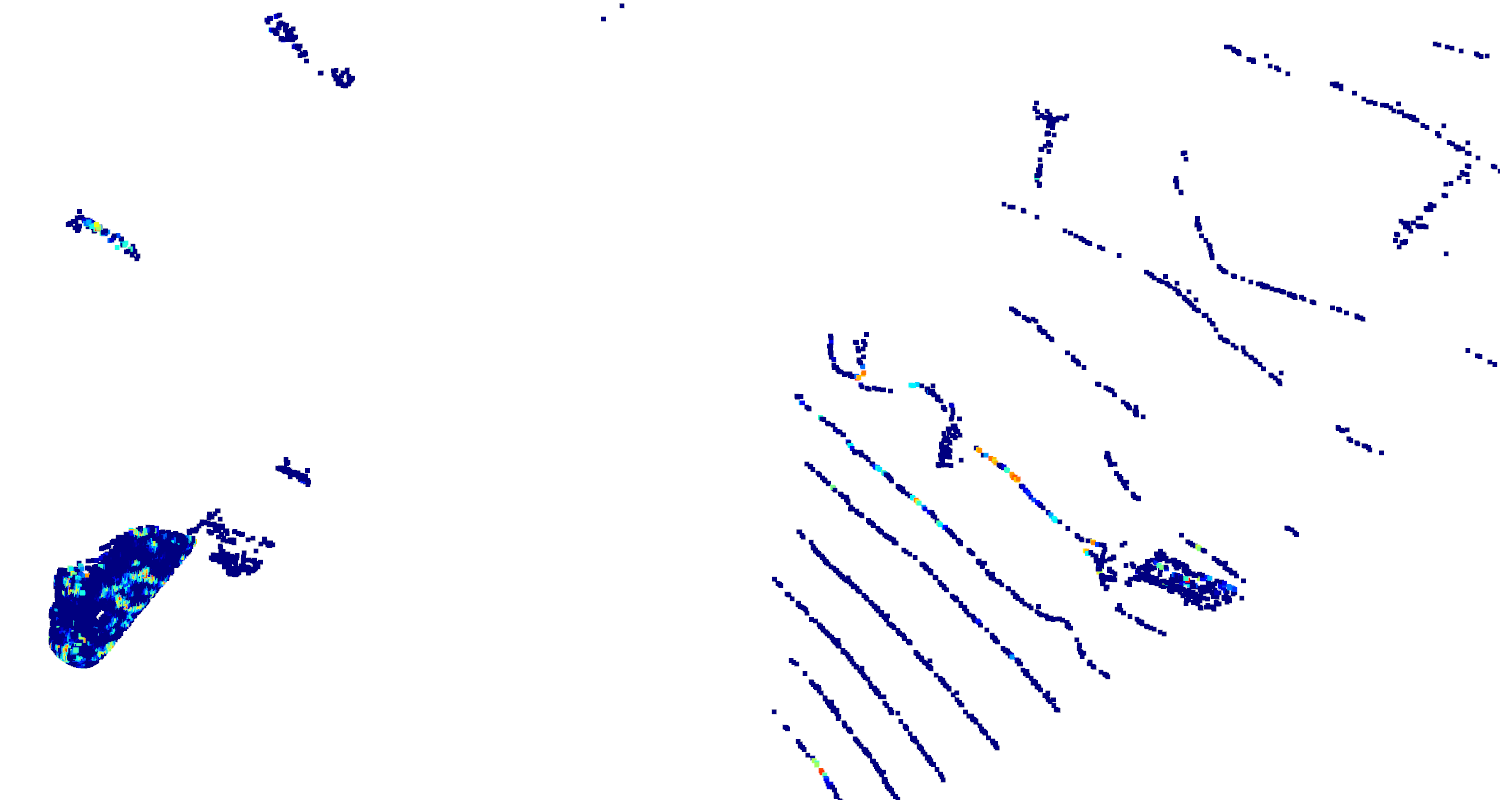}
    \end{subfigure}

    

    \rule{0.95\textwidth}{0.1pt}

    \begin{subfigure}{0.48\columnwidth}
        \centering
        \includegraphics[width=1\columnwidth, trim={0cm 0cm 0cm 1cm}, clip]{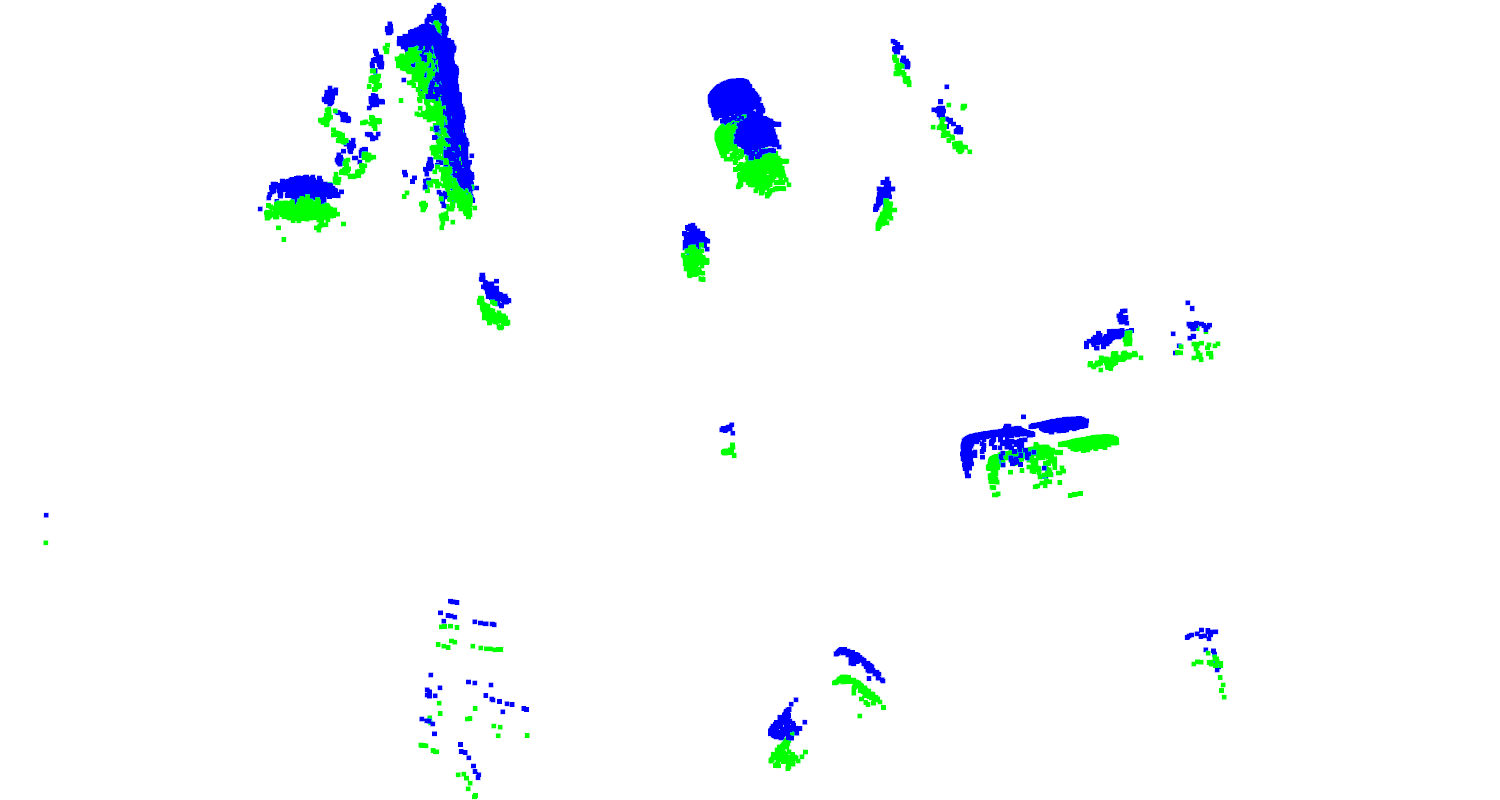}
    \end{subfigure}
    \begin{subfigure}{0.48\columnwidth}
        \centering
        \includegraphics[width=1\columnwidth, trim={0cm 0cm 0cm 1cm}, clip]{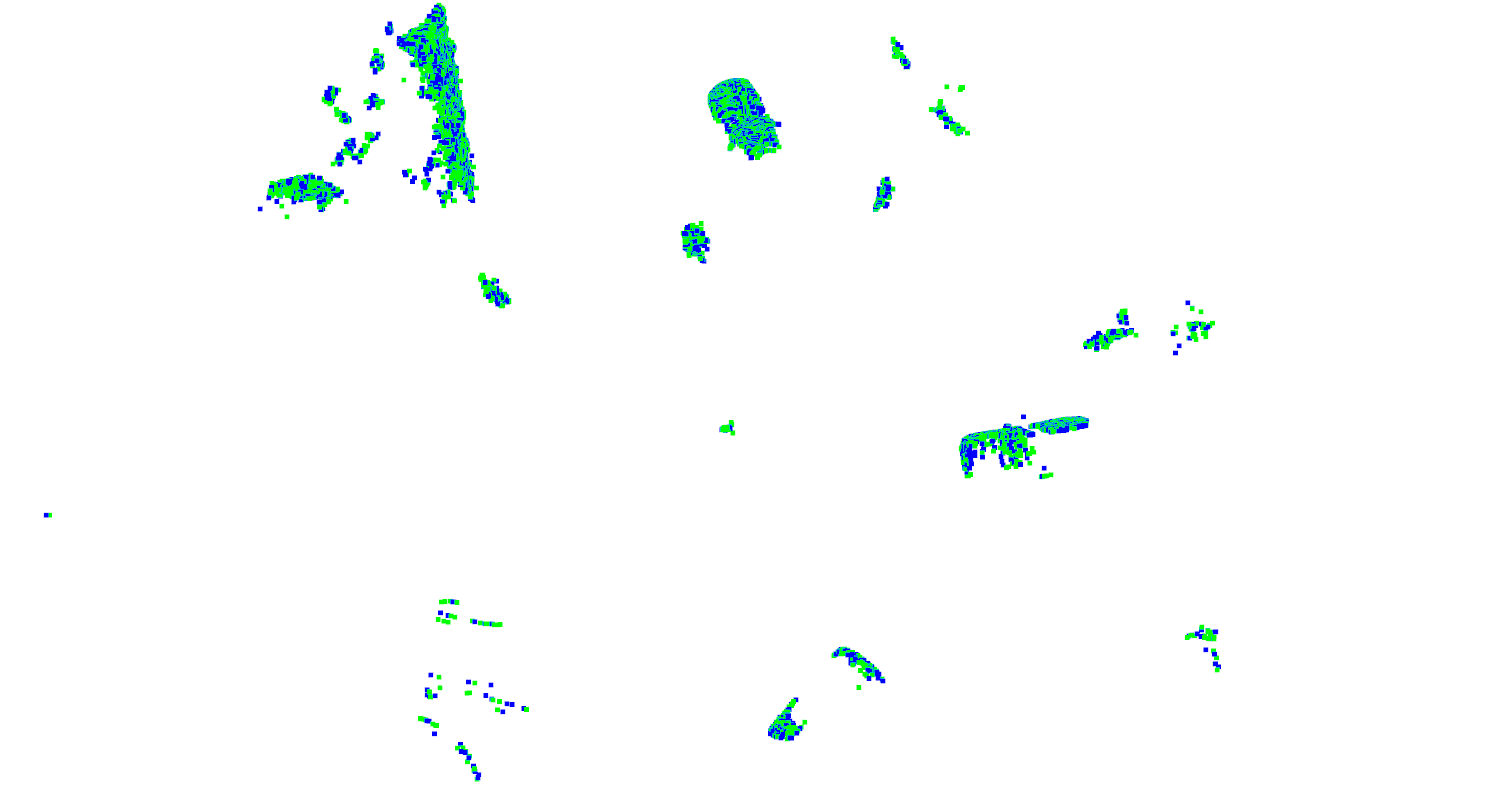}
    \end{subfigure}
    \begin{subfigure}{0.48\columnwidth}
        \centering
        \includegraphics[width=1\columnwidth, trim={0cm 0cm 0cm 1cm}, clip]{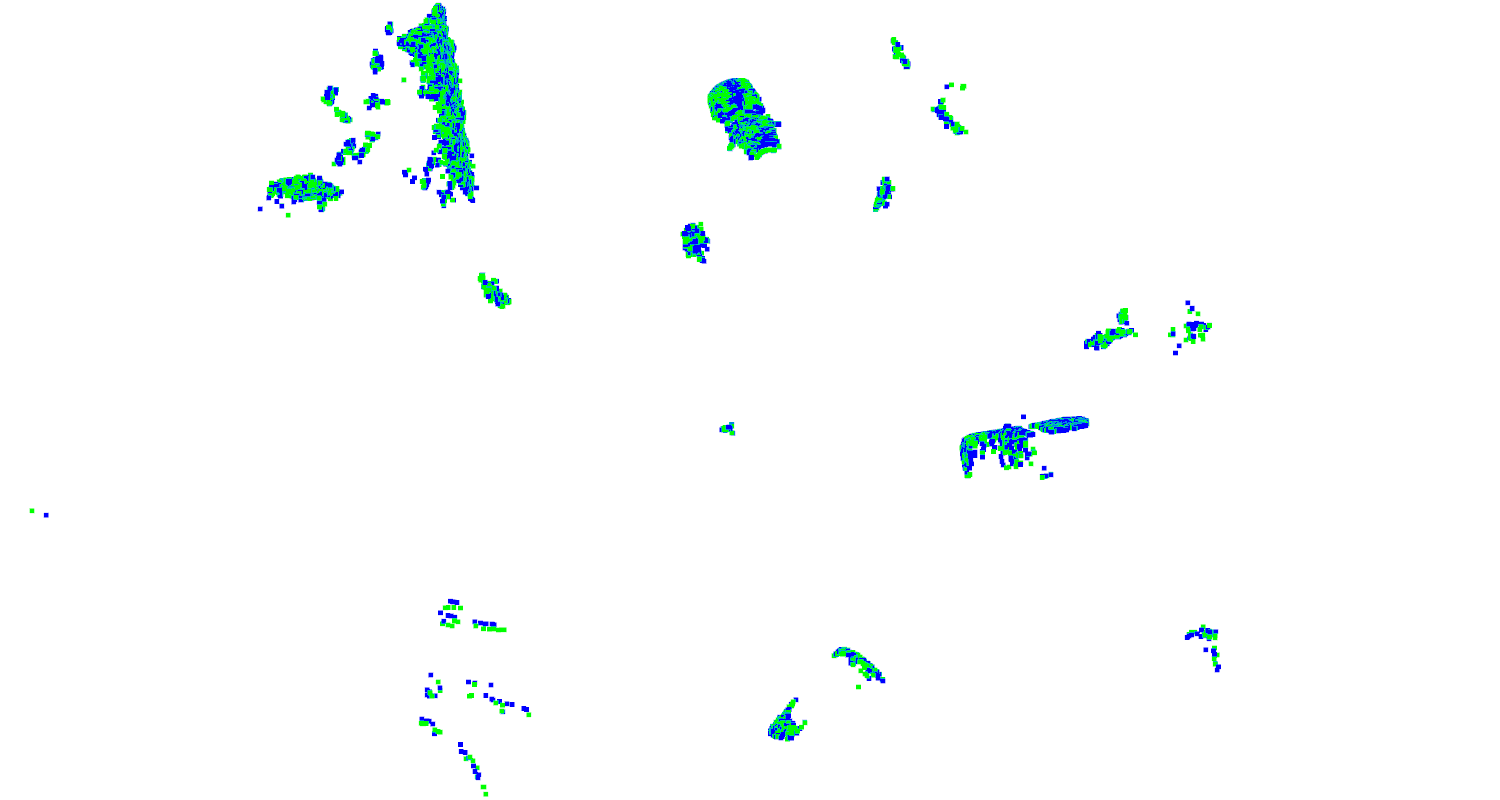}
    \end{subfigure}
    \begin{subfigure}{0.48\columnwidth}
        \centering
        \includegraphics[width=1\columnwidth, trim={0cm 0cm 0cm 1cm}, clip]{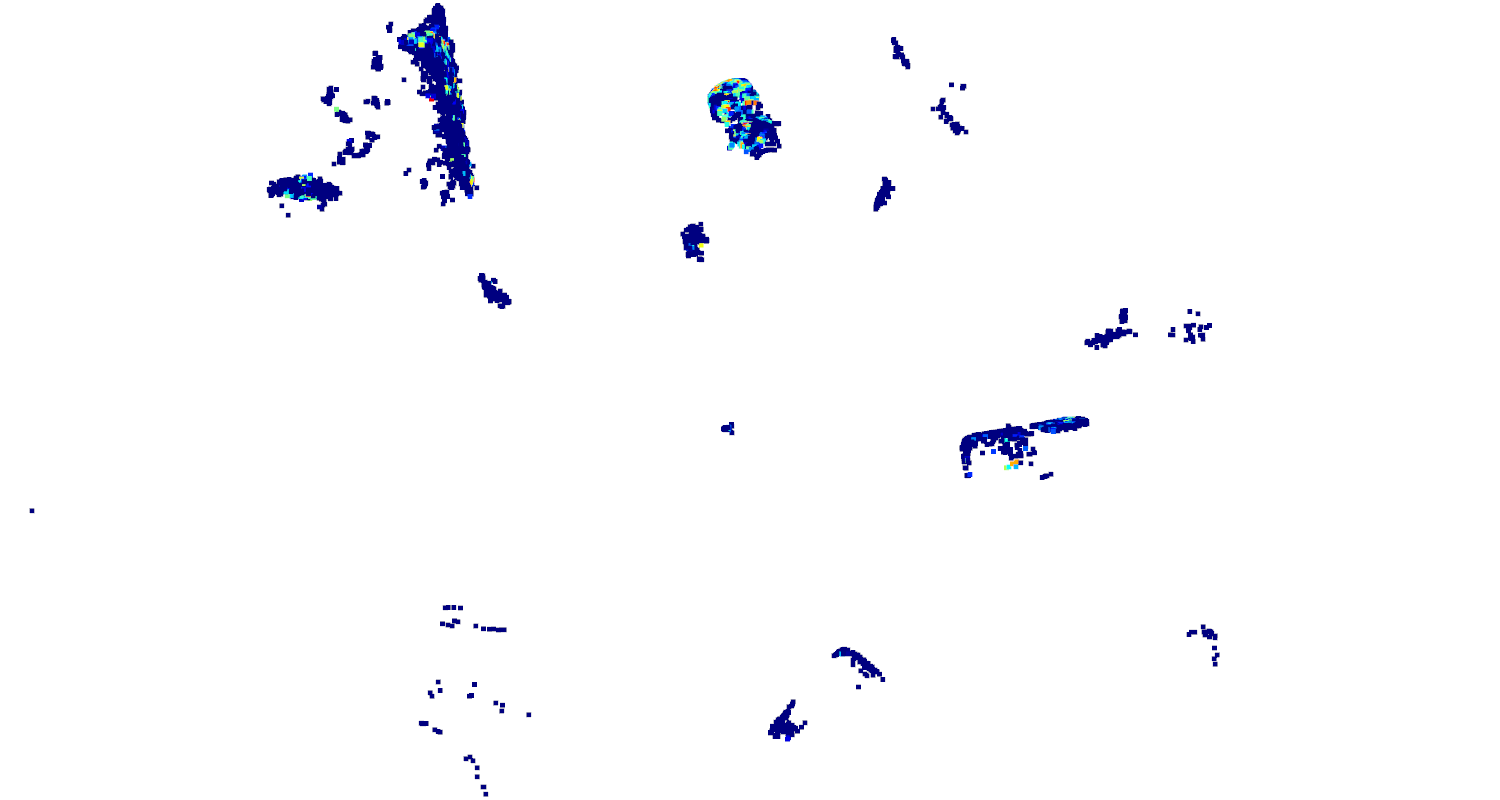}
    \end{subfigure}

    \rule{0.95\textwidth}{0.1pt}
    
    \vspace{2mm}
    
    \begin{subfigure}{0.48\columnwidth}
        \centering
        \includegraphics[width=1\columnwidth, trim={0cm 0cm 0cm 1cm}, clip]{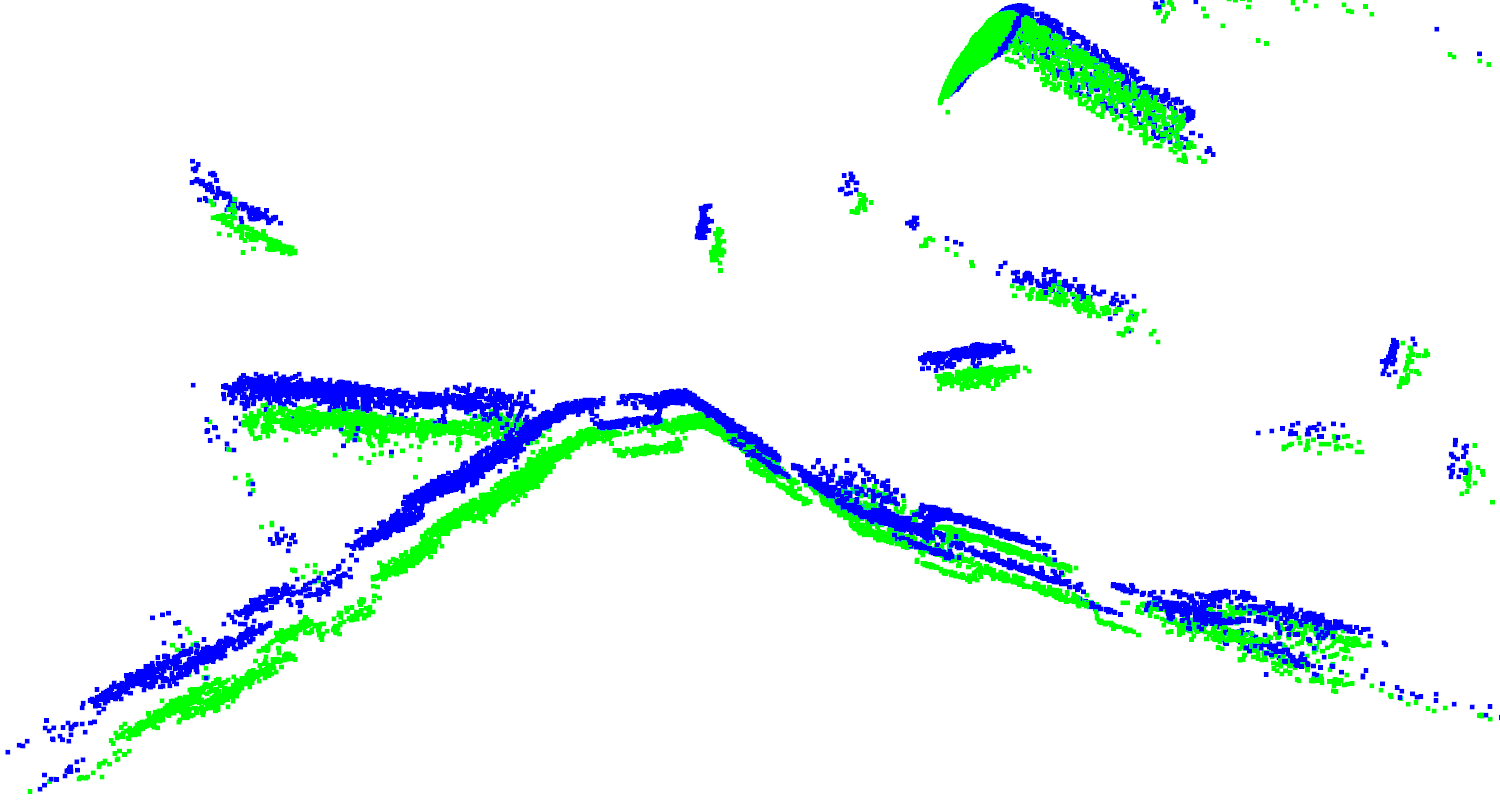}
        \caption*{PC1 and PC2}
    \end{subfigure}
    \begin{subfigure}{0.48\columnwidth}
        \centering
        \includegraphics[width=1\columnwidth, trim={0cm 0cm 0cm 1cm}, clip]{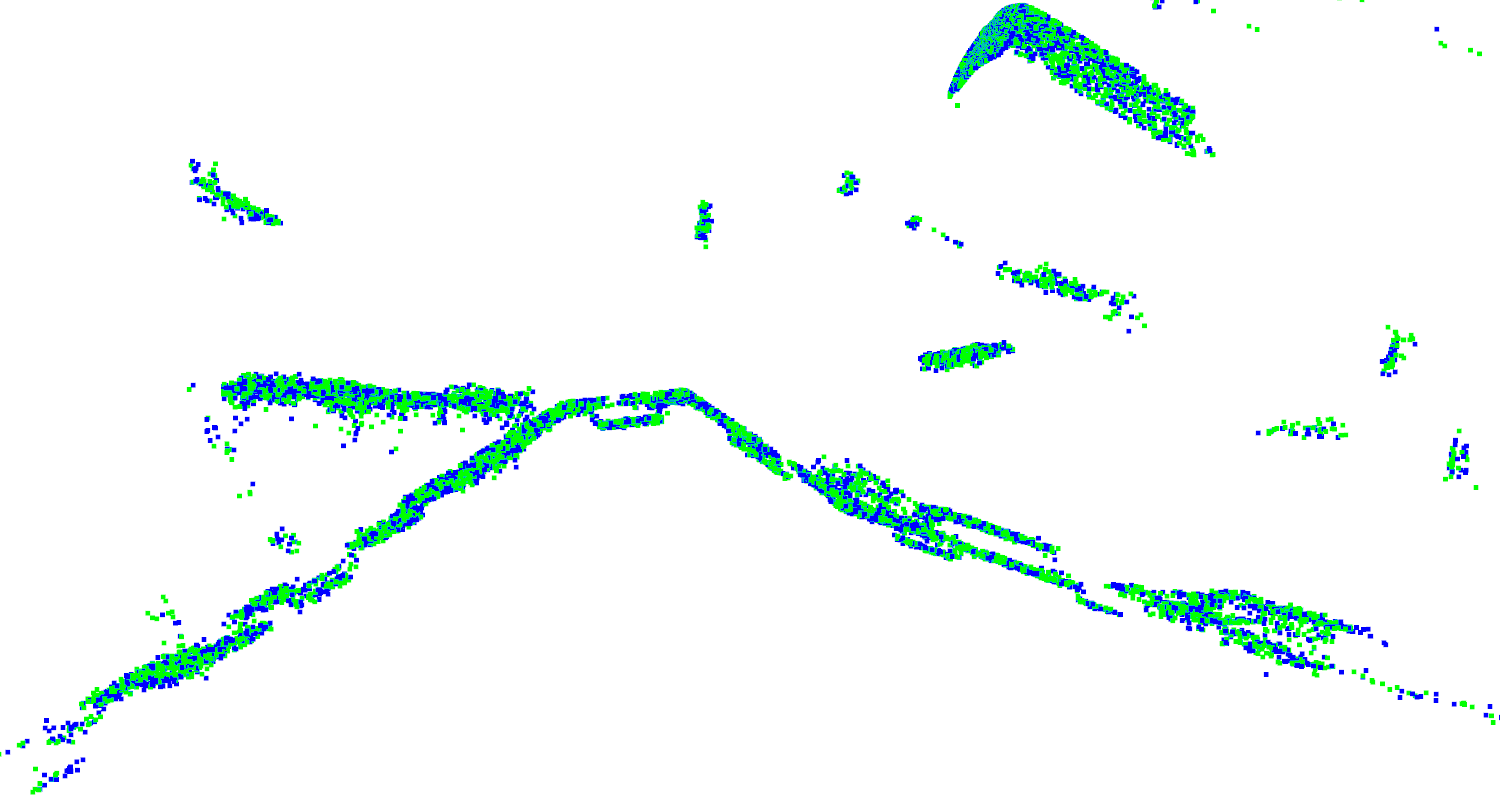}
        \caption*{PC1+GT and PC2}
    \end{subfigure}
    \begin{subfigure}{0.48\columnwidth}
        \centering
        \includegraphics[width=1\columnwidth, trim={0cm 0cm 0cm 1cm}, clip]{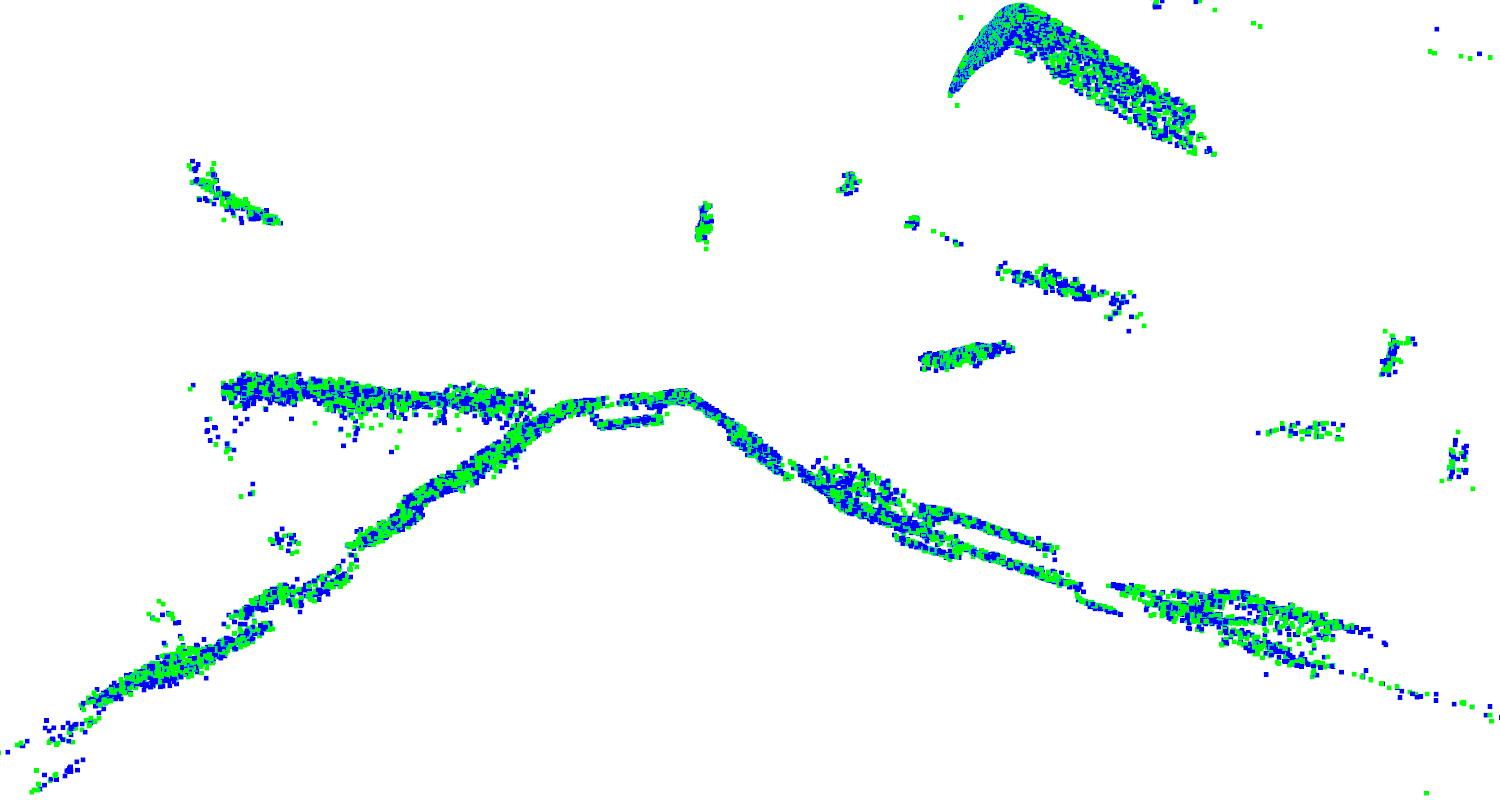}
        \caption*{PC1+Pred and PC2}
    \end{subfigure}
    \begin{subfigure}{0.48\columnwidth}
        \centering
        \includegraphics[width=1\columnwidth, trim={0cm 0cm 0cm 1cm}, clip]{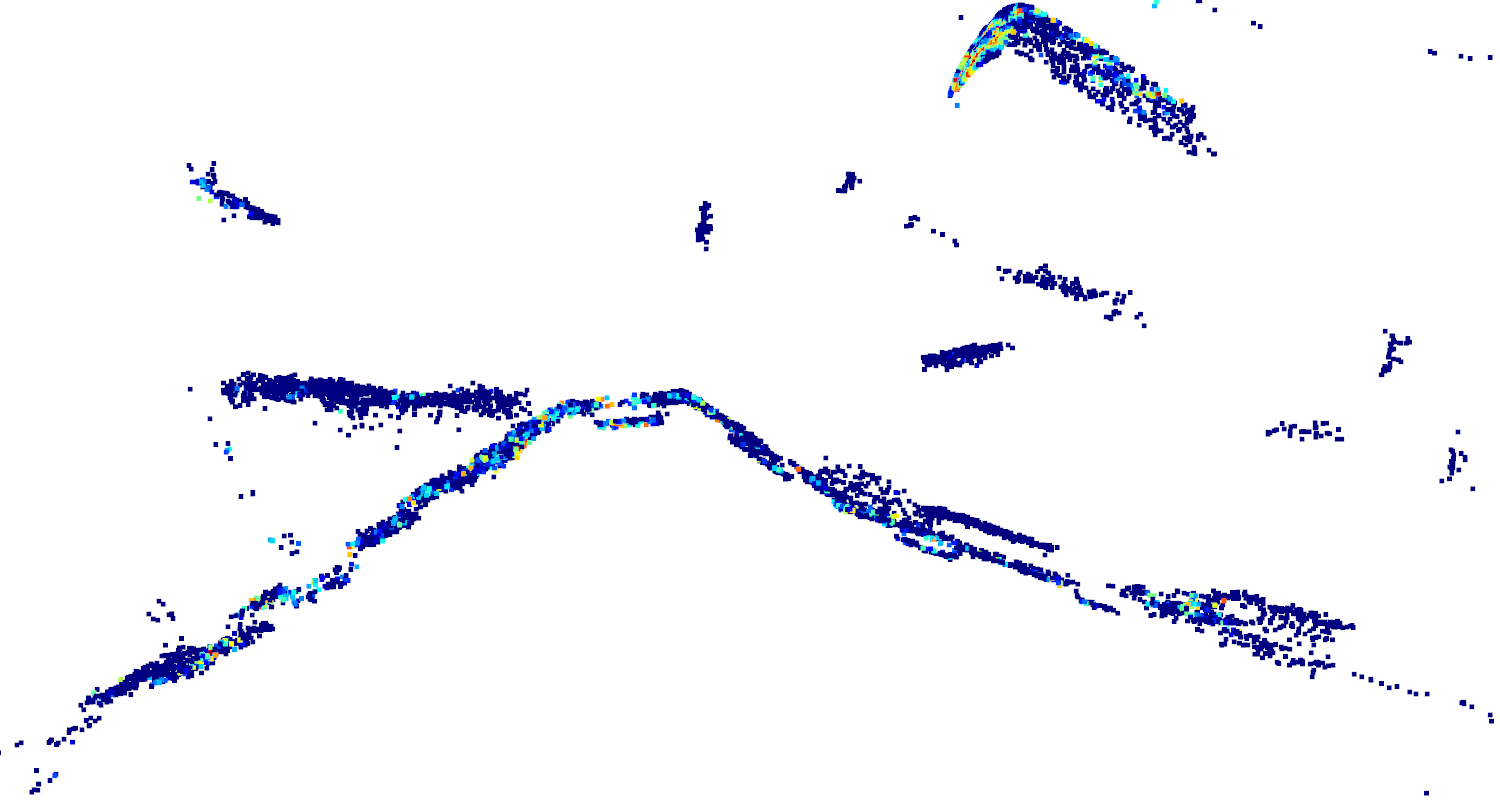}
        \caption*{Error}
    \end{subfigure}
    
    \caption{More qualitative results on the KITTI~\cite{kitti2015} scene flow dataset.}
    \label{fig:res-kitti-supp}
    
\end{figure*}

\begin{figure*}[tp]
    \centering
    
    \rule{0.95\textwidth}{0.1pt}
 
    \begin{subfigure}{0.48\columnwidth}
        \centering
        \includegraphics[width=1\columnwidth, trim={0cm 0cm 0cm 1cm}, clip]{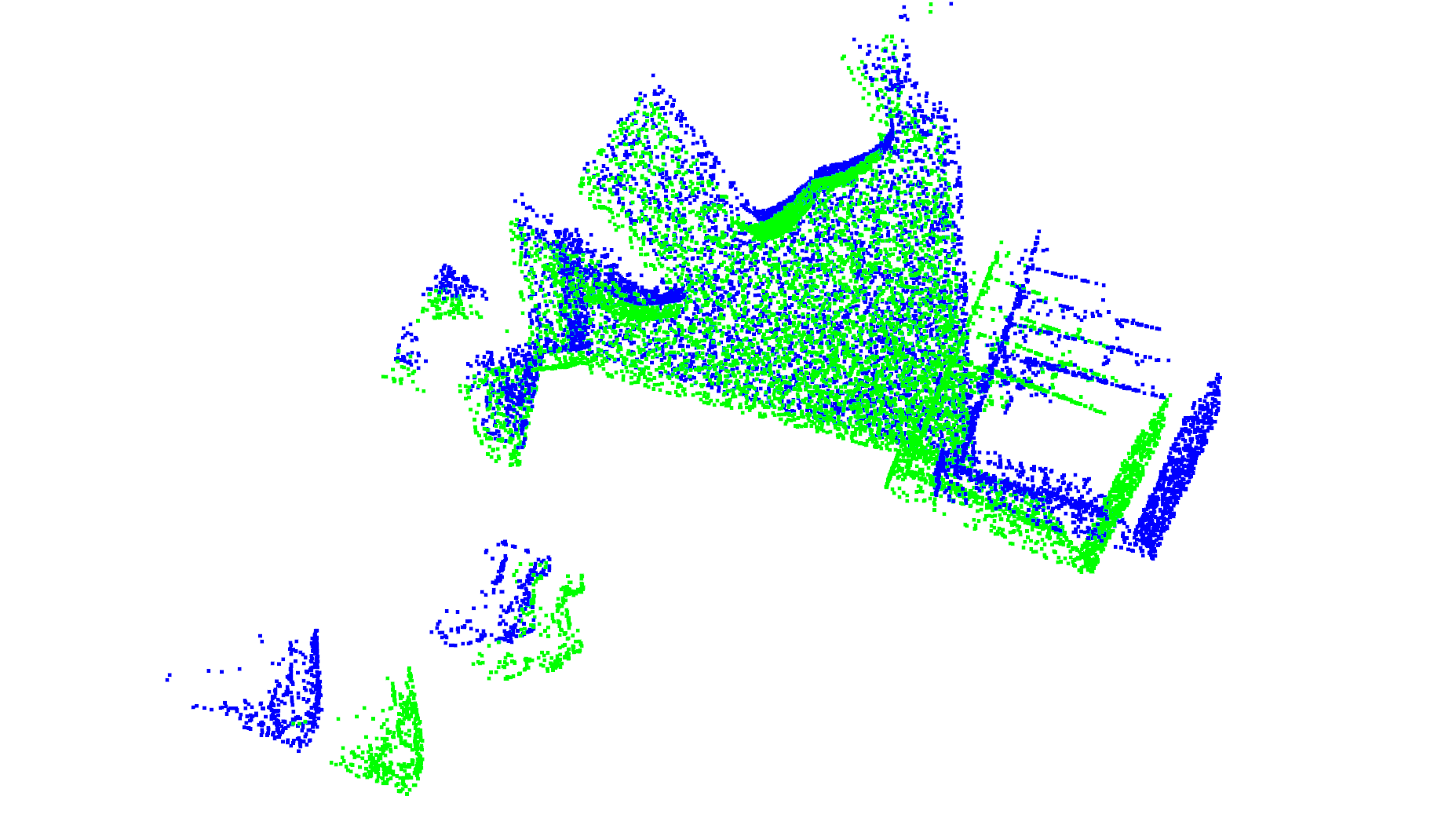}
    \end{subfigure}
    \begin{subfigure}{0.48\columnwidth}
        \centering
        \includegraphics[width=1\columnwidth, trim={0cm 0cm 0cm 1cm}, clip]{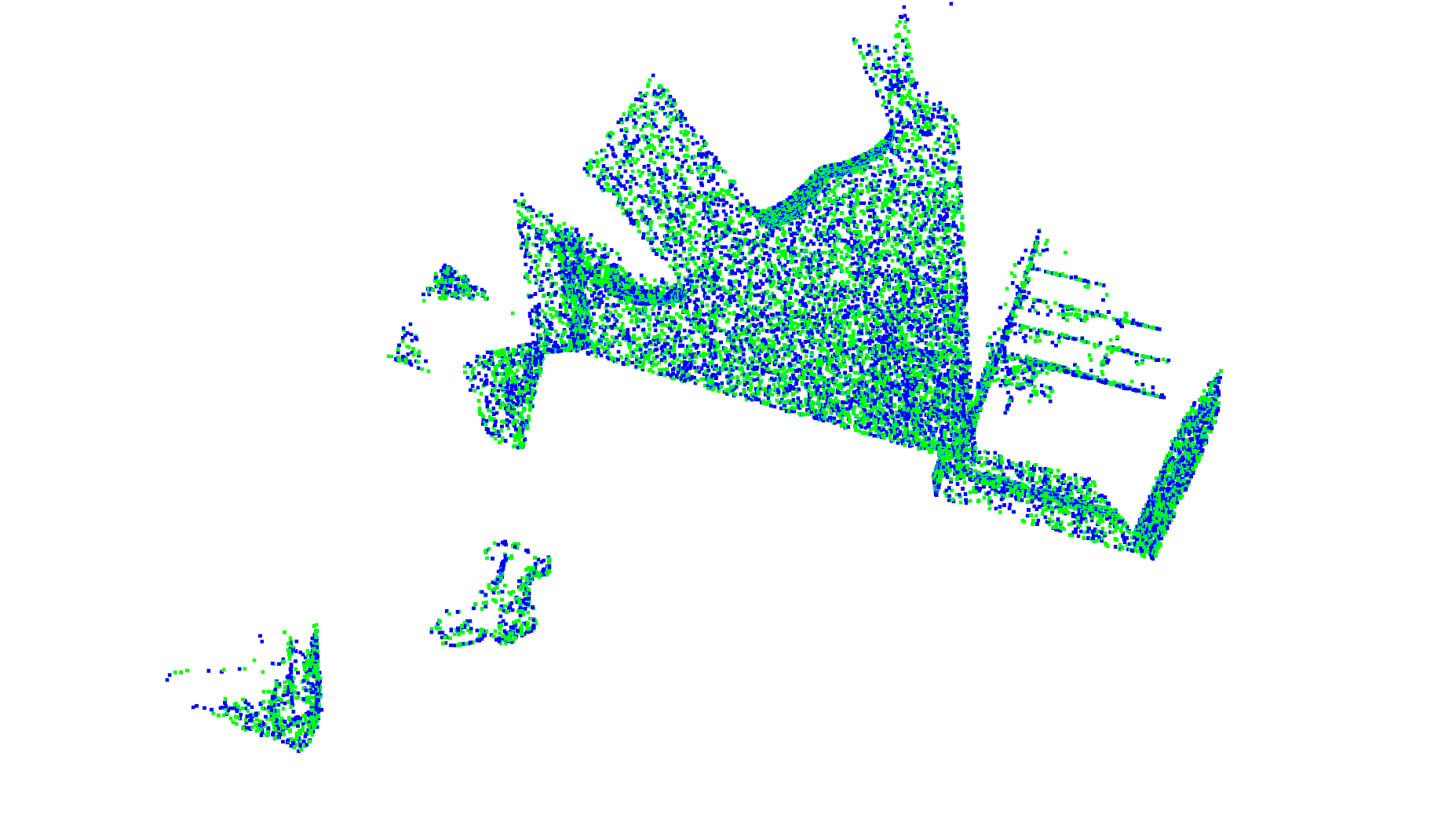}
    \end{subfigure}
    \begin{subfigure}{0.48\columnwidth}
        \centering
        \includegraphics[width=1\columnwidth, trim={0cm 0cm 0cm 1cm}, clip]{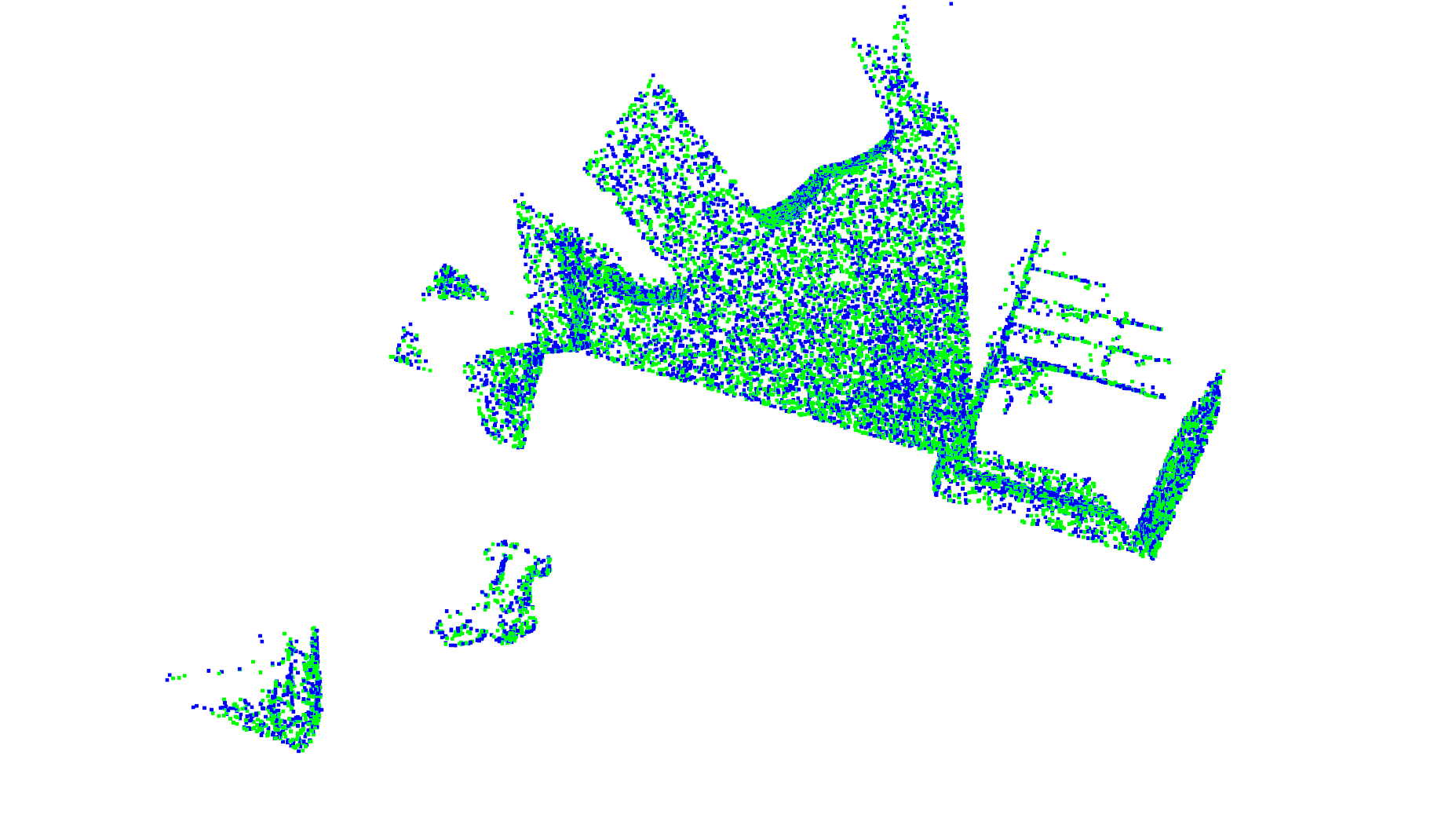}
    \end{subfigure}
    \begin{subfigure}{0.48\columnwidth}
        \centering
        \includegraphics[width=1\columnwidth, trim={0cm 0cm 0cm 1cm}, clip]{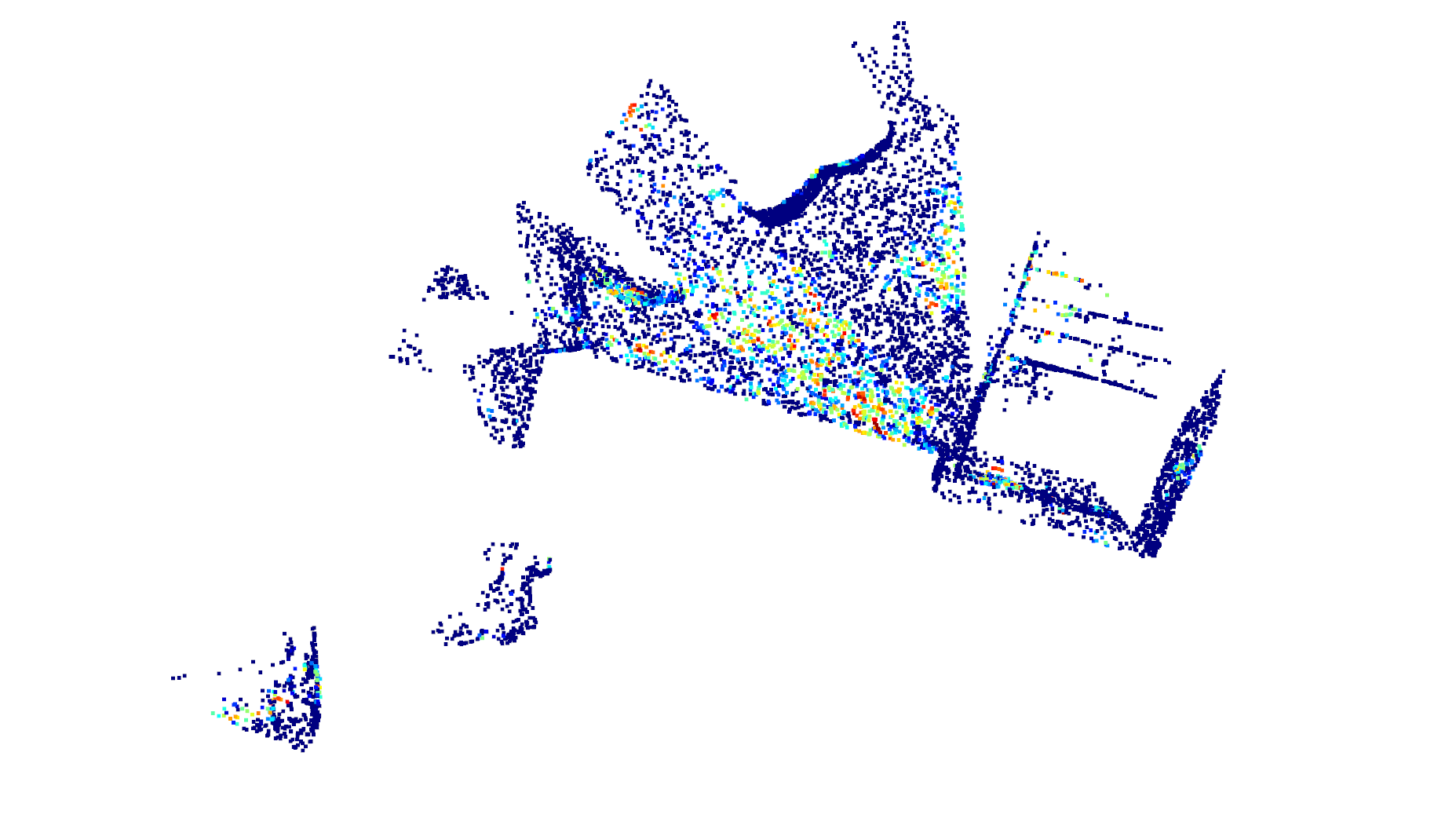}
    \end{subfigure}
    
    \rule{0.95\textwidth}{0.1pt}

    \begin{subfigure}{0.48\columnwidth}
        \centering
        \includegraphics[width=1\columnwidth, trim={0cm 0cm 0cm 0cm}, clip]{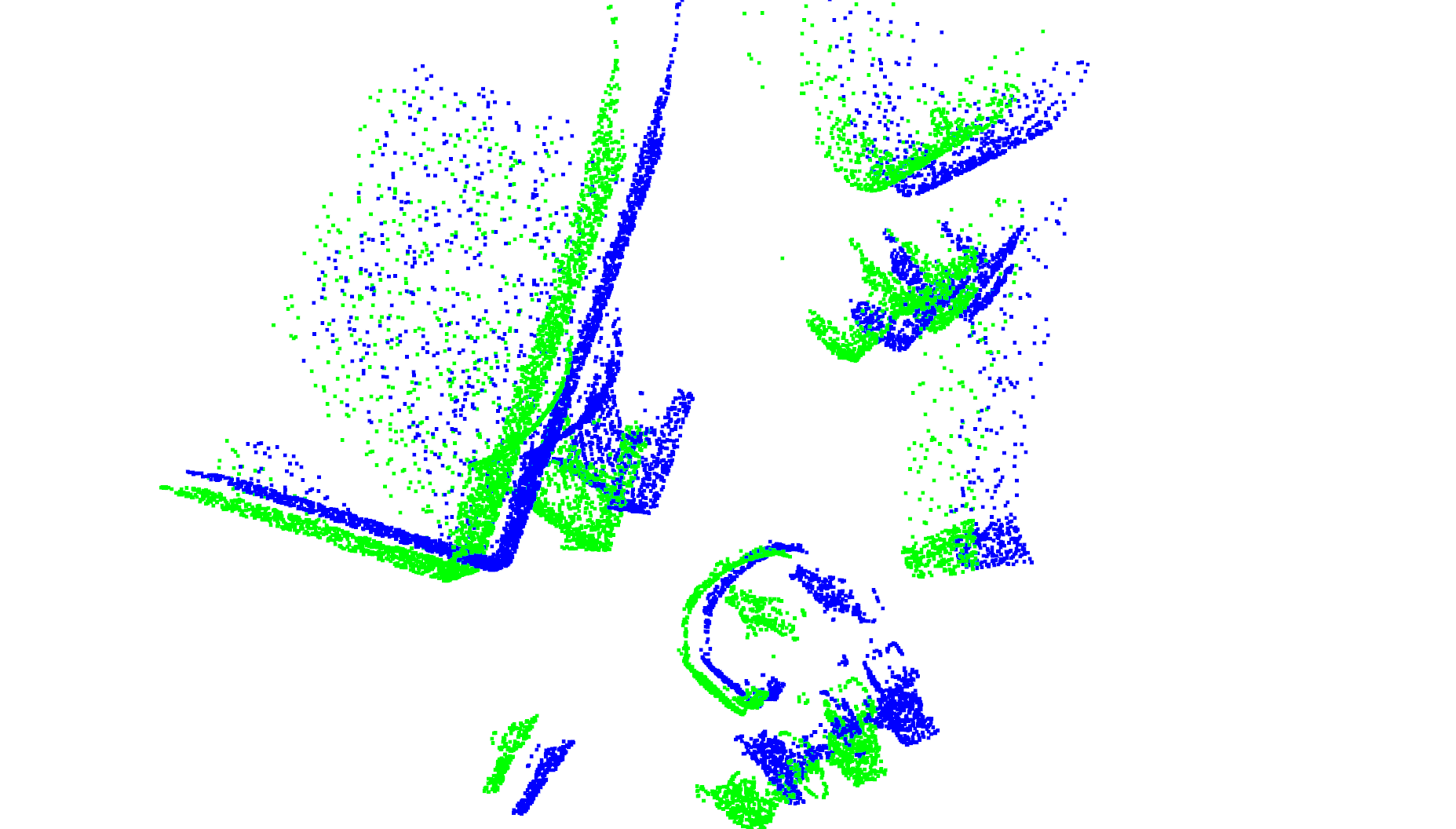}
    \end{subfigure}
    \begin{subfigure}{0.48\columnwidth}
        \centering
        \includegraphics[width=1\columnwidth, trim={0cm 0cm 0cm 0cm}, clip]{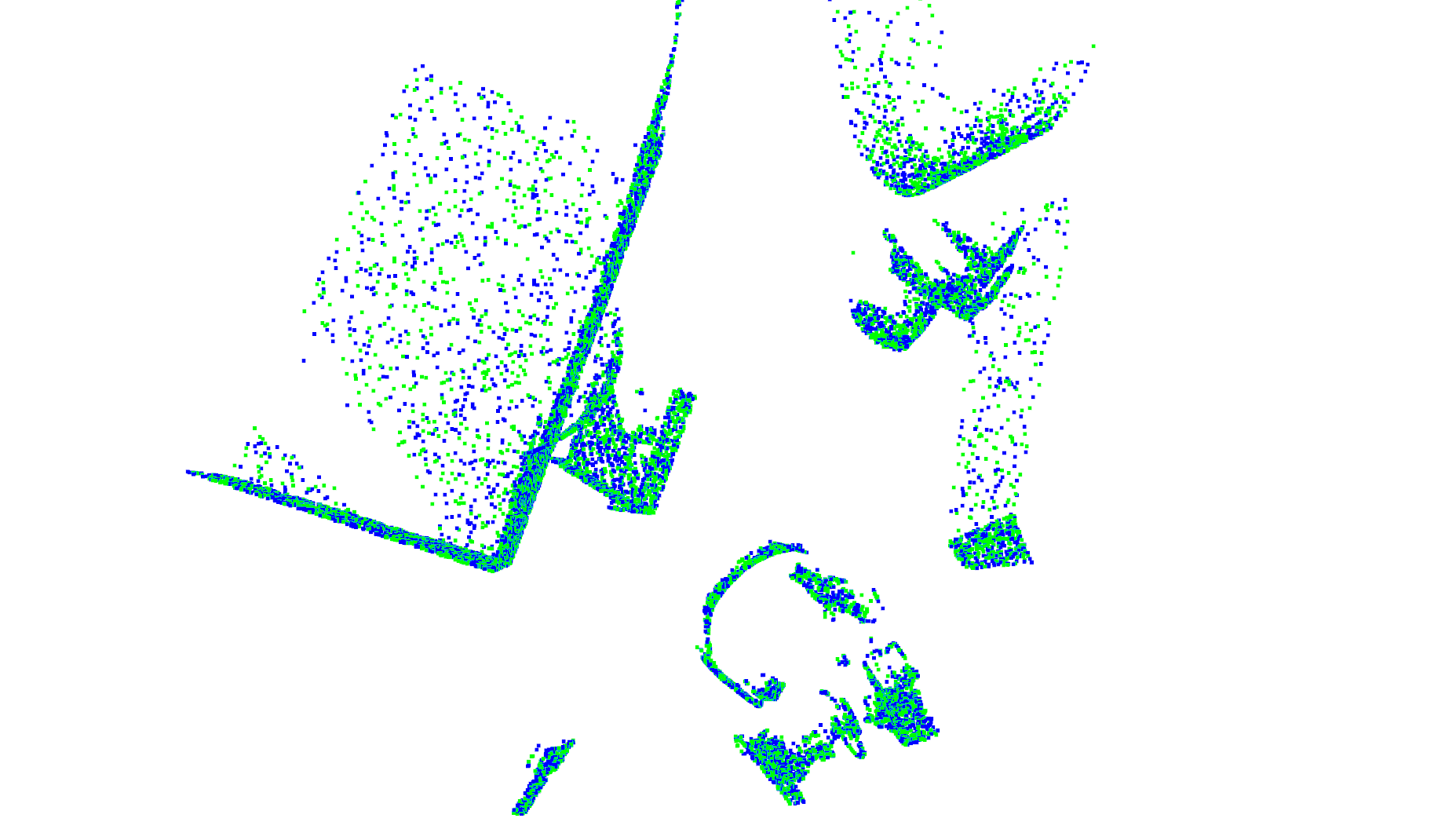}
    \end{subfigure}
    \begin{subfigure}{0.48\columnwidth}
        \centering
        \includegraphics[width=1\columnwidth, trim={0cm 0cm 0cm 0cm}, clip]{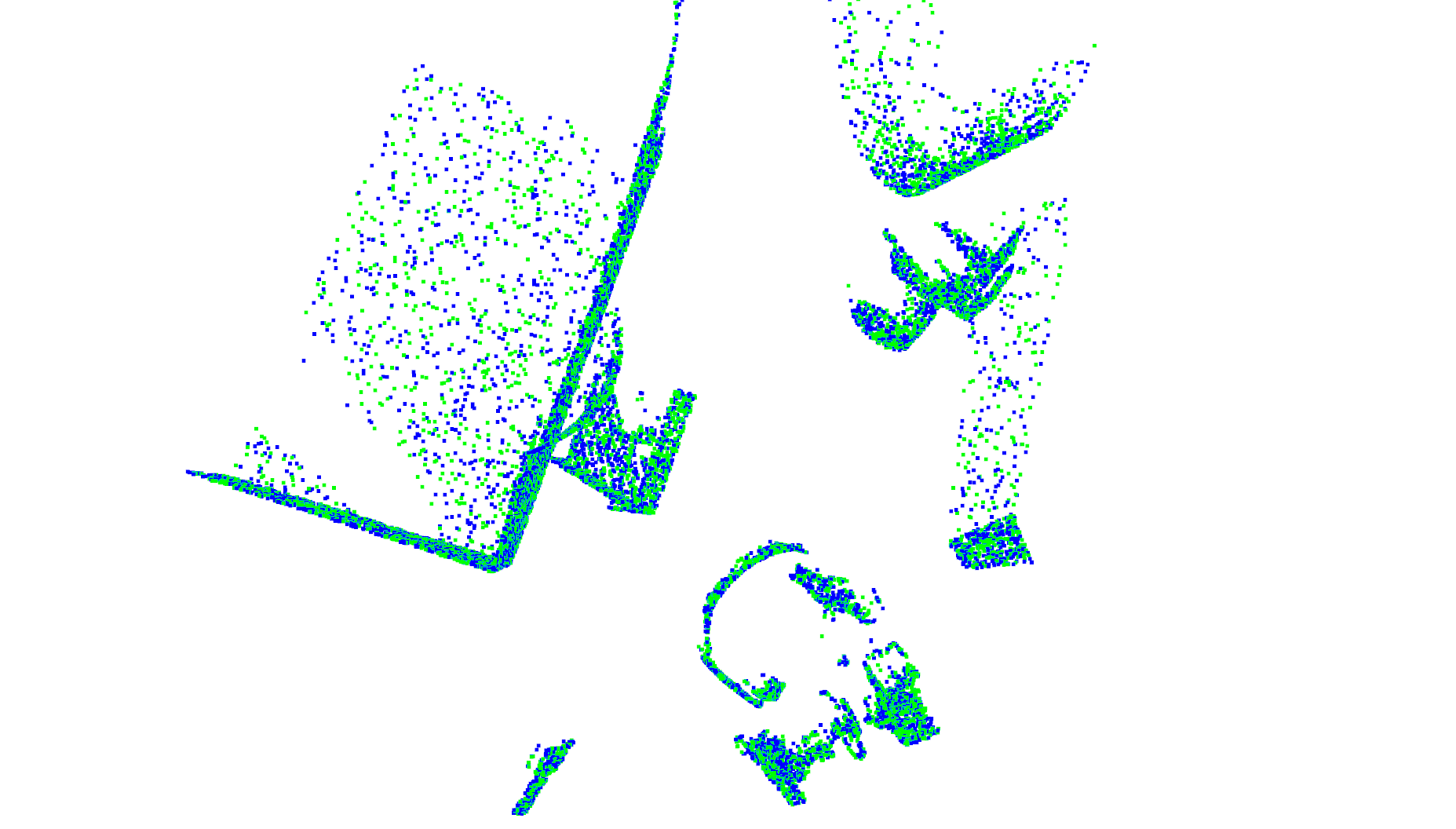}
    \end{subfigure}
    \begin{subfigure}{0.48\columnwidth}
        \centering
        \includegraphics[width=1\columnwidth, trim={0cm 0cm 0cm 0cm}, clip]{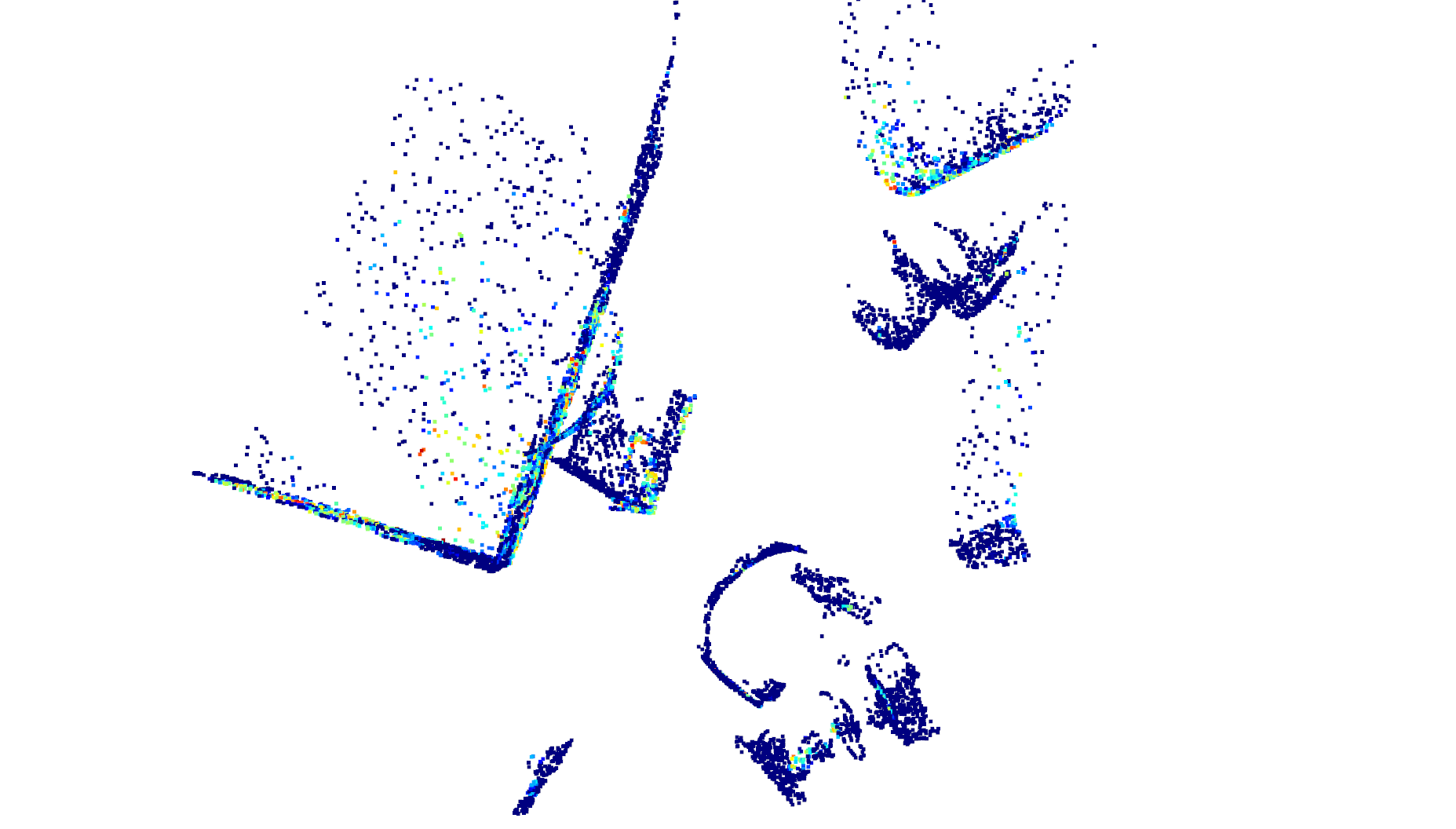}
    \end{subfigure}

    \rule{0.95\textwidth}{0.1pt}

    \begin{subfigure}{0.48\columnwidth}
        \centering
        \includegraphics[width=1\columnwidth, trim={0cm 0cm 0cm 0cm}, clip]{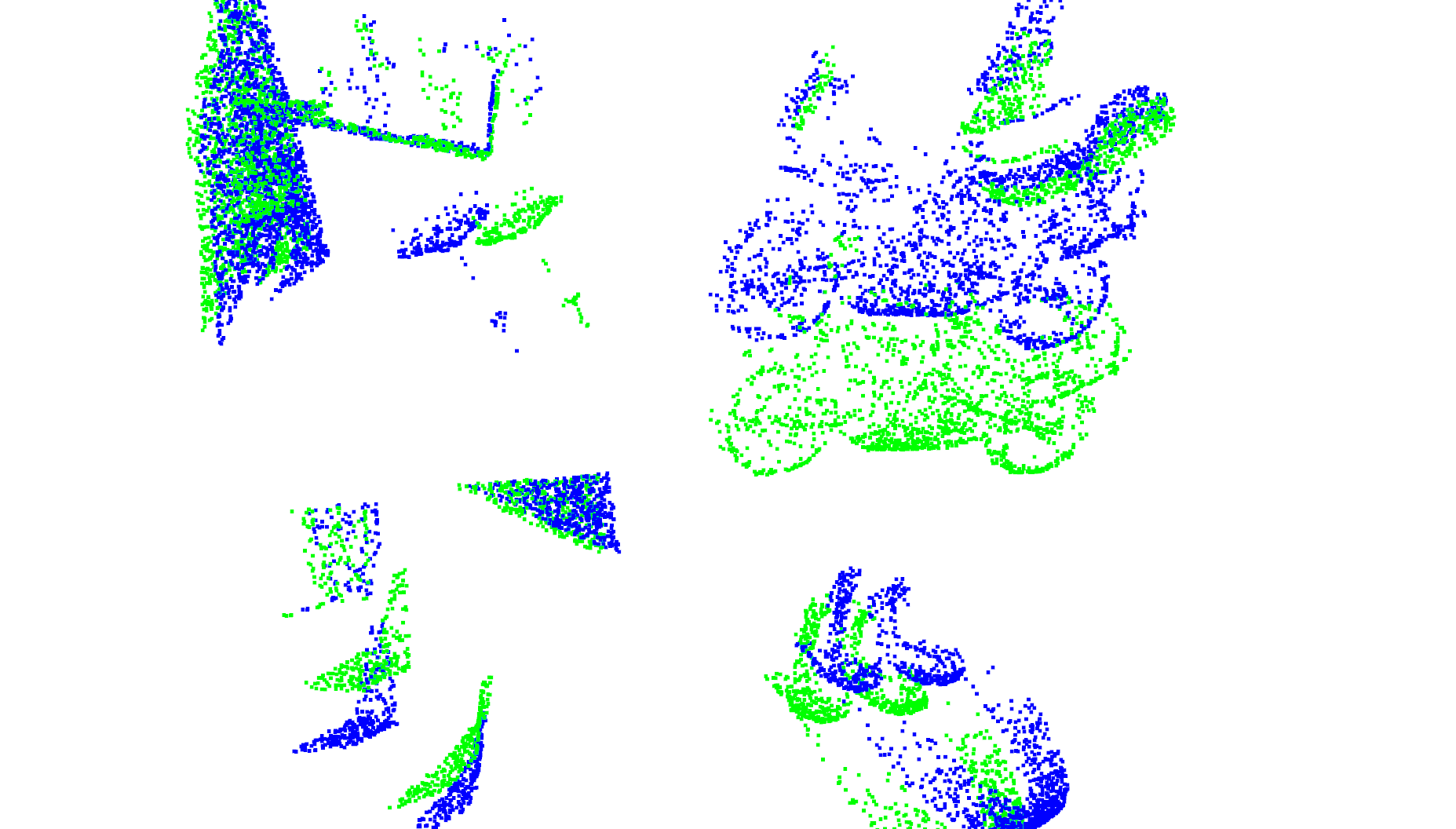}
    \end{subfigure}
    \begin{subfigure}{0.48\columnwidth}
        \centering
        \includegraphics[width=1\columnwidth, trim={0cm 0cm 0cm 0cm}, clip]{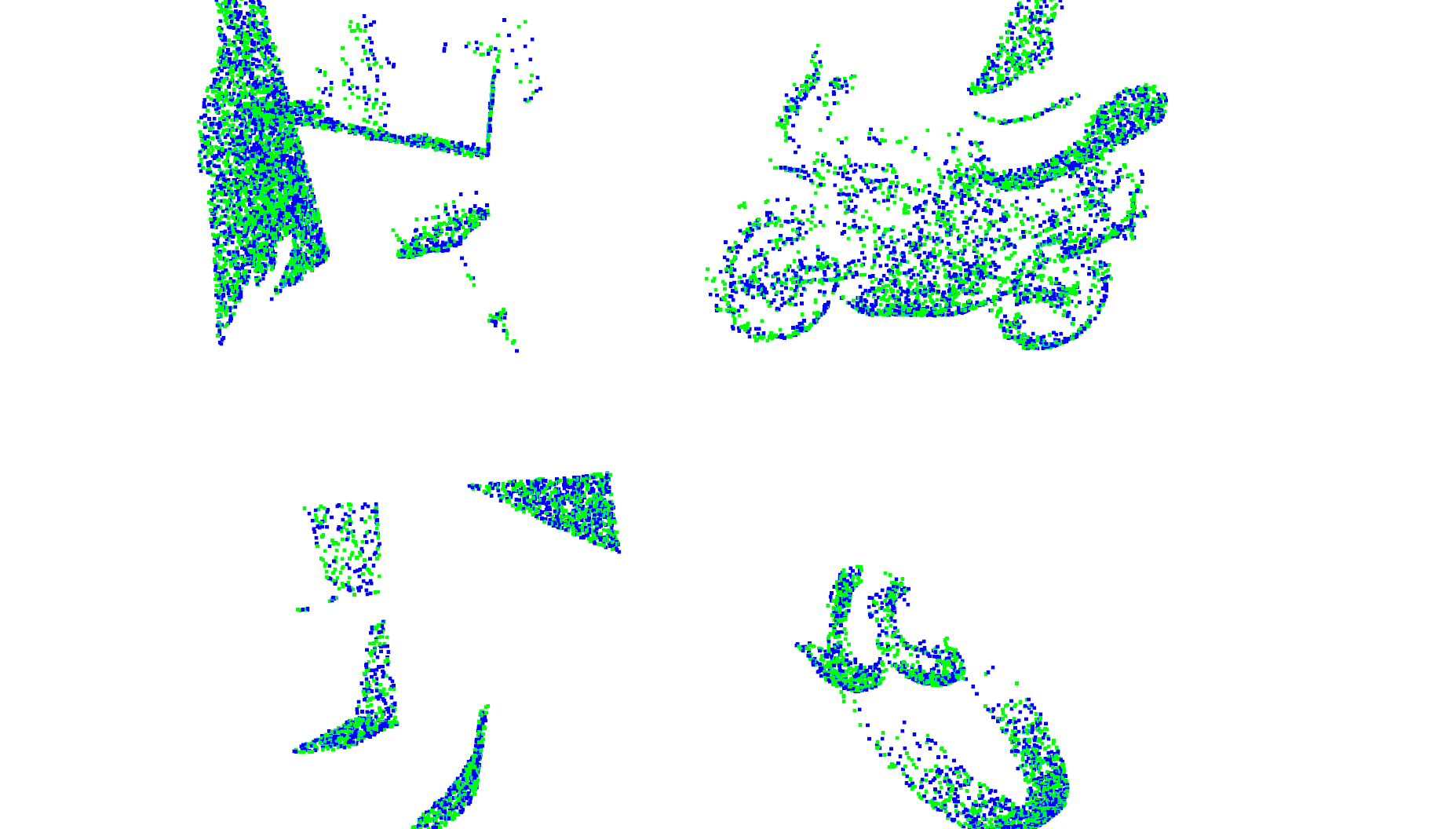}
    \end{subfigure}
    \begin{subfigure}{0.48\columnwidth}
        \centering
        \includegraphics[width=1\columnwidth, trim={0cm 0cm 0cm 0cm}, clip]{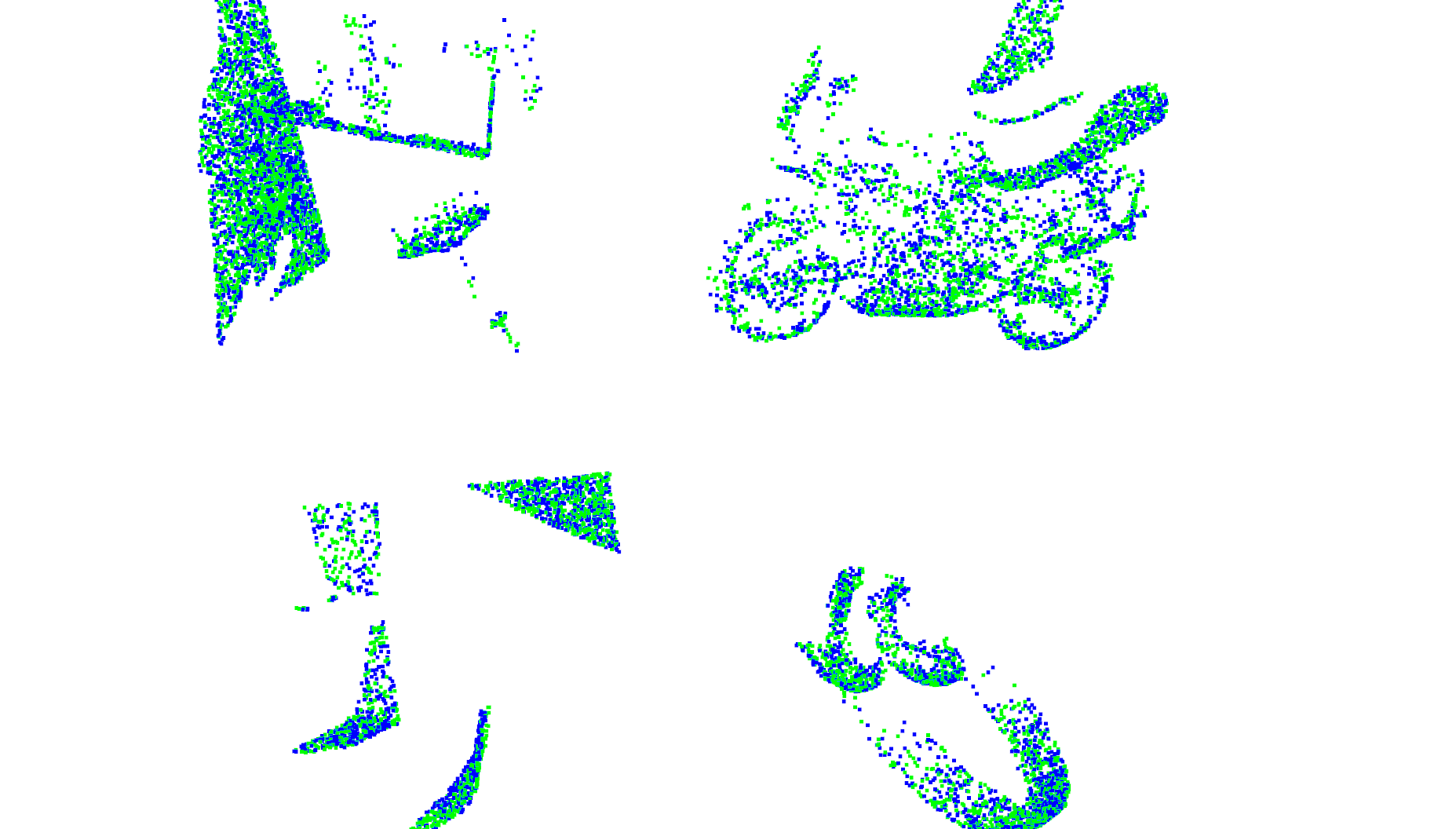}
    \end{subfigure}
    \begin{subfigure}{0.48\columnwidth}
        \centering
        \includegraphics[width=1\columnwidth, trim={0cm 0cm 0cm 0cm}, clip]{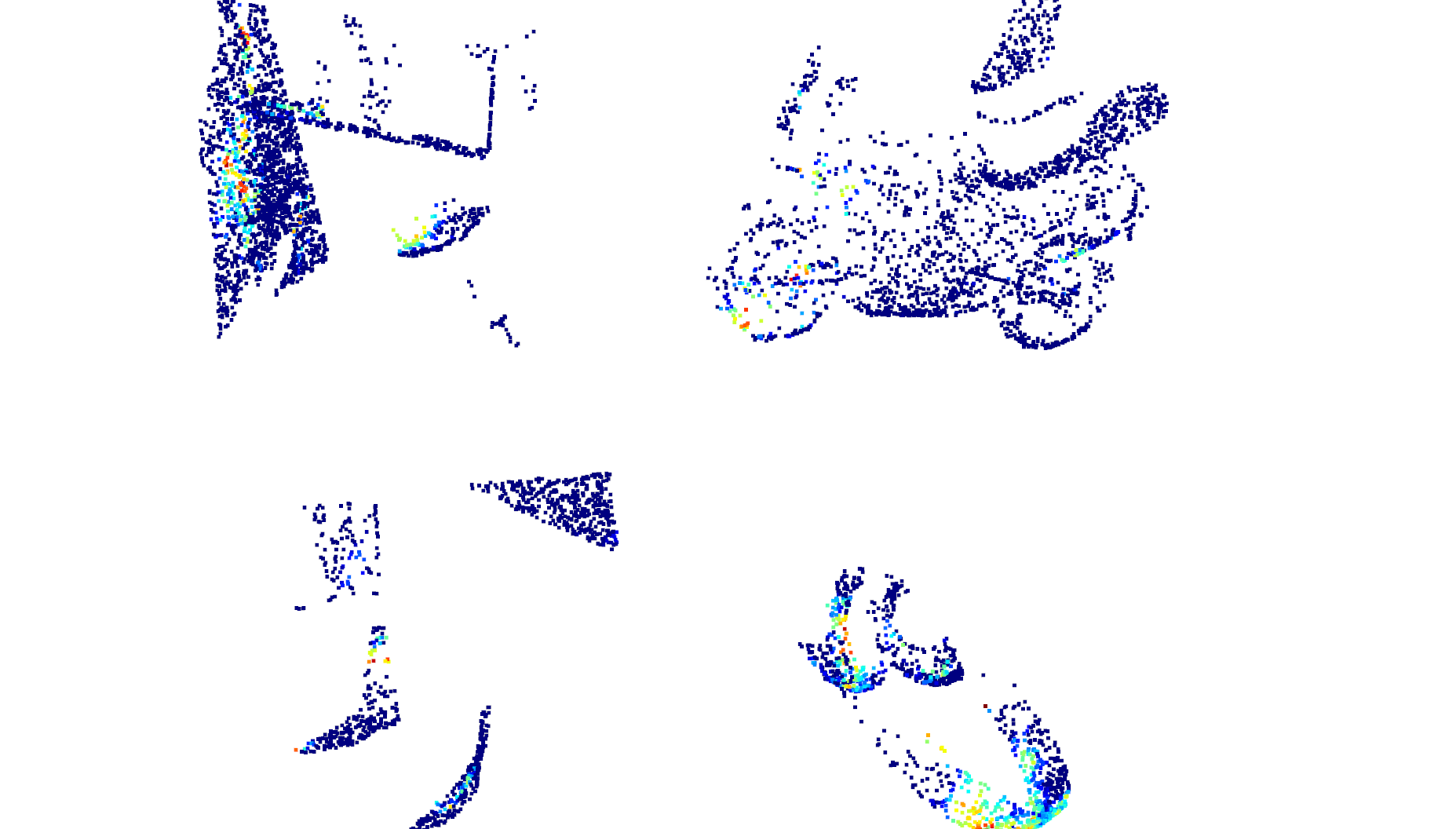}
    \end{subfigure}

    \rule{0.95\textwidth}{0.1pt}

    \begin{subfigure}{0.48\columnwidth}
        \centering
        \includegraphics[width=1\columnwidth, trim={0cm 0cm 0cm 0cm}, clip]{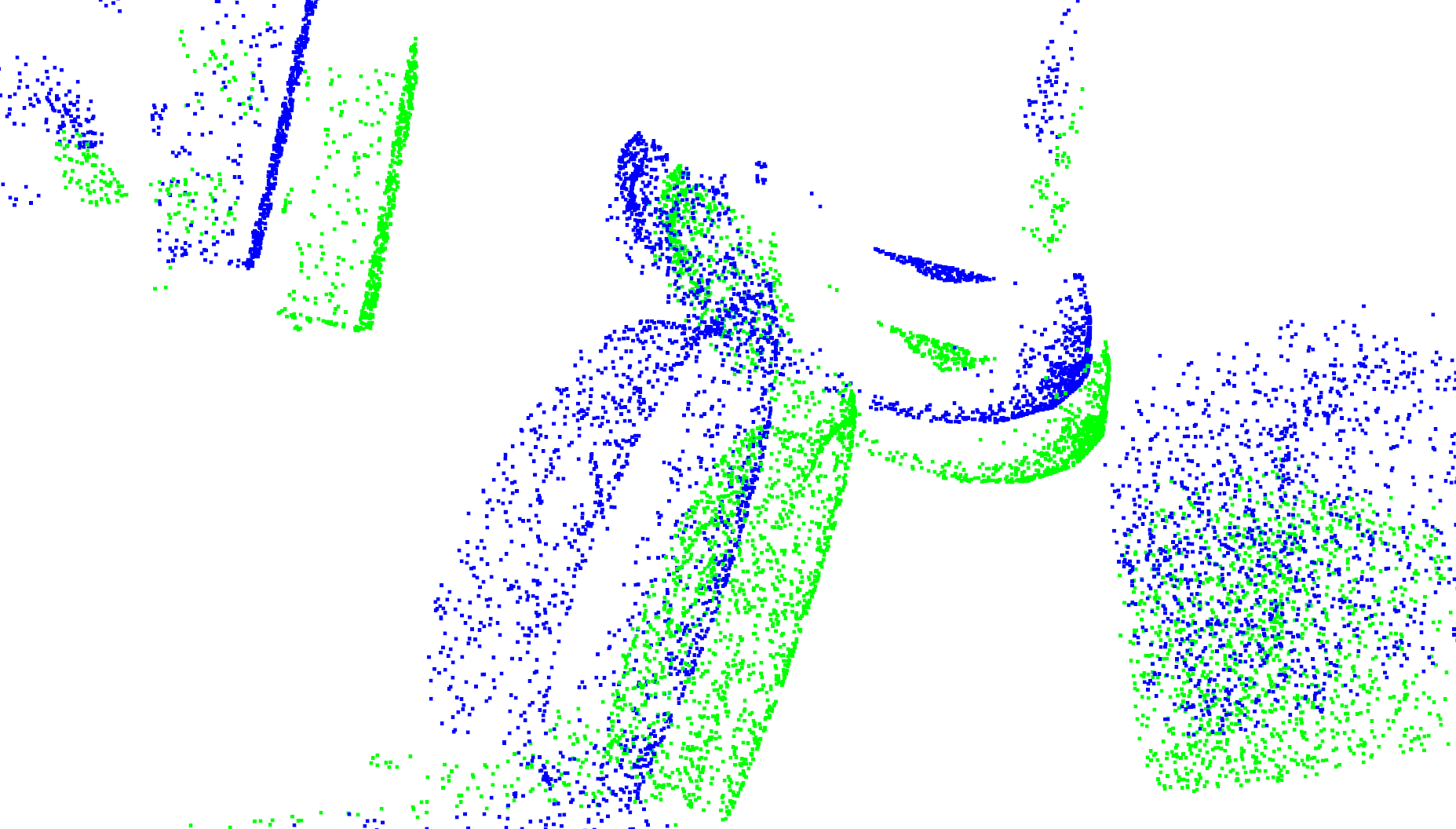}
    \end{subfigure}
    \begin{subfigure}{0.48\columnwidth}
        \centering
        \includegraphics[width=1\columnwidth, trim={0cm 0cm 0cm 0cm}, clip]{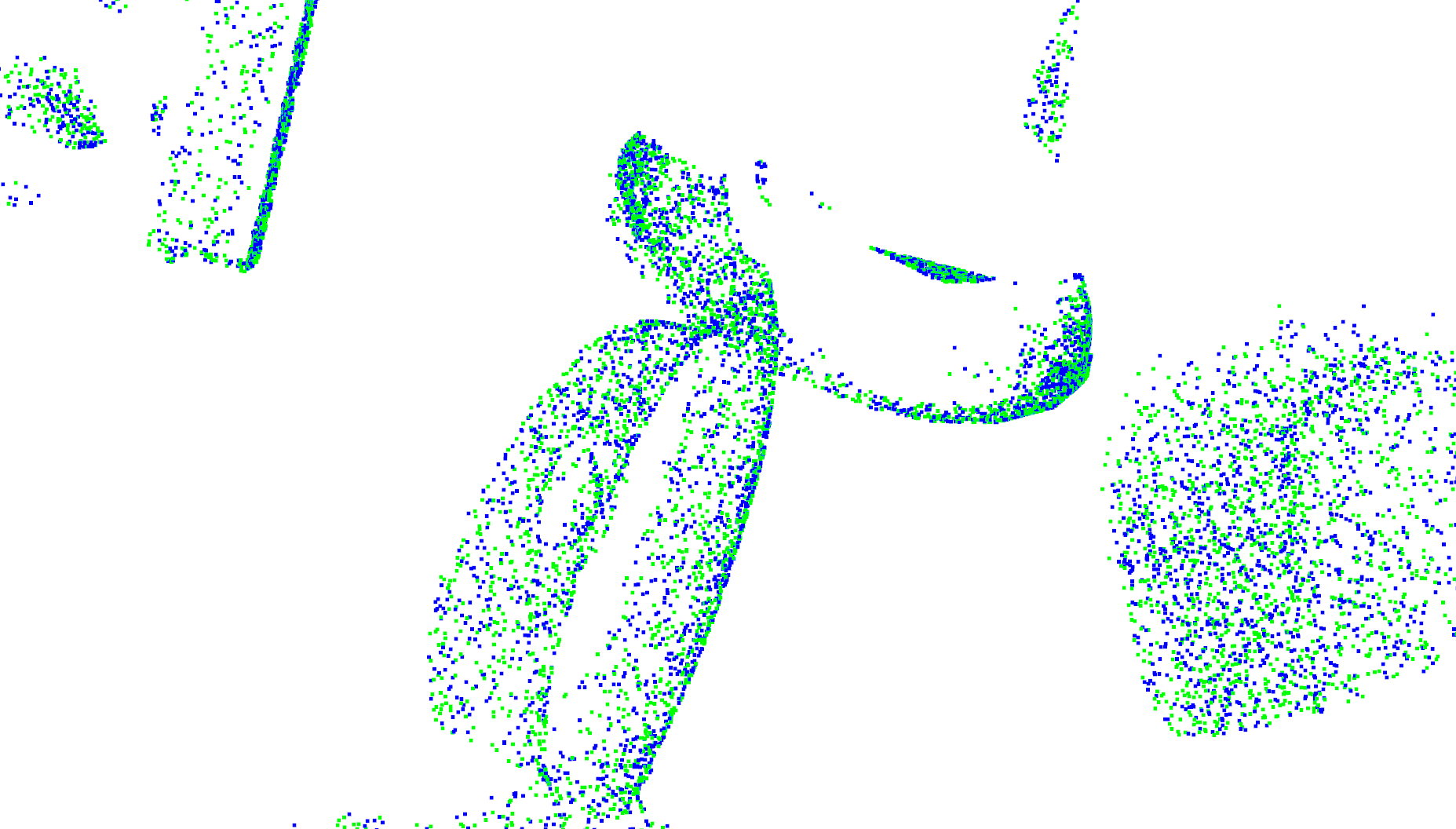}
    \end{subfigure}
    \begin{subfigure}{0.48\columnwidth}
        \centering
        \includegraphics[width=1\columnwidth, trim={0cm 0cm 0cm 0cm}, clip]{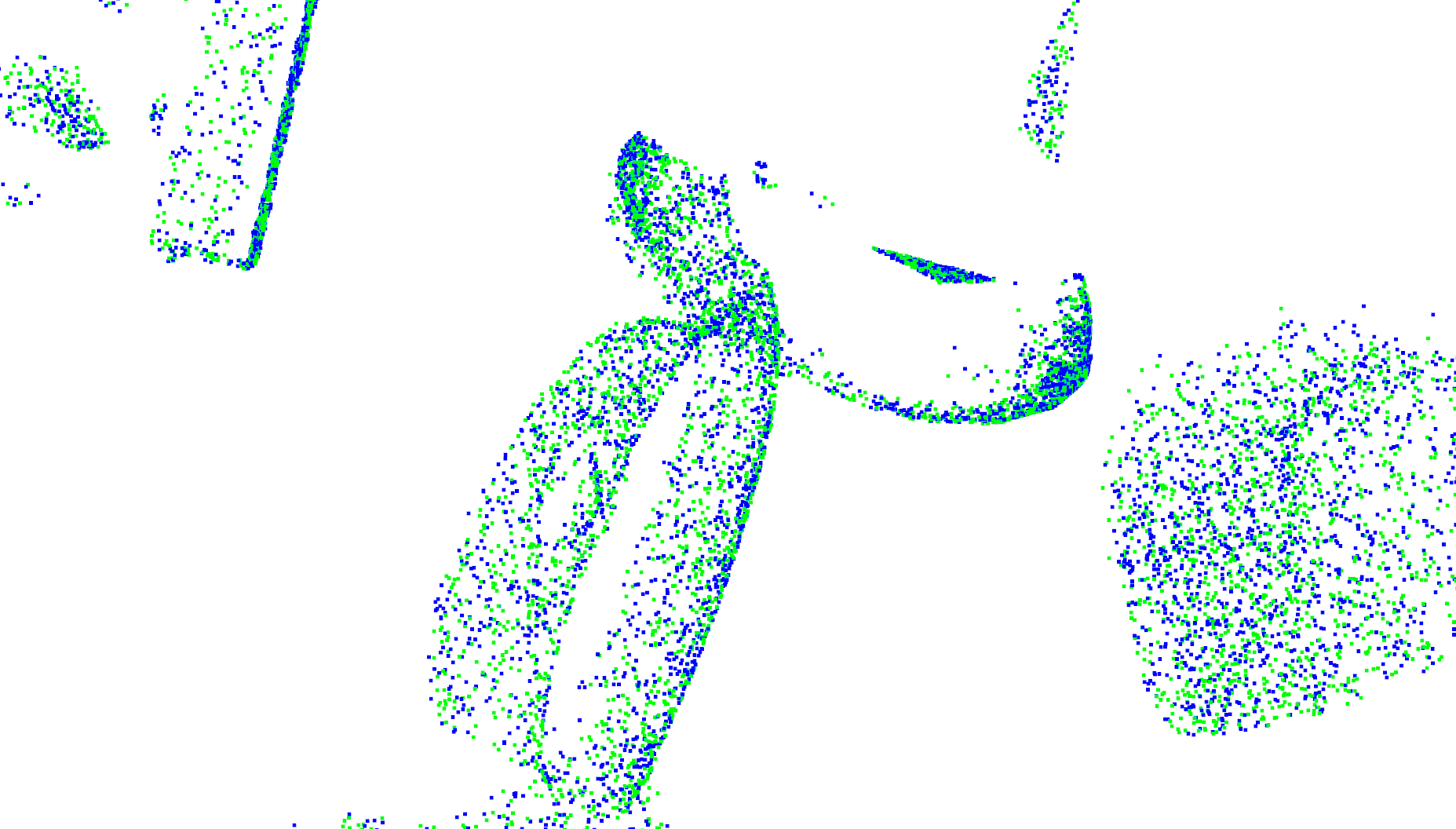}
    \end{subfigure}
    \begin{subfigure}{0.48\columnwidth}
        \centering
        \includegraphics[width=1\columnwidth, trim={0cm 0cm 0cm 0cm}, clip]{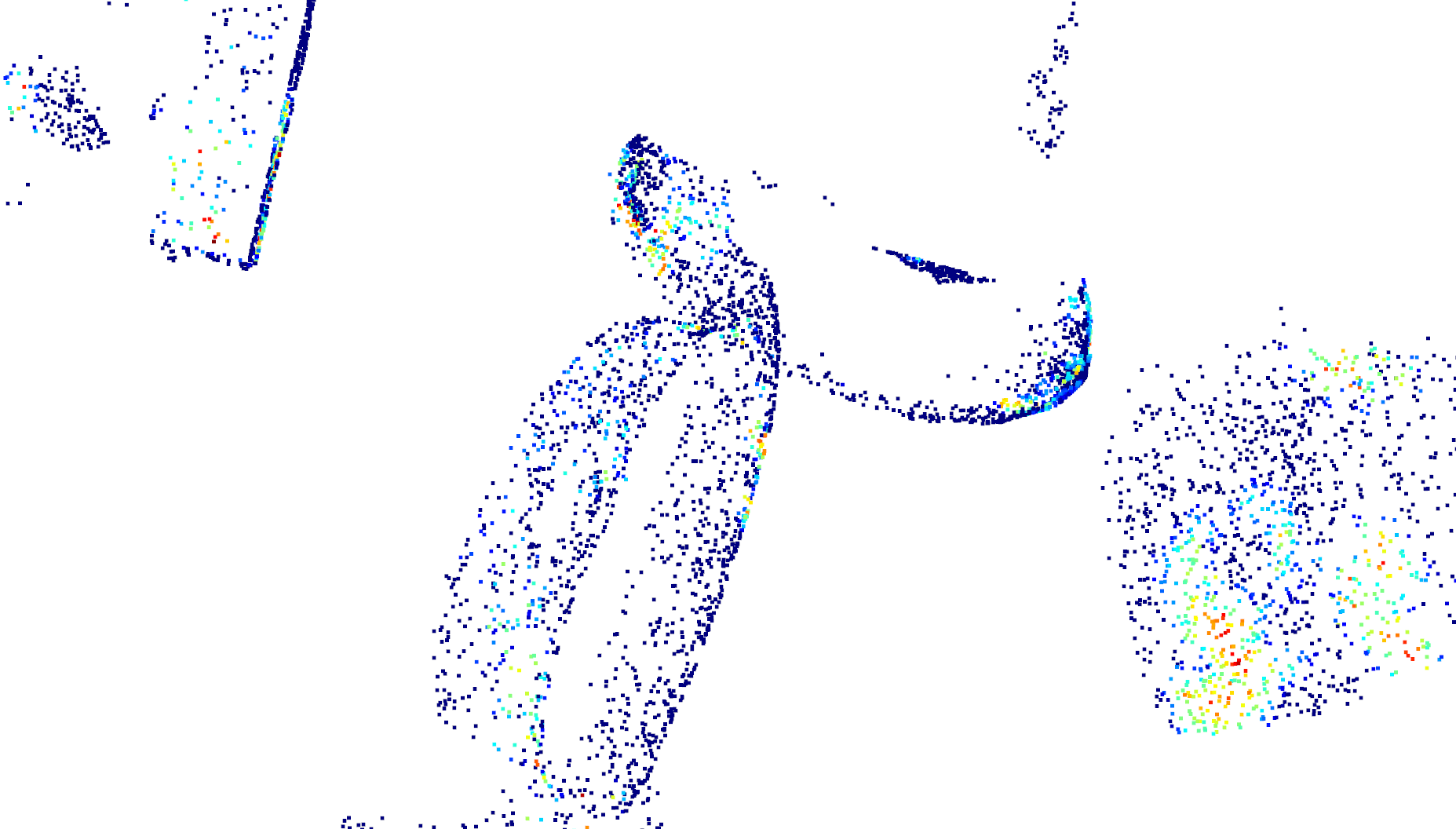}
    \end{subfigure}

    \rule{0.95\textwidth}{0.1pt}

    \begin{subfigure}{0.48\columnwidth}
        \centering
        \includegraphics[width=1\columnwidth, trim={0cm 0cm 0cm 0cm}, clip]{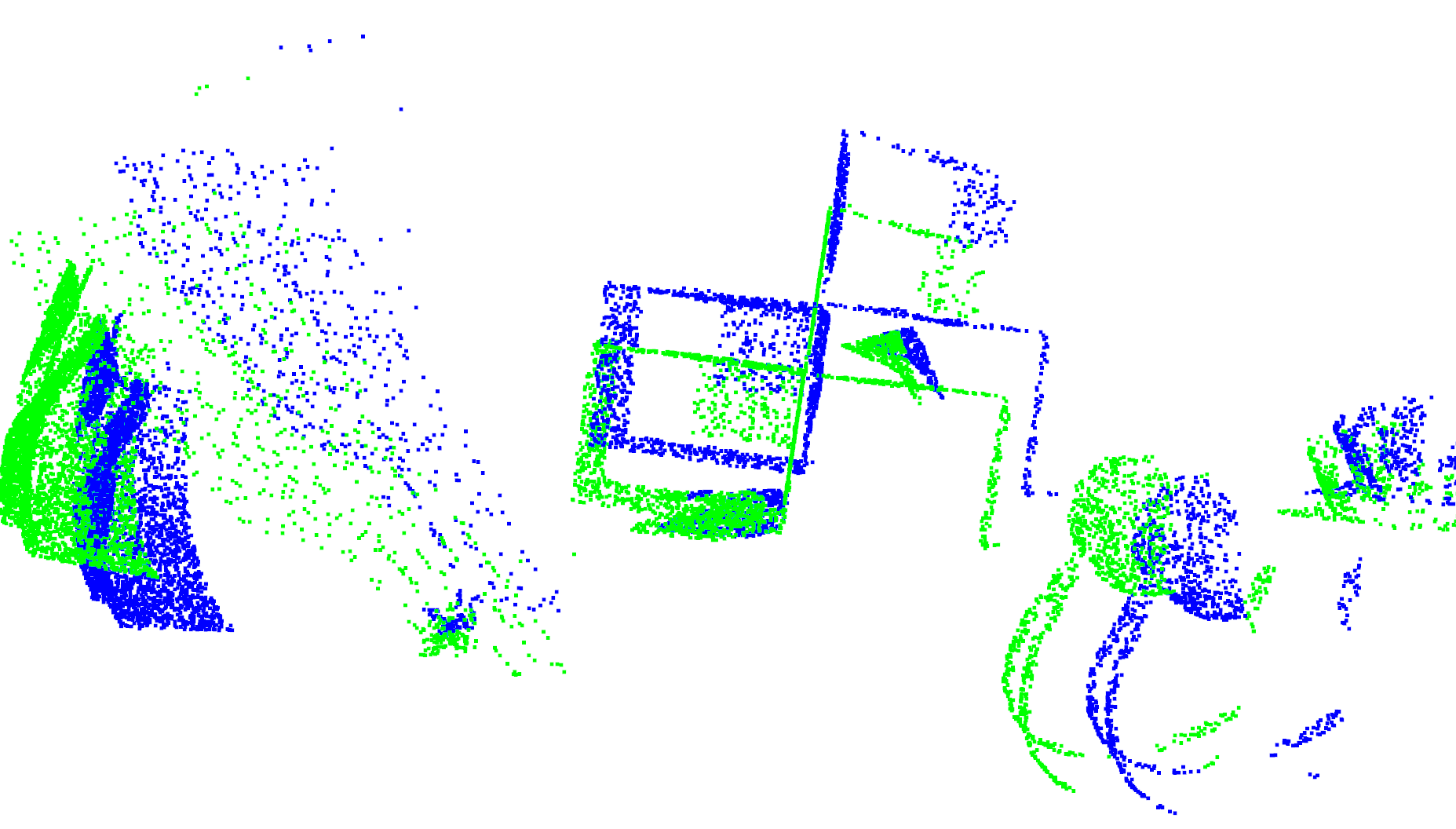}
    \end{subfigure}
    \begin{subfigure}{0.48\columnwidth}
        \centering
        \includegraphics[width=1\columnwidth, trim={0cm 0cm 0cm 0cm}, clip]{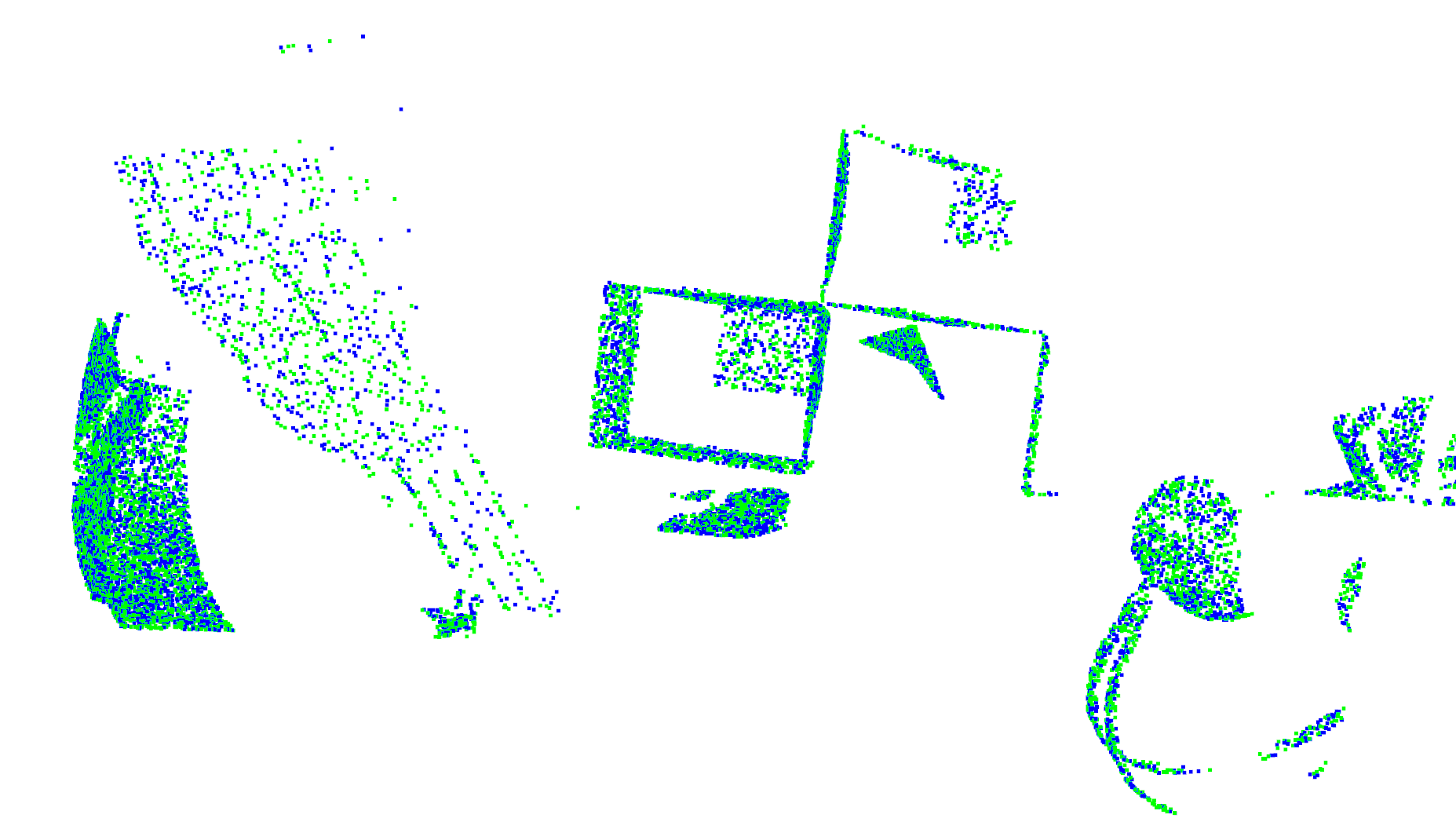}
    \end{subfigure}
    \begin{subfigure}{0.48\columnwidth}
        \centering
        \includegraphics[width=1\columnwidth, trim={0cm 0cm 0cm 0cm}, clip]{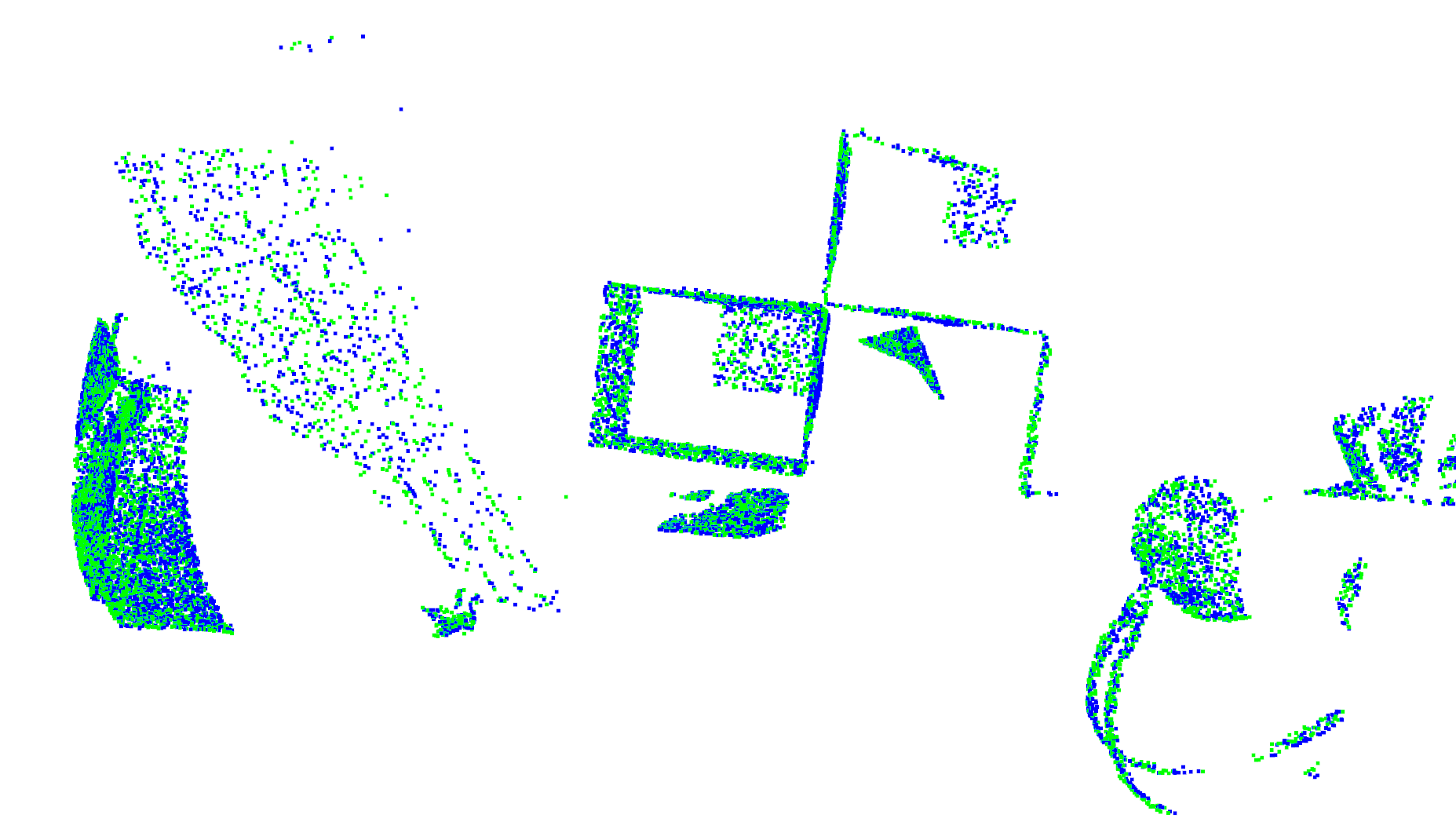}
    \end{subfigure}
    \begin{subfigure}{0.48\columnwidth}
        \centering
        \includegraphics[width=1\columnwidth, trim={0cm 0cm 0cm 0cm}, clip]{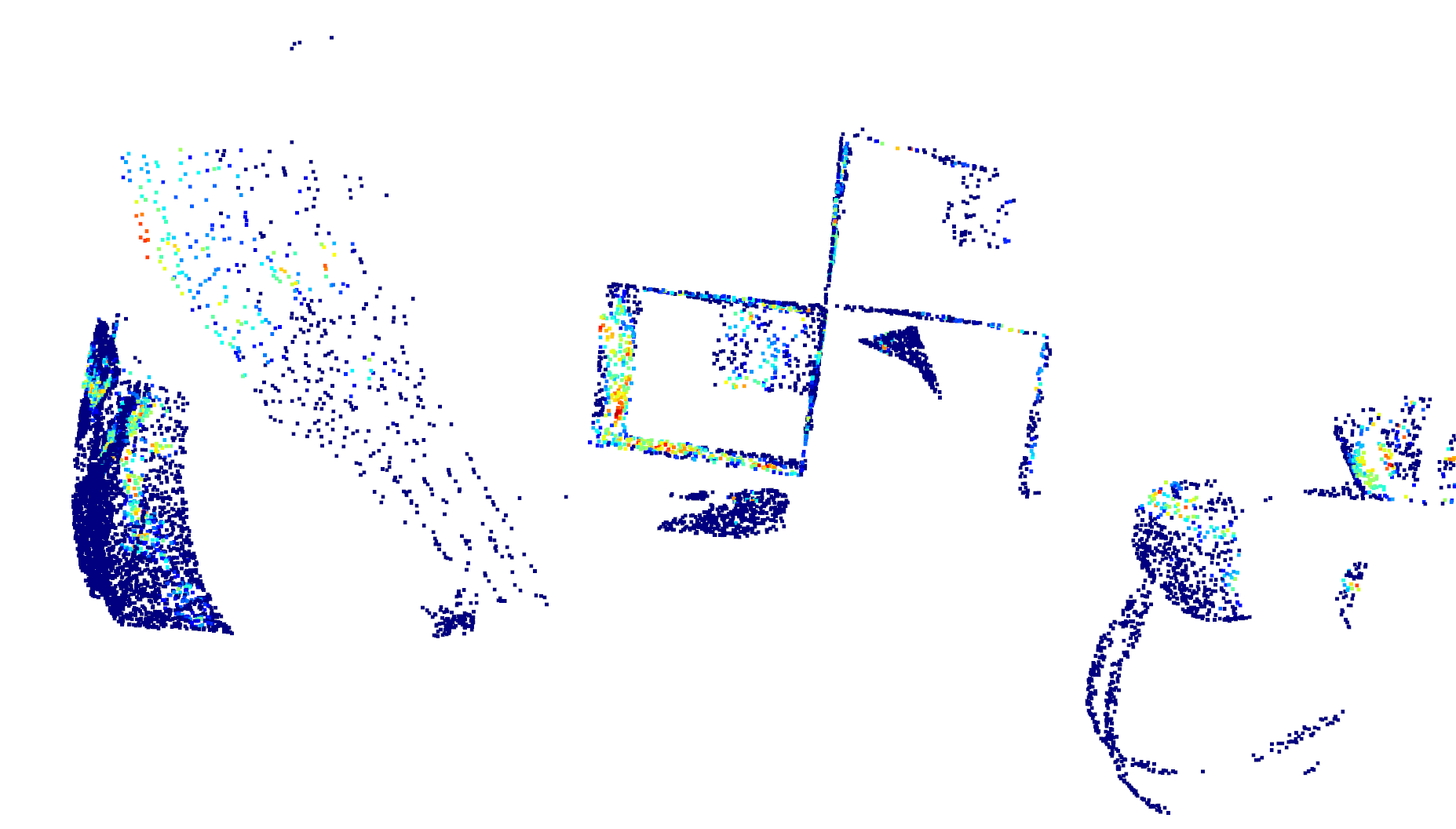}
    \end{subfigure}

    \rule{0.95\textwidth}{0.1pt}

    \begin{subfigure}{0.48\columnwidth}
        \centering
        \includegraphics[width=1\columnwidth, trim={0cm 0cm 0cm 0cm}, clip]{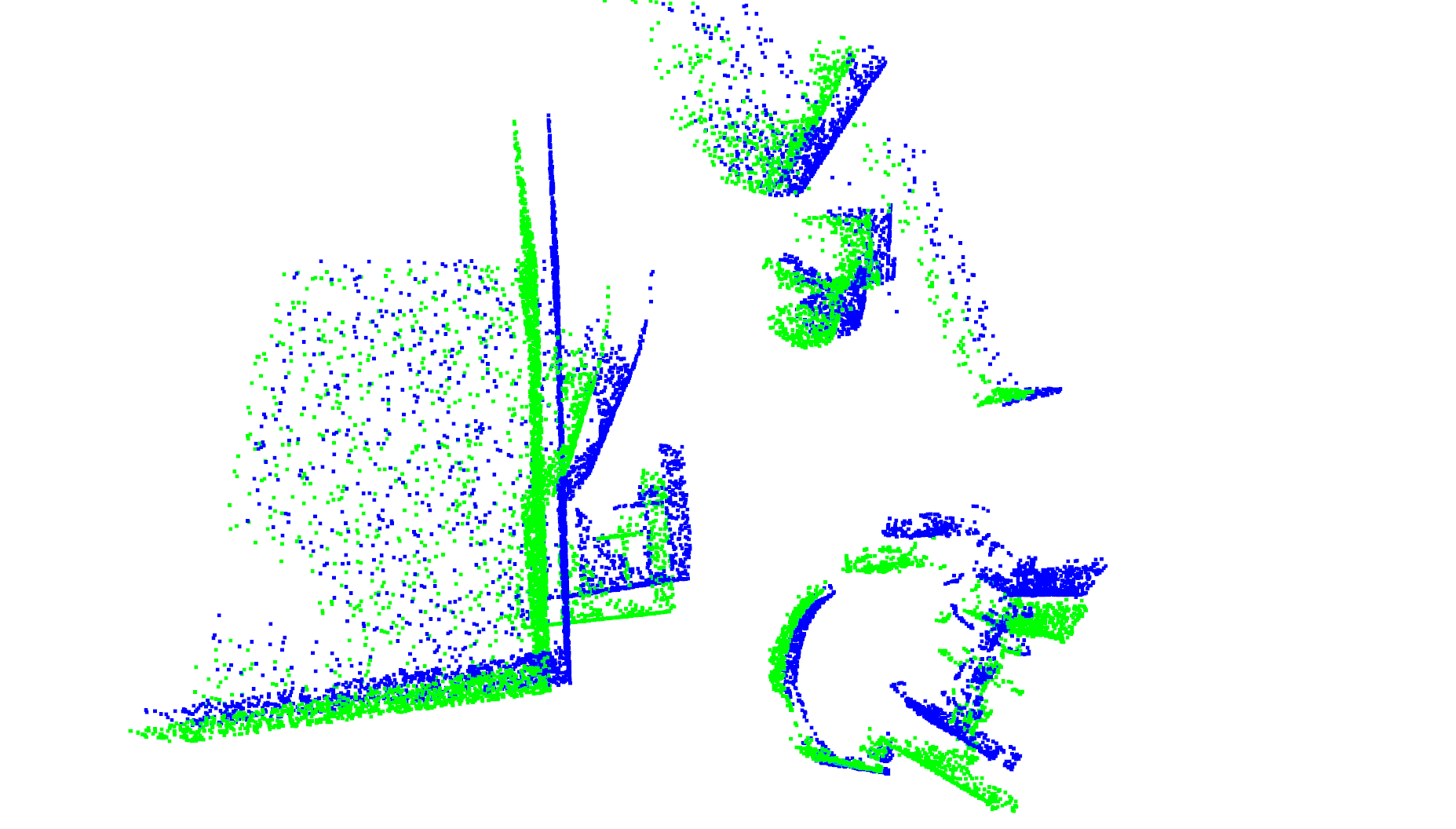}
    \end{subfigure}
    \begin{subfigure}{0.48\columnwidth}
        \centering
        \includegraphics[width=1\columnwidth, trim={0cm 0cm 0cm 0cm}, clip]{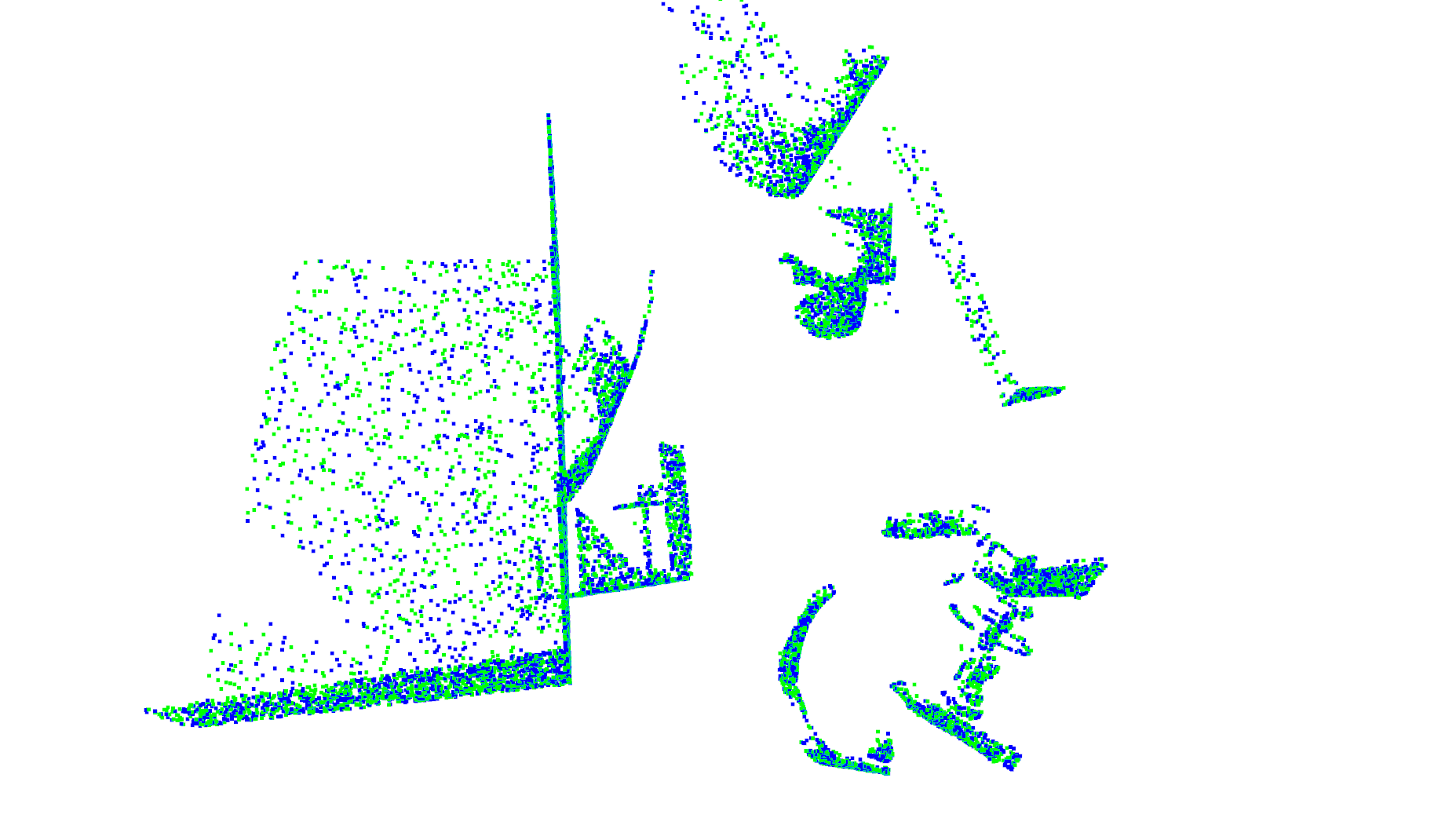}
    \end{subfigure}
    \begin{subfigure}{0.48\columnwidth}
        \centering
        \includegraphics[width=1\columnwidth, trim={0cm 0cm 0cm 0cm}, clip]{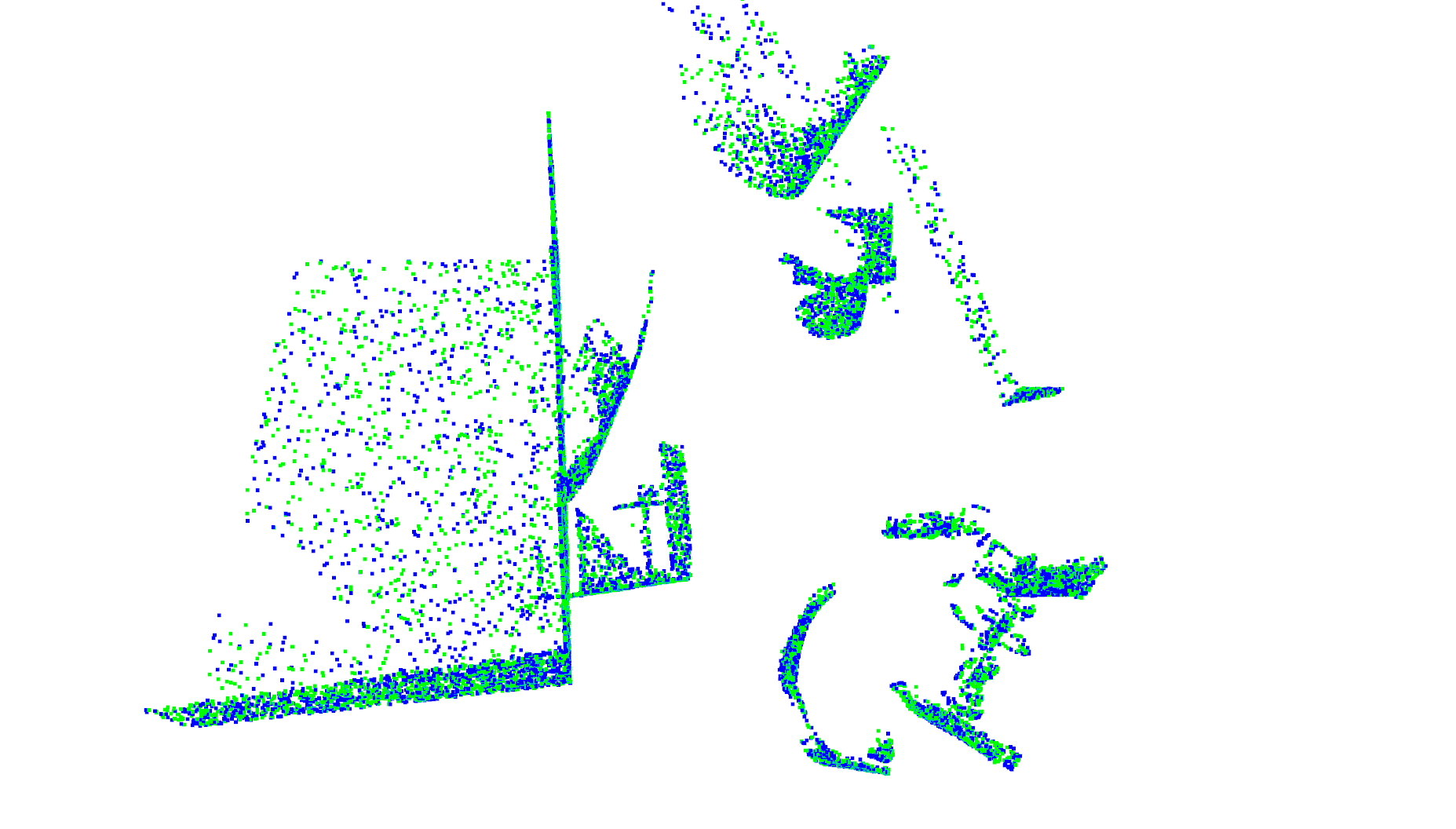}
    \end{subfigure}
    \begin{subfigure}{0.48\columnwidth}
        \centering
        \includegraphics[width=1\columnwidth, trim={0cm 0cm 0cm 0cm}, clip]{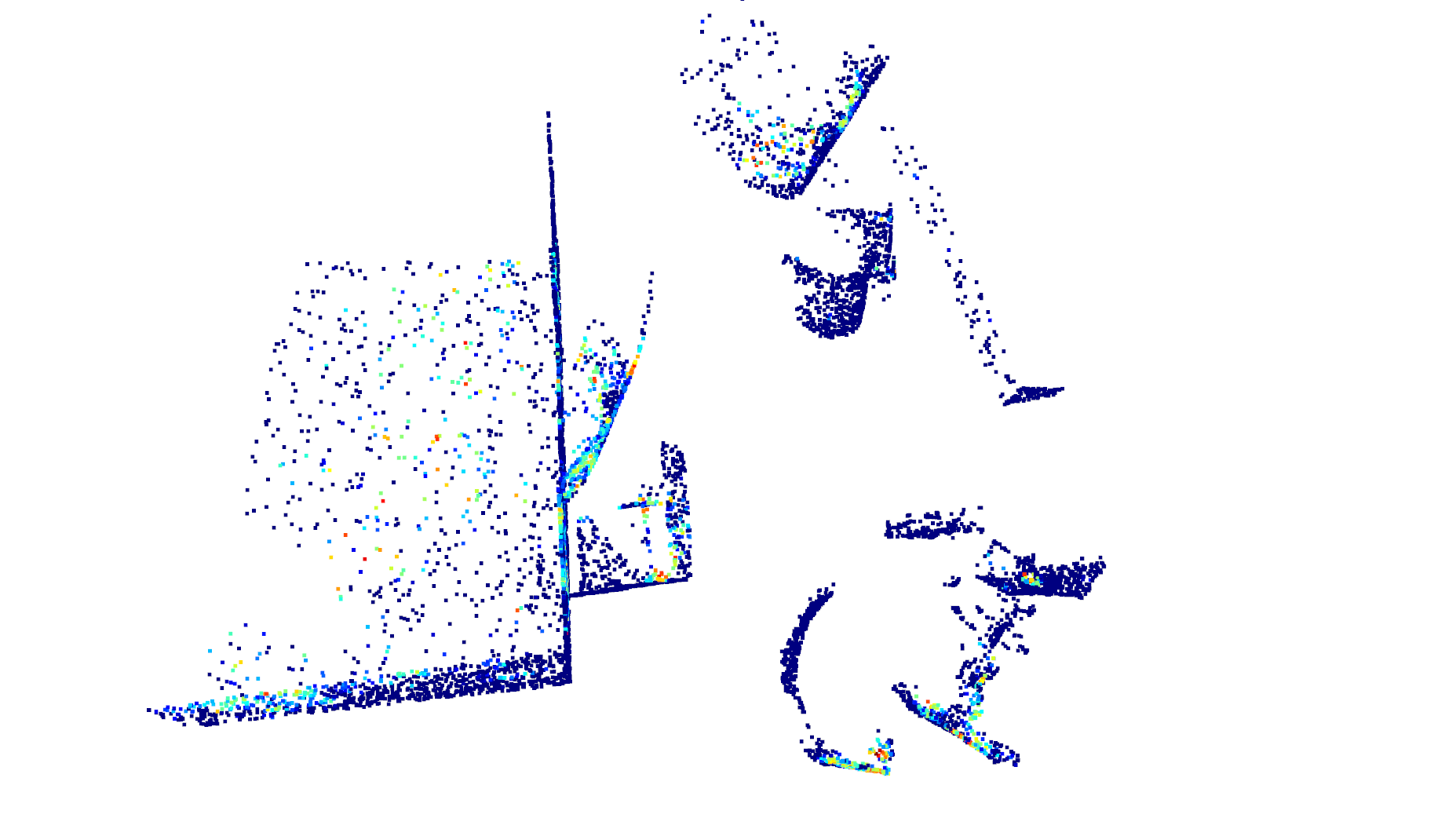}
    \end{subfigure}

    \rule{0.95\textwidth}{0.1pt}
    
    \begin{subfigure}{0.48\columnwidth}
        \centering
        \includegraphics[width=1\columnwidth, trim={0cm 0cm 0cm 0cm}, clip]{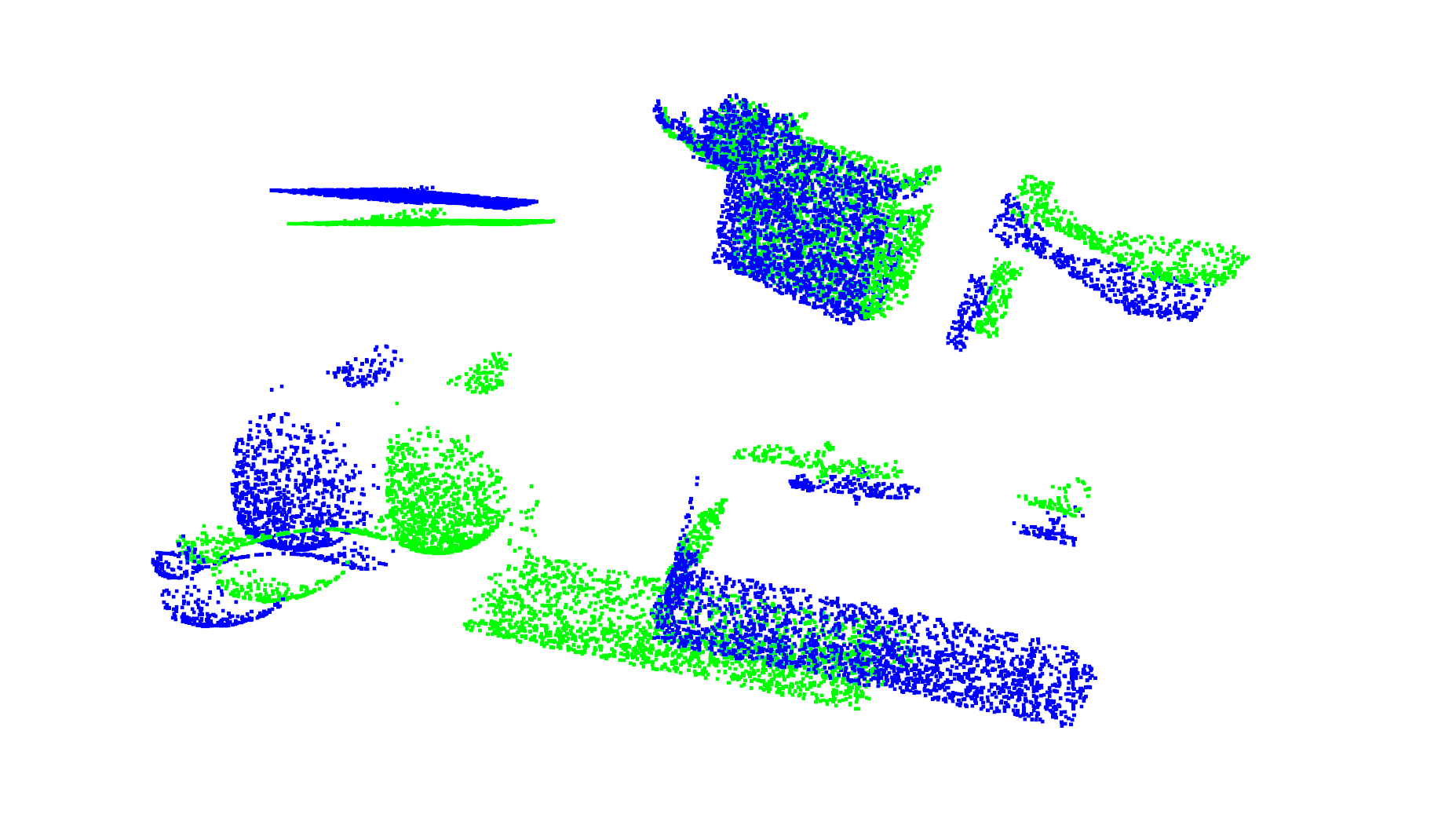}
    \end{subfigure}
    \begin{subfigure}{0.48\columnwidth}
        \centering
        \includegraphics[width=1\columnwidth, trim={0cm 0cm 0cm 0cm}, clip]{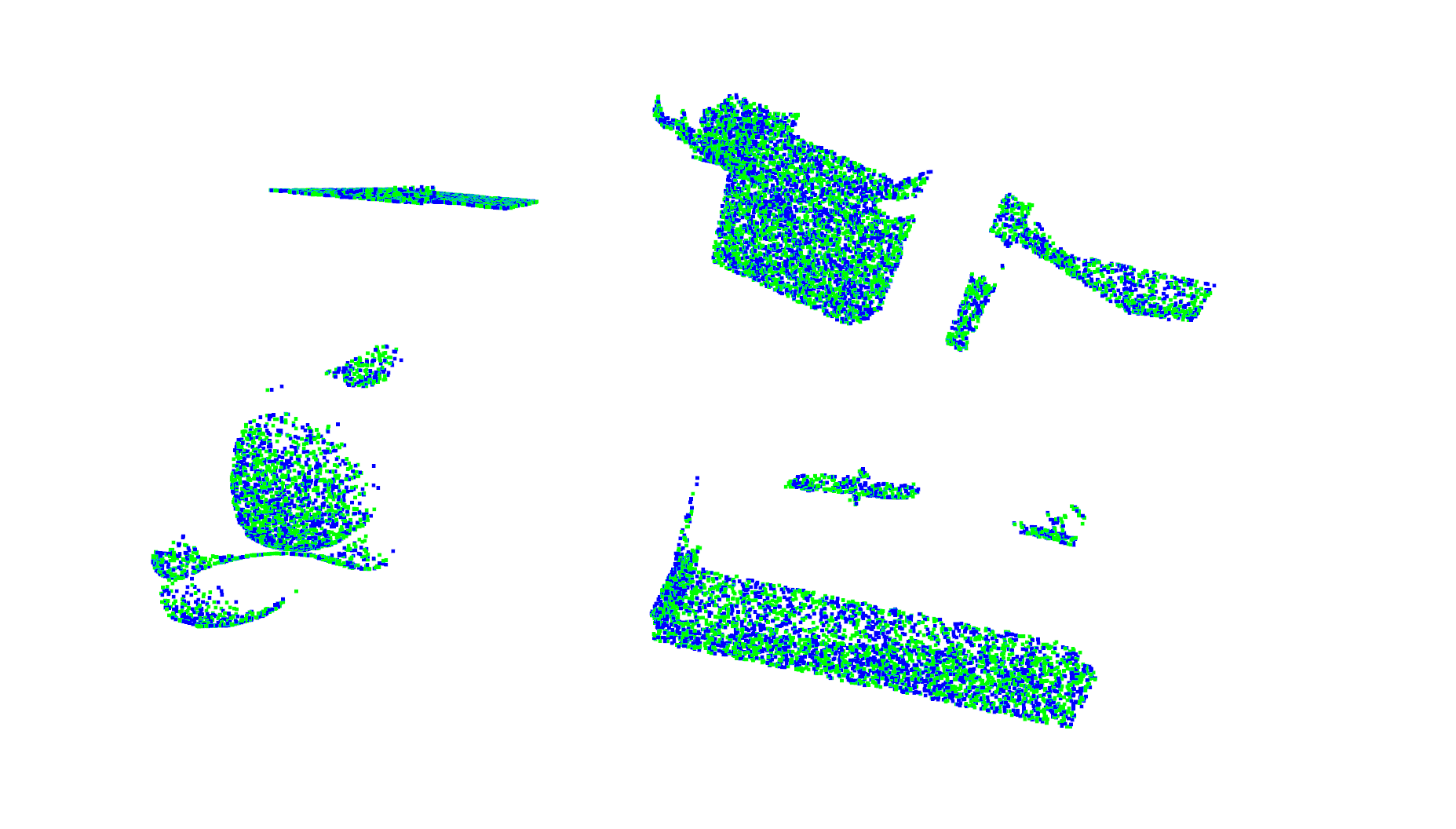}
    \end{subfigure}
    \begin{subfigure}{0.48\columnwidth}
        \centering
        \includegraphics[width=1\columnwidth, trim={0cm 0cm 0cm 0cm}, clip]{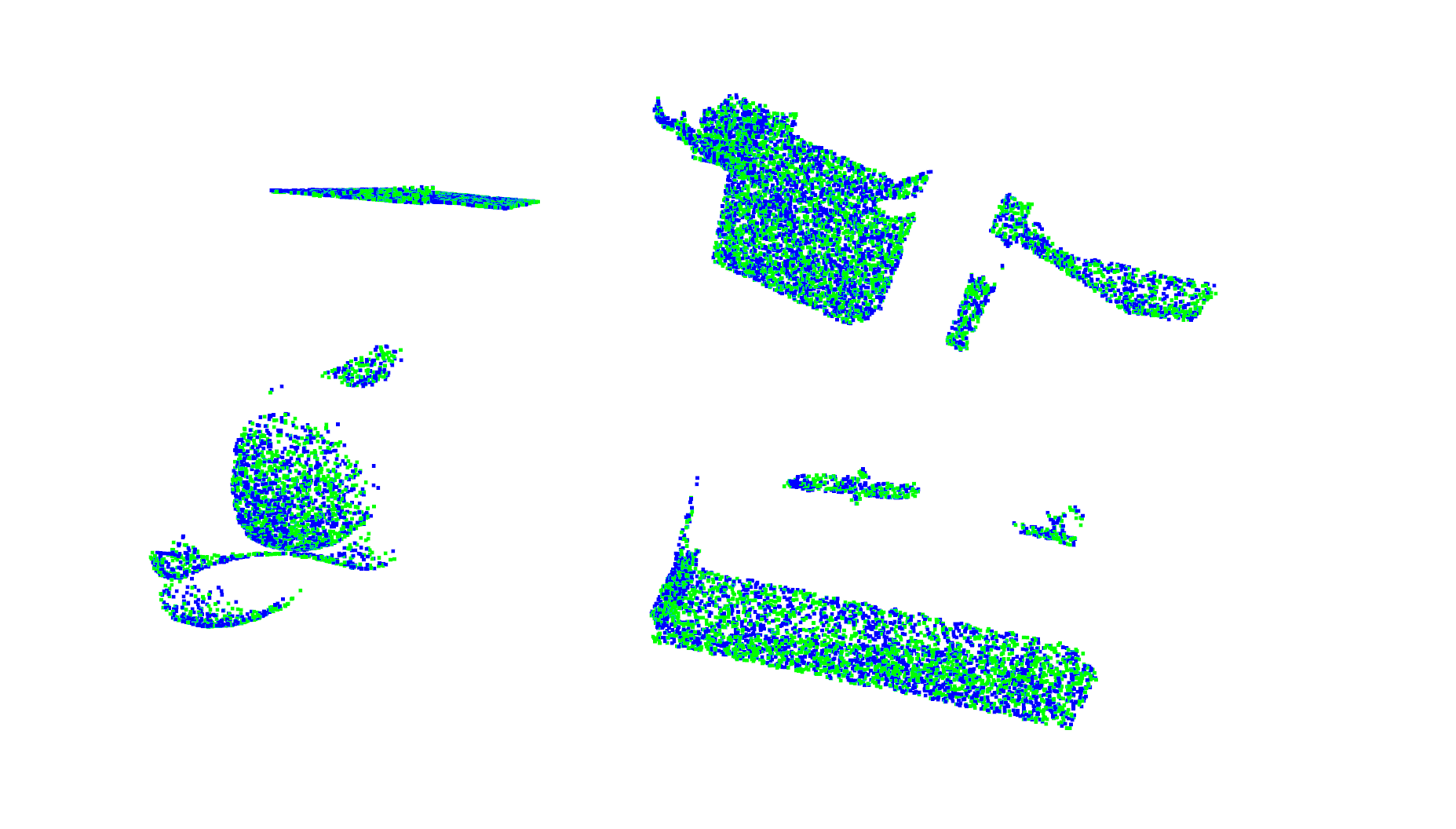}
    \end{subfigure}
    \begin{subfigure}{0.48\columnwidth}
        \centering
        \includegraphics[width=1\columnwidth, trim={0cm 0cm 0cm 0cm}, clip]{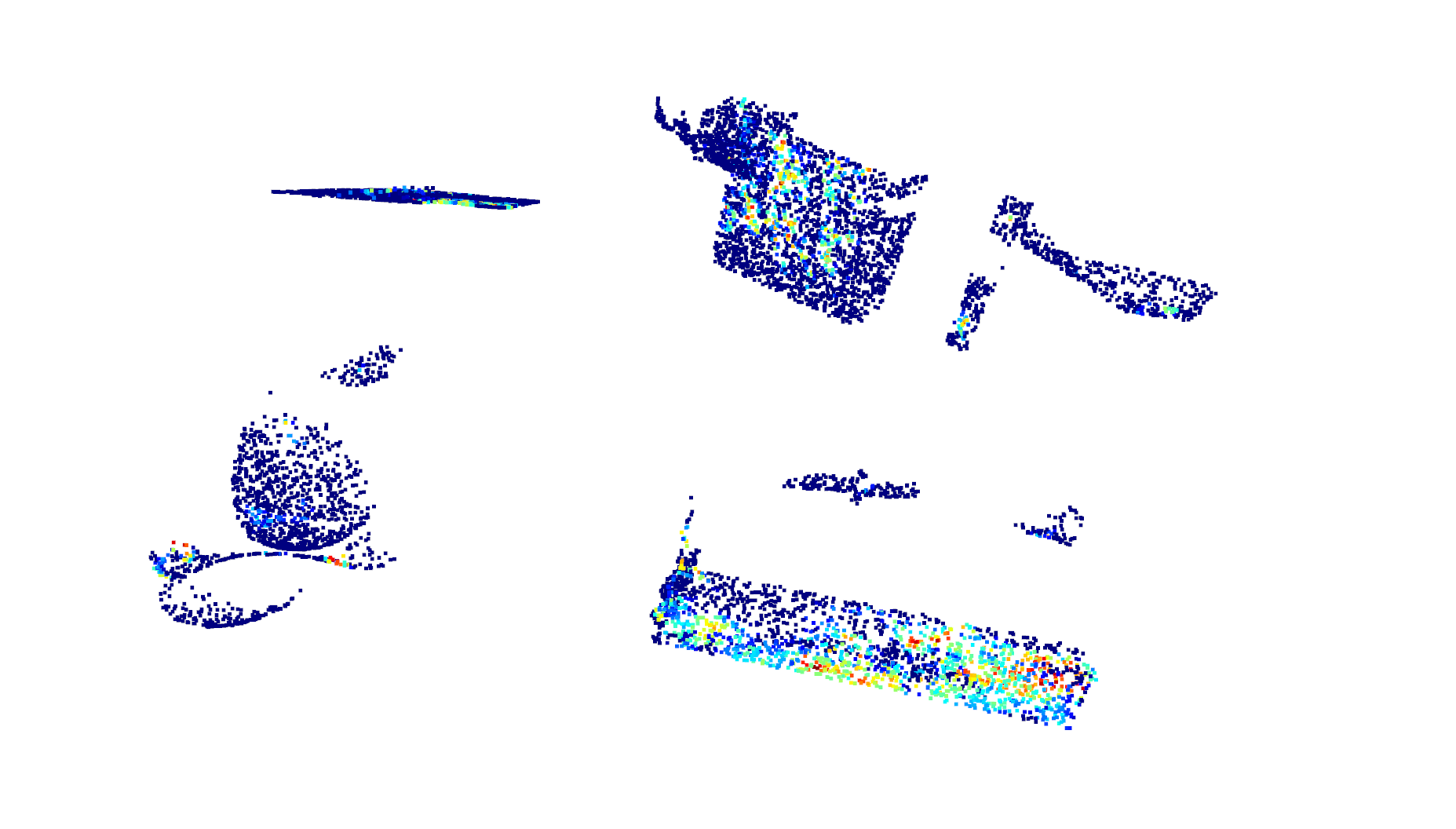}
    \end{subfigure}

    \rule{0.95\textwidth}{0.1pt}

    \begin{subfigure}{0.48\columnwidth}
        \centering
        \includegraphics[width=1\columnwidth, trim={0cm 0cm 0cm 0cm}, clip]{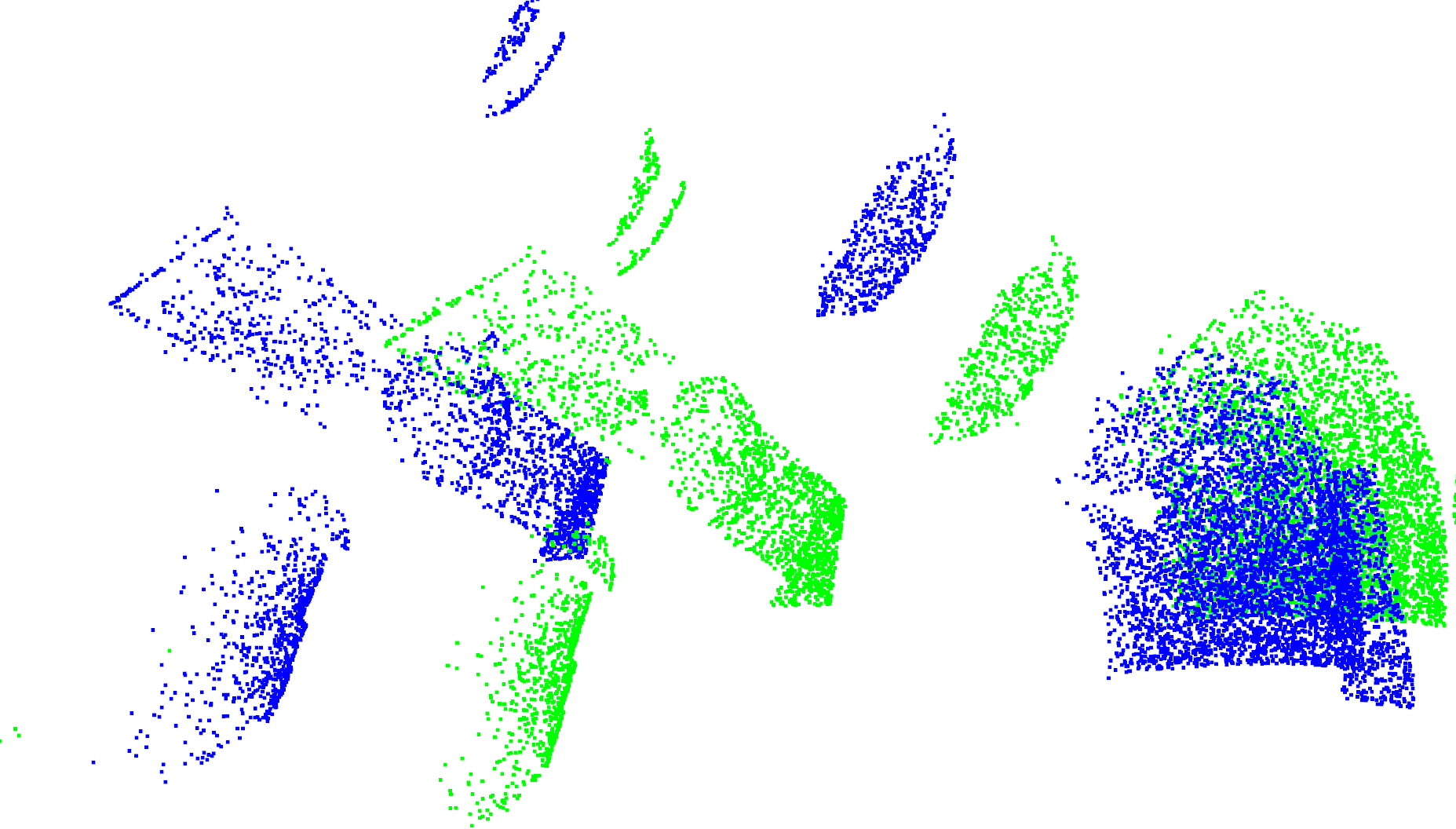}
    \end{subfigure}
    \begin{subfigure}{0.48\columnwidth}
        \centering
        \includegraphics[width=1\columnwidth, trim={0cm 0cm 0cm 0cm}, clip]{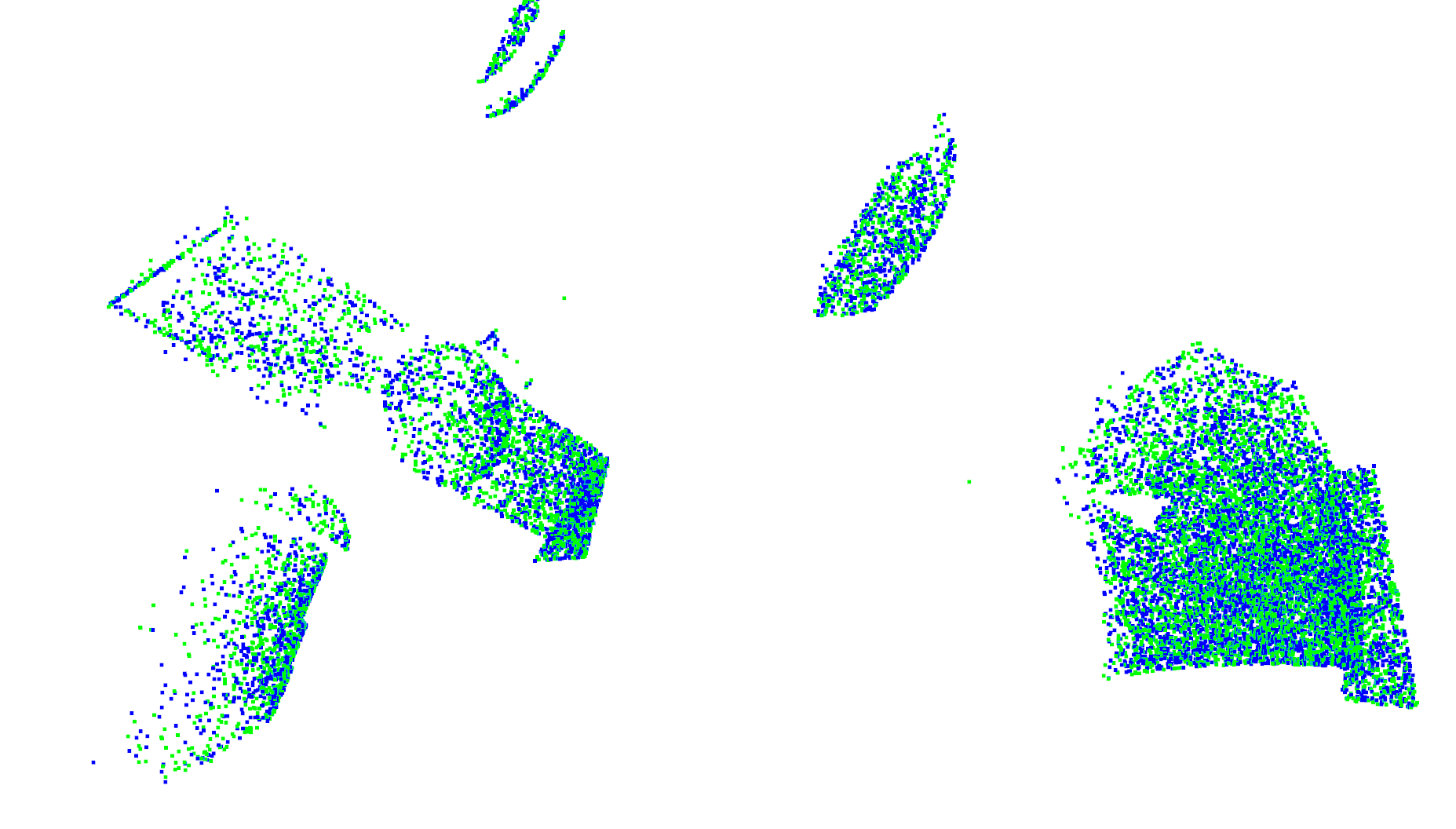}
    \end{subfigure}
    \begin{subfigure}{0.48\columnwidth}
        \centering
        \includegraphics[width=1\columnwidth, trim={0cm 0cm 0cm 0cm}, clip]{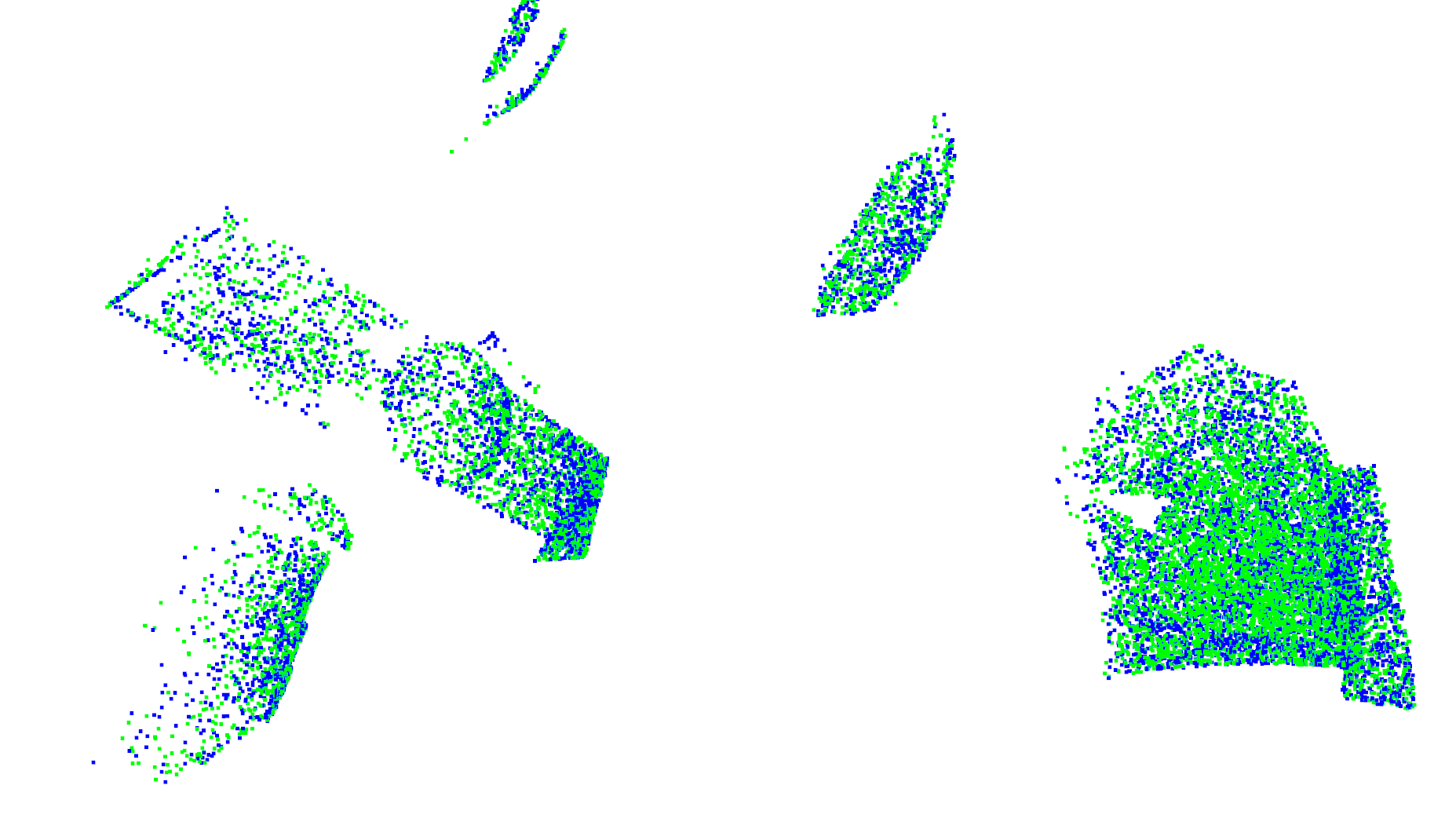}
    \end{subfigure}
    \begin{subfigure}{0.48\columnwidth}
        \centering
        \includegraphics[width=1\columnwidth, trim={0cm 0cm 0cm 0cm}, clip]{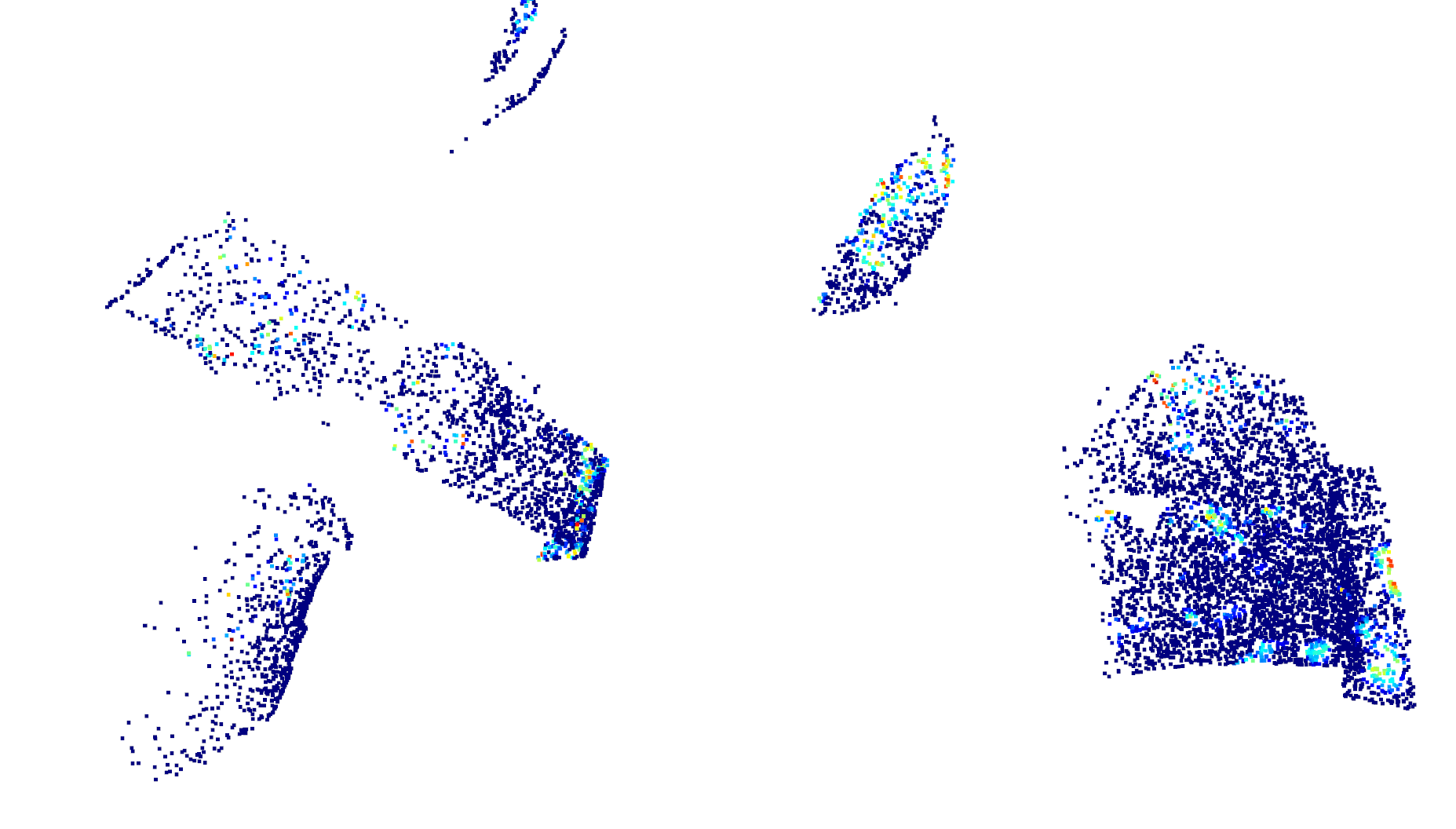}
    \end{subfigure}

    \rule{0.95\textwidth}{0.1pt}
    
    
    \begin{subfigure}{0.48\columnwidth}
        \centering
        \includegraphics[width=1\columnwidth, trim={0cm 0cm 0cm 0cm}, clip]{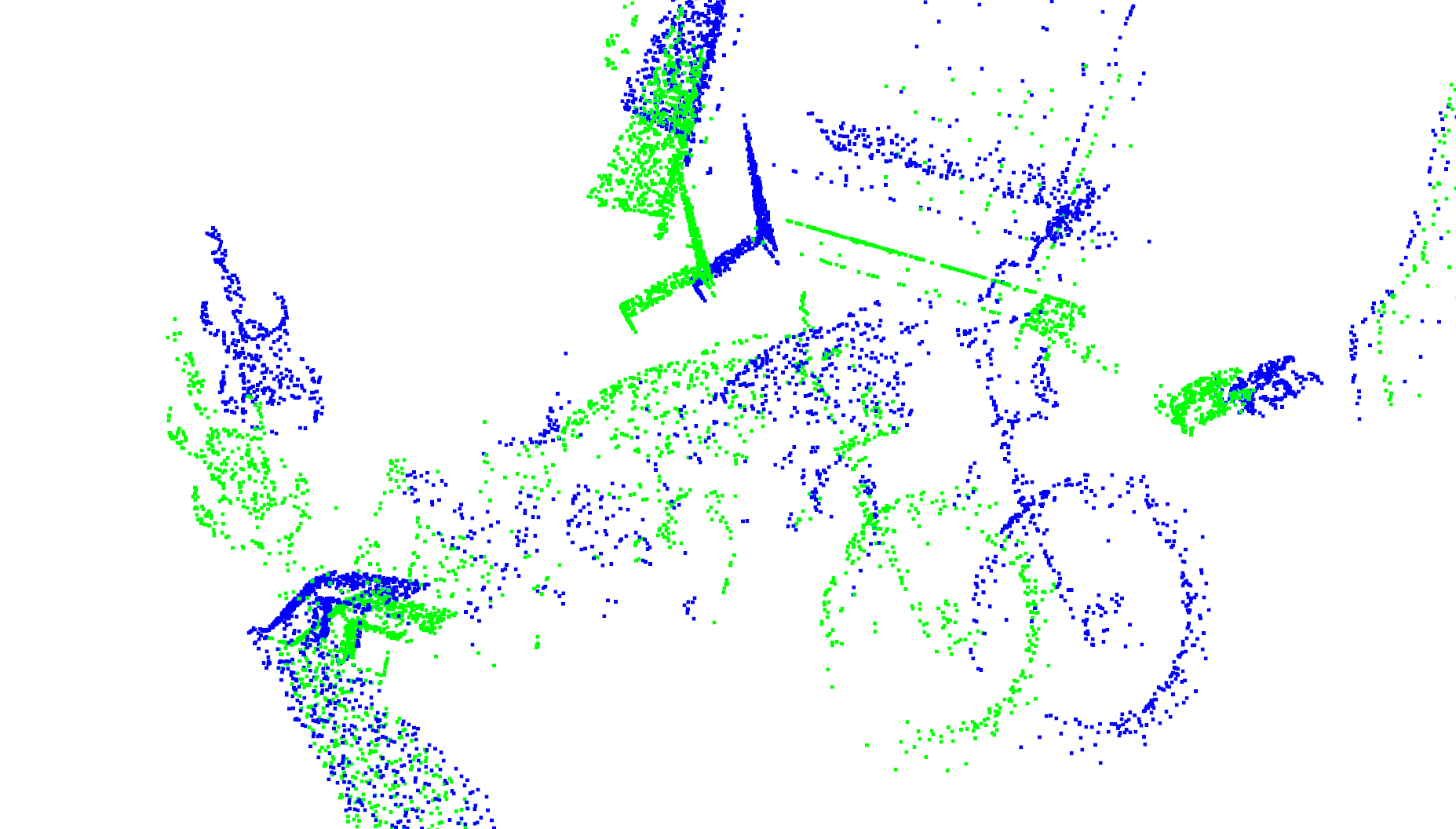}
        \caption*{PC1 and PC2}
    \end{subfigure}
    \begin{subfigure}{0.48\columnwidth}
        \centering
        \includegraphics[width=1\columnwidth, trim={0cm 0cm 0cm 0cm}, clip]{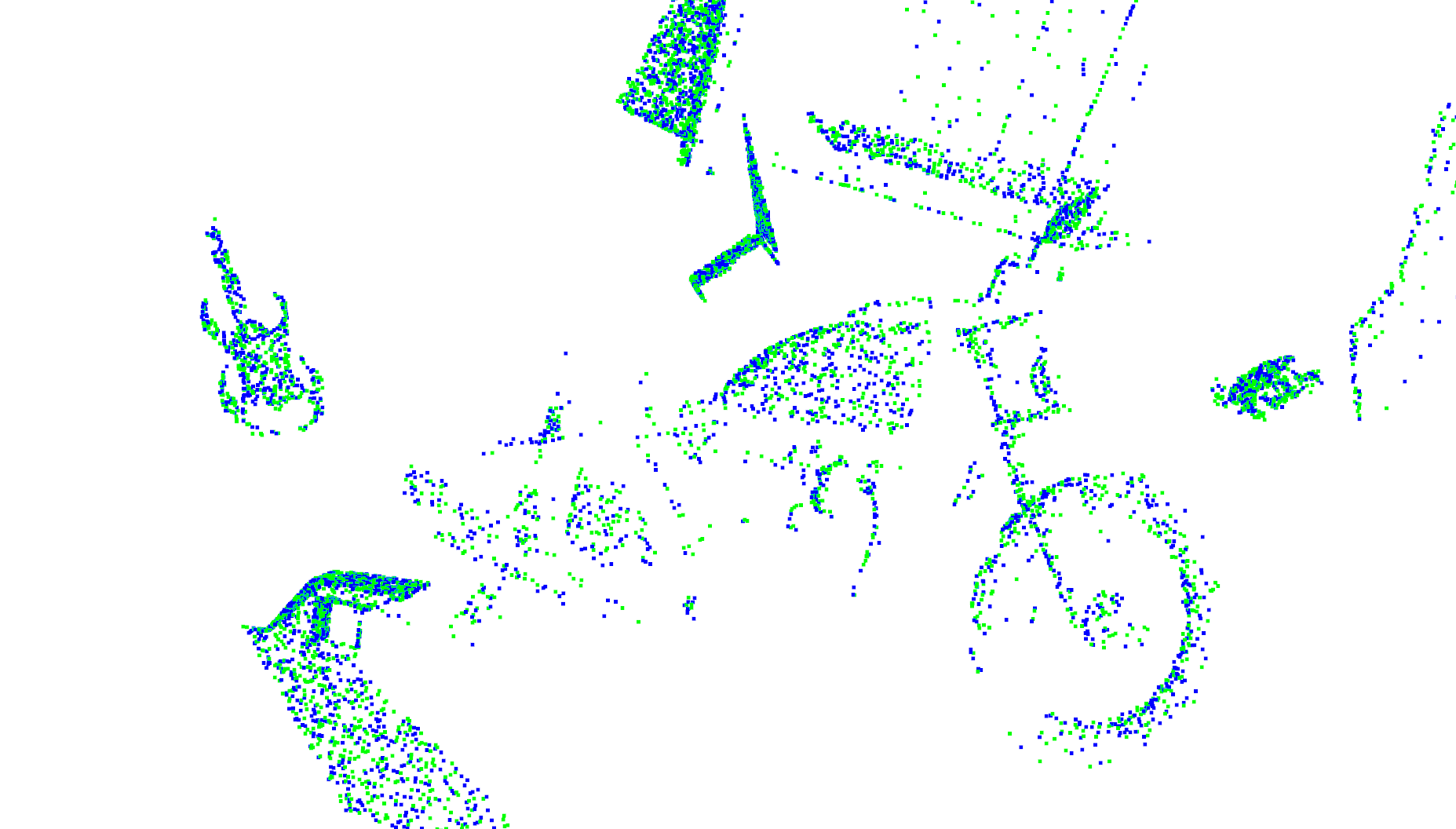}
        \caption*{PC1+GT and PC2}
    \end{subfigure}
    \begin{subfigure}{0.48\columnwidth}
        \centering
        \includegraphics[width=1\columnwidth, trim={0cm 0cm 0cm 0cm}, clip]{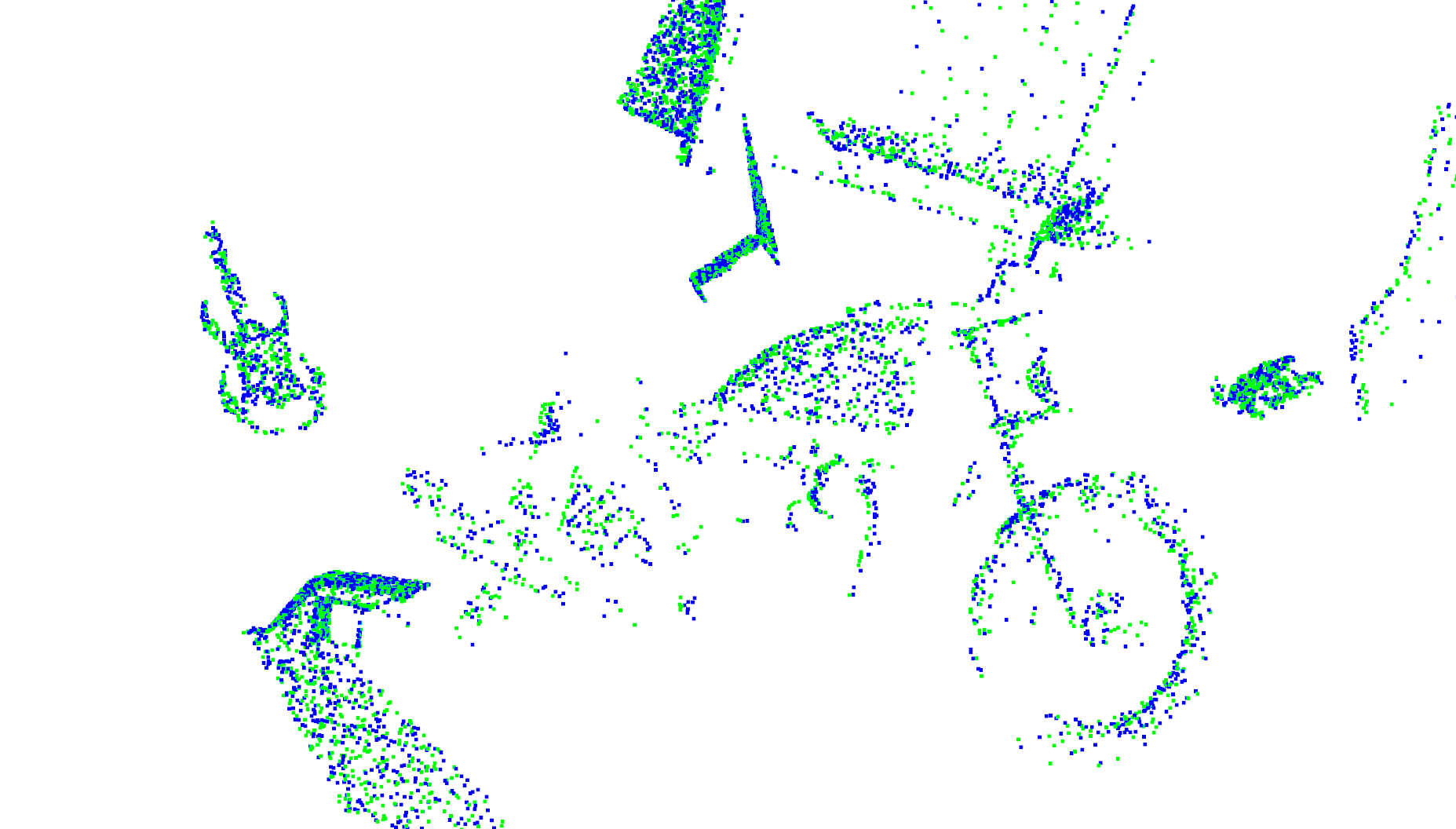}
        \caption*{PC1+Pred and PC2}
    \end{subfigure}
    \begin{subfigure}{0.48\columnwidth}
        \centering
        \includegraphics[width=1\columnwidth, trim={0cm 0cm 0cm 0cm}, clip]{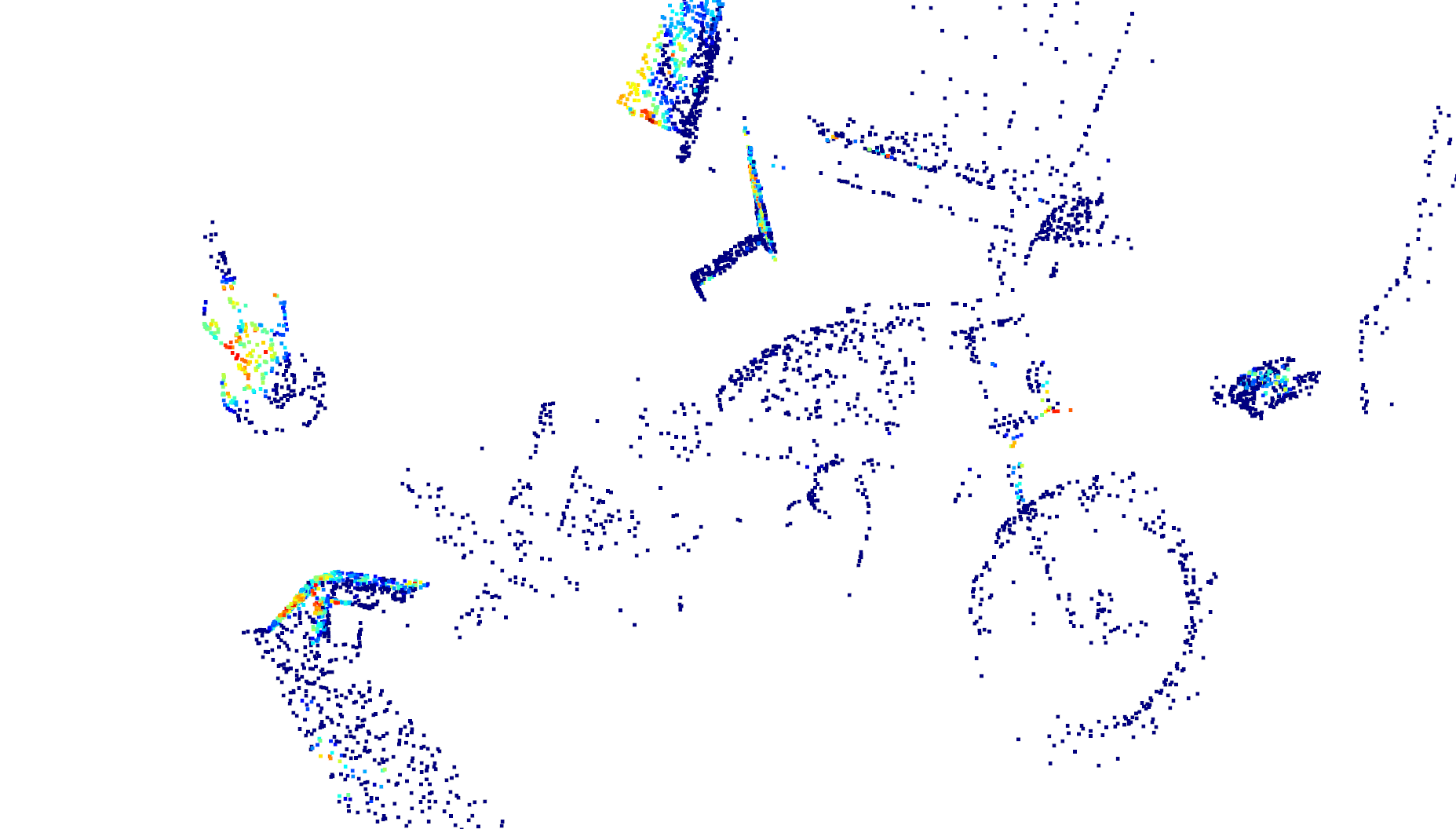}
        \caption*{Error}
    \end{subfigure}
    
    \caption{Qualitative results on the FlyingThings3D~\cite{ft3d} scene flow dataset.}
    \label{fig:res-kitti-supp}
    
\end{figure*}

{\small
\bibliographystyle{ieee_fullname}
\bibliography{reference}
}